\documentclass[journal]{IEEEtran}

\usepackage{multirow}
\usepackage{amsmath}  
\usepackage{amssymb}  
\usepackage{color,array}
\usepackage{tabularx}
\usepackage{xcolor}
\usepackage{graphicx}
\usepackage{setspace}  
\usepackage{makecell}
\usepackage{subcaption}
\usepackage{caption}
\DeclareCaptionLabelSeparator{periodspace}{.\quad}
\usepackage{graphics} 
\usepackage{epsfig} 
\usepackage{times} 
\usepackage{amsmath} 
\usepackage{amssymb}  

\usepackage{bbm} 

\usepackage[linesnumbered,ruled,vlined]{algorithm2e}  
\usepackage{subfloat}

\usepackage{booktabs} 
\usepackage{gensymb}  
\usepackage{bm}  
\usepackage{underscore}  

\usepackage{amsmath} 
\usepackage{amssymb}  

\usepackage{threeparttable}
\usepackage{enumerate}

\usepackage{pifont}
\newcommand{\Checkmark}{\ding{51}}
\newcommand{\XSolidBrush}{\ding{55}}
\usepackage{cite}
\usepackage[colorlinks,urlcolor=blue,linkcolor=blue,citecolor=blue]{hyperref}
\usepackage{xcolor,colortbl}

\usepackage{listings} 
\begin{document}
\title{\LARGE \bf
Follow-Bench: A Unified Motion Planning Benchmark for Socially-Aware Robot Person Following
}

\author{Hanjing~Ye$^{1}$, Weixi~Situ$^{1}$, Jianwei~Peng$^{1}$, Yu~Zhan$^{1}$,
    Bingyi~Xia$^{1}$, Kuanqi~Cai$^{2,3}$, Hong~Zhang$^{1,*}$%
    \thanks{This work was supported in part by Shenzhen Science and Technology Program (No.\ ZDYJ20251211141000001), in part by Shenzhen Science and Technology Program (No.\ SGDX20240115111759002), in part by Meituan Academy of Robotics Shenzhen, in part by the Shenzhen Association for Science and Technology (No.\ XHXS2025-003), and in part by High level of special funds (G03034K003) from Southern University of Science and Technology, Shenzhen, China.}%
    \thanks{$^{1}$Shenzhen Key Laboratory of Robotics and Computer Vision, Southern University of Science and Technology (SUSTech), Shenzhen, China.}%
    \thanks{$^{2}$Human-Robot Interfaces and Interaction Laboratory, Istituto Italiano Di Tecnologia (IIT), Genova, Italy.}%
    \thanks{$^{3}$Swiss Federal Technology Institute of Lausanne (EPFL), Lausanne, Switzerland.}%
    \thanks{$^{*}$Corresponding author: hzhang@sustech.edu.cn.}%
}

\maketitle
\thispagestyle{empty}
\pagestyle{empty}

\begin{abstract}
    Robot person following (RPF)---mobile robots that follow and assist a specific person---has emerging applications in personal assistance, security patrols, eldercare, and logistics. To be effective, such robots must follow the target while ensuring safety and comfort for both the target and surrounding people. In this work, we present the first comprehensive study of RPF, which (i) surveys representative scenarios, motion-planning methods, and evaluation metrics with a focus on safety and comfort; (ii) introduces Follow-Bench, a unified benchmark simulating diverse scenarios, including various target trajectory patterns, crowd dynamics, and environmental layouts; and (iii) re-implements eight representative RPF planners, ensuring that both safety and comfort are systematically considered. Extensive simulation experiments provide a quantitative study of safety- and comfort-related physical indicators, while real-world trials of two MPC-based implementations on a differential-drive robot illustrate deployment feasibility and practical following behaviors. These results reveal open challenges and future research directions. All code, data, and deployment scripts are publicly available at \href{https://follow-bench.github.io/}{https://follow-bench.github.io/.}
\end{abstract}
\def\abstractname{Note to Practitioners}
\begin{abstract}
    This work addresses the challenge of developing RPF systems that maintain both safety (collision-free and continuous target tracking) and comfort (socially acceptable distances and smooth motion) in human-populated environments. While existing surveys provide broad overviews, they lack a systematic discussion of motion-planning principles that jointly consider safety and comfort. To fill this gap, we present a comprehensive survey of RPF-related scenarios, methods, and evaluation metrics, and introduce Follow-Bench, a unified benchmark built on a lightweight simulator for systematic assessment of RPF planners under diverse target, crowd, and environmental conditions. Through systematic simulation comparisons and limited real-world demonstrations of two planners, practitioners can examine planning trade-offs and deployment challenges, and use Follow-Bench to develop and evaluate their own solutions. This research is broadly applicable to RPF scenarios in personal assistance, security patrol, eldercare, and logistics.
\end{abstract}


\section{INTRODUCTION}


\IEEEPARstart{R}{obot} person following (RPF)\cite{islam2019person,socialSurvey2024ijrr,honig2018toward,eirale2025human} is a fundamental capability in human-robot interaction, with applications ranging from personal assistance and security to service robotics. RPF is particularly important for tasks such as eldercare and guided tours in complex, dynamic environments, where robots must be capable of localizing\cite{ye2023robot}, identifying\cite{ye2024reid}, and accompanying\cite{sekiguchi2021uncertainty} a target individual in a socially aware manner. To this end, accompanying behaviors should adhere to principles of safety, comfort, politeness, and context-aware responsiveness~\cite{francis2025principles}.

Although prior surveys\cite{islam2019person,honig2018toward,eirale2025human} have provided broad overviews of its core modules (e.g., perception, tracking and human-robot interaction) and social factors, a dedicated discussion of motion-planning principles that explicitly account for \textbf{safety and comfort} remains lacking.
This paper addresses this gap by centering on these two fundamental criteria, which jointly define how robots can accompany humans in a socially acceptable and physically secure manner.
Prior comfort-aware RPF studies\cite{sun2016ROBIO,montesdeoca2022person,liu2019robot} consider factors such as user comfort, following distance, motion smoothness, and environmental context, but remain tied to individual models rather than a unified, reproducible benchmark for comparing planners under safety- and comfort-critical scenarios.
Here, safety denotes continuous target tracking and reliable collision avoidance, while comfort primarily refers to maintaining appropriate interpersonal position with respect to both the target and surrounding pedestrians. Guided by these requirements, we comprehensively review representative scenarios, planning methods, and evaluation metrics, propose a general RPF-driven motion-planning pipeline, and re-implement eight popular planners to systematically analyze their performance.

To enable systematic performance evaluation, we introduce \textbf{Follow-Bench}, a unified benchmark for RPF planners under diverse conditions, including various target trajectory patterns, pedestrian-flow patterns, and environmental layouts. Unlike existing RPF benchmarks---which focus primarily on perception (e.g., identifying a specific target as \cite{ye2025tpt,chen2017person}) or follow high-level language instructions in simplified environments (e.g., \cite{zhang2024uni} built on Habitat 3.0\cite{puig2023habitat})---Follow-Bench targets the motion-planning capabilities required in realistic, socially interactive settings.

Most existing RPF studies oversimplify human behavior by modeling pedestrians as randomly moving agents, neglecting realistic human interactions, and typically restrict evaluation to domestic or lab-scale environments with only a few pedestrians. Consequently, very few RPF approaches have been systematically evaluated in socially interactive scenarios, for evaluating the motion-planning capabilities for the purpose of accompanying the target person continuously and naturally while avoiding collision with others. For example, Repiso \textit{et al.}\cite{repisoadaptive2020,repiso2024adaptive} evaluate their planners in a simulator that extends the social force model (SFM)\cite{helbing1995social} to account for interactions. However, their evaluations mainly focus on side-following, with simulated environments largely limited to a narrow passage formed by two obstacles and dynamic pedestrians modeled primarily as random crowds. They therefore do not systematically cover broader pedestrian-flow patterns such as parallel flows, circular flows, and perpendicular crossings, or diverse environmental layouts such as doorways, intersections, corridors, and cluttered spaces. Moreover, their scenarios are not released as an open-source benchmark.

In this paper, we address these limitations by introducing Follow-Bench, which explicitly emphasizes two critical requirements for RPF: (i) continuously and naturally following a target person, and (ii) proactively avoiding obstacles and other pedestrians in socially interactive environments. Unlike independent navigation tasks\cite{socialSurvey2024ijrr,stratton2024characterizing,chen2019crowd,tsoi2022sean2}, which focus primarily on collision avoidance, these dual requirements underscore a fundamental trade-off between safety and comfort. Through extensive simulation experiments with eight re-implemented RPF planners, we examine this trade-off using physical indicators; real-world trials of two planners additionally illustrate deployment feasibility and practical limitations. In summary, the main contributions of this paper are:

\begin{itemize}
    \item \textbf{A comprehensive review} of existing RPF-related motion planners, scenarios and evaluation metrics with an emphasis on safety and comfort.
    \item \textbf{A unified benchmark (Follow-Bench)}, implemented in a lightweight simulator with realistic social interactions, for evaluating RPF motion-planning capabilities across diverse target trajectory patterns, pedestrian flows, and environmental layouts.
    \item \textbf{A systematic analysis and comparison} of eight re-implemented RPF planners in Follow-Bench, complemented by real-world feasibility demonstrations of two planners, with discussions of limitations and future research directions.
\end{itemize}

The remainder of the paper is organized as follows: Sec.~\ref{sec:review} reviews existing RPF research with a focus on representative scenarios, evaluation metrics, and motion planners. Sec.~\ref{sec:bench} describes Follow-Bench, including the simulator, scenarios, experimental setup, and evaluation protocols. Simulation and real-world results are presented in Sec.~\ref{sec:simResults}. Sec.~\ref{sec:discussion} discusses challenges, limitations, and future directions, and Sec.~\ref{sec:conclusion} concludes the paper.

\section{Literature Review} \label{sec:review}
The behaviors of autonomous mobile robots in RPF tasks can vary substantially depending on factors such as the application context, robot type, and the assisted person's preferences~\cite{islam2019person,honig2018toward}. In this paper, we focus on two critical and quantifiable requirements that are frequently prioritized: \textbf{safety} and \textbf{comfort}.
\begin{itemize}
    \item \textbf{Safety} is an objective requirement, ensuring that the robot avoids collisions while maintaining continuous accompaniment of the target person.
    \item \textbf{Comfort}, in contrast, is inherently more subjective and is closely related to user preferences. It is physically manifested in the robot's motion patterns and in its relative positioning with respect to the target person and surrounding pedestrians, which can be understood through the lens of proxemics~\cite{hall1963system}.
\end{itemize}
Guided by these two requirements, we comprehensively review representative scenarios (Sec.~\ref{sec:scenarios}) that challenge safety and comfort, and the evaluation metrics (Sec.~\ref{sec:metrics}) used to assess performance with respect to these requirements. Furthermore, the reviewed methods (Sec.~\ref{sec:methods}) are analyzed with an emphasis on how they are designed to address these two key requirements.

\subsection{Scenarios} \label{sec:scenarios}
RPF can occur in a wide range of real-world scenarios. Previous surveys~\cite{islam2019person,honig2018toward,eirale2025human} have typically categorized these scenarios by application domains or broad environmental contexts (e.g., indoor/outdoor, open/crowded spaces). However, such coarse classifications limit the ability to quantify scenario complexity, which is essential for benchmarking the capability of RPF planners. In practice, each scenario arises from a specific combination of factors---the target person's walking behavior, the spatial layout of the environment, and the level of human interference surrounding the robot. Together, these factors determine how challenging it is for a robot to maintain both safety and comfort during following. To enable a more systematic description and comparison of existing studies, we categorize RPF scenarios along three key dimensions: target trajectory pattern, crowd dynamics, and environmental scale and layout, as summarized in Table~\ref{tab:scenarios}.

\subsubsection{Target Trajectory Pattern}
This dimension focuses on how the target person moves. In simple tests, the person follows a predefined shortest path on a map. In more realistic settings, however, their motion includes turns  that increase the difficulty of following. Therefore, the primary source of trajectory complexity lies in the target's turning behaviors, which we classify into smooth and sharp turns according to their difficulty.

Smooth turns are typically associated with natural obstacle avoidance or casual walking patterns. These include standard L-shaped turns and square paths~\cite{sekiguchi2021uncertainty, Kuderer2014Approach, Park2013Autonomous, hu2014tie, yao2021iros, montesdeoca2022person}, circular or figure-eight-shaped trajectories~\cite{sekiguchi2021uncertainty, bayoumi2016learning}, and consecutive S-shaped motions~\cite{leisiazar2025adapting, nikdel2020lbgp, montesdeoca2022person}.
In contrast, sharp turns, which involve abrupt changes in direction, are particularly challenging for RPF systems. A common example is the U-turn~\cite{sekiguchi2021uncertainty, bayoumi2016learning, leisiazar2025adapting, nikdel2020lbgp, eirale2022human, karunarathne2018model}, though other forms such as obtuse-angle turns and triangular trajectories~\cite{sekiguchi2021uncertainty, peng2024dual} also pose significant difficulties.
Leisiazar \emph{et al.}~\cite{leisiazar2025adapting} explicitly evaluated their RPF algorithm's performance under varying target turning radii, highlighting the sensitivity of RPF systems to trajectory geometry.   

The aforementioned turning behaviors typically involve simultaneous changes in both orientation and position during walking. However, in some scenarios, humans may adjust their orientation to a random direction. They would first reorient themselves---i.e., rotate in place---before proceeding in a different direction. This decoupling of orientation and translational movement has been explicitly considered in several RPF-related studies~\cite{leisiazar2025adapting}. A notable distinction is made by Hu~\textit{et al.}\cite{hu2014tie} between step turns and spin turns, with the latter referring to in-place rotations that are particularly common during sharp directional changes. These behaviors introduce additional challenges for RPF systems, especially in target orientation estimation\cite{zhao2024human,hu2014tie} and following position adjustment to avoid obstructing the target's intended path~\cite{sekiguchi2021uncertainty}.

\begin{table}[t]
\centering
\caption{Different dimensions of existing RPF scenarios and extended scenarios (marked in \textbf{bold*}) that are mentioned in related social navigation literature.}
\renewcommand{\arraystretch}{1.2} 
\scalebox{0.62}{%
\begin{tabular}{lll}
\toprule
\textbf{\makecell{Dimension}} & \multicolumn{2}{c}{\textbf{Category and References}} \\
\midrule
\midrule
\multirow{6}{1.7cm}{Target\\Trajectory\\Pattern} & \multirow{2}{*}{Sharp Turn} & Triangle trajectory \cite{sekiguchi2021uncertainty, peng2024dual} \\
& & U-Turn \cite{sekiguchi2021uncertainty, bayoumi2016learning, leisiazar2025adapting, nikdel2020lbgp, eirale2022human, karunarathne2018model} \\
\cmidrule(lr){2-3}
& \multirow{3}{*}{Smooth Turn} & Square; L-Turn \cite{sekiguchi2021uncertainty, Kuderer2014Approach, Park2013Autonomous, hu2014tie, yao2021iros, montesdeoca2022person} \\
& & Circle; ``8'' trajectory \cite{sekiguchi2021uncertainty, bayoumi2016learning} \\
& & Several Turns; ``S'' trajectory \cite{leisiazar2025adapting, nikdel2020lbgp, montesdeoca2022person} \\
\cmidrule(lr){2-3}
& \multicolumn{2}{l}{Rotate then Walk \cite{hu2014tie, sekiguchi2021uncertainty, leisiazar2025adapting}} \\
\midrule
\multirow{10}{1.7cm}{Environmental\\Scale \&\\Layout} & \multirow{3}{*}{Scale} & Small (10m*10m) \cite{eirale2022human} \\
& & Mid (30m*30m) \cite{morales2012how, Bruckschen2020Human} \\
& & Large (100m*100m) \cite{yuan2018laser, karunarathne2018model} \\
\cmidrule(lr){2-3}
& \multirow{2}{*}{Narrow Layouts} & Hallway; Corridor \cite{Park2013Autonomous, eirale2022human, montesdeoca2022person, gross2017roreas, kollmitz2015Time} \\
& & Doorway; Passageway \cite{repisoadaptive2020, xiao2023Learning} \\
\cmidrule(lr){2-3}
& \multirow{1}{*}{Cluttered Obstacles} & Columns; Obstacle Course \cite{Kuderer2014Approach, xiao2023Learning} \\
\cmidrule(lr){2-3}
& \multirow{3}{*}{Semantic Layouts} & Intersection \cite{yuan2018laser, kollmitz2015Time} \\
& & \textbf{Elevator*} \\
& & \textbf{Structured Environments*} \\
\cmidrule(lr){2-3}
& \multirow{1}{*}{3D Topographies} & Stairs; Ramps \cite{zhan2025monocular}; Uneven Terrains \cite{zhan2025monocular,scheidemann2024obstacle} \\
\midrule
\multirow{12}{1.7cm}{Crowd\\Dynamics} & \multirow{4}{*}{Pedestrian Number} & One \cite{gross2017roreas} \\
& & Mid (2-10) \cite{goldhoorn2014continuous, leisiazar2025adapting, chen2017person} \\
& & Large (10-20) \cite{kastner2022human} \\
& & \textbf{Dense Crowds*} \\
\cmidrule(lr){2-3}
& \multirow{6}{*}{Directionality and Pattern} & Parallel Traffic \cite{gross2017roreas, liu2019robot, repiso2017line} \\
& & Random Crowds \cite{repisoadaptive2020,repiso2024adaptive} \\
& & \textbf{Perpendicular Traffic*}\cite{wang2022metrics, kollmitz2015Time} \\
& & \textbf{Circular Traffic*}\cite{wang2022metrics} \\
& & \textbf{Join / Leave a Crowd*}\cite{tsoi2022sean2} \\
\cmidrule(lr){2-3}
& \multirow{3}{*}{Motion Policy} & SFM / ESFM \cite{liu2019robot, ferrer2014proactive, kastner2022human} \\
& & \textbf{ORCA*} \\
& & CV \cite{leisiazar2023mcts, peng2024dual} \\
\bottomrule
\end{tabular}
} 
\label{tab:scenarios} 
\vspace{-0.3cm}
\end{table}

\subsubsection{Environmental Scale \& Layout} 
This dimension concerns the environmental scale and layout in which RPF is performed. Small or medium-sized indoor spaces~\cite{eirale2022human,morales2012how,Bruckschen2020Human} often cause frequent occlusions due to dense obstacles and limited visibility, but re-finding the target is easier within a restricted area. Conversely, in large-scale venues such as shopping malls or open plazas~\cite{karunarathne2018model}, occlusions are fewer, yet the search and re-identification process becomes harder because the target can move towards a much broader region.

In addition to scale, environmental layout is also a relevant factor in RPF complexity. Narrow structures---hallways or corridors, for example---force the robot to keep a safe yet functional distance from both the target person and nearby pedestrians while navigating tight clearances \cite{Park2013Autonomous,eirale2022human,montesdeoca2022person,gross2017roreas,kollmitz2015Time}. When a wide space funnels into a doorway or passage, the follower often must switch its relative position to maintain line-of-sight and safety \cite{repisoadaptive2020,xiao2023Learning}. Cluttered obstacles similarly demand a constantly shifting balance between obstacle avoidance and target pursuit \cite{Kuderer2014Approach,xiao2023Learning}. In addition, semantic cues embedded in the layout call for context-sensitive behavior: at intersections, the robot must anticipate cross-traffic and blind corners \cite{yuan2018laser,kollmitz2015Time}; in elevators, it should shrink its proxemic radius to respect the tighter personal space \cite{francis2025principles}; and in highly repetitive patterns—waiting-room bench grids, supermarket aisles, or office cubicles—it can exploit spatial regularity to plan more predictable and efficient paths. Finally, 3D topographical variations, such as ramps, stairs, and uneven terrains, introduce additional challenges related to stable human localization\cite{zhan2025monocular}, locomotion and safe navigation\cite{scheidemann2024obstacle}. Collectively, these factors underscore that successful RPF requires strategies tuned not only to environmental scale but also to the nuanced features of the surrounding layouts.

\subsubsection{Crowd Dyanmics} 

The complexity of RPF increases in crowded scenes due to occlusion and collision risks caused by non-target pedestrians. Especially due to human-human interactions, pedestrian future motion is diverse and varied. In existing RPF simulated environments, some methods assume simplified crowd motion with constant velocity models \cite{leisiazar2023mcts,peng2024dual} without considering interactions, while some methods simulate the interactions via SFM or Extended SFM \cite{liu2019robot,ferrer2014proactive,kastner2022human}. ORCA is another interaction simulation method, widely adopted in social navigation \cite{francis2025principles,samavi2024sicnav}---shows promise for dense crowds, but its application to RPF remains unexplored.

As for the pedestrian number settings, current evaluations primarily target sparse scenarios. Some studies assume one other person\cite{gross2017roreas} or a small group of 2-10 people \cite{goldhoorn2014continuous,leisiazar2025adapting,chen2017person}. A few studies evaluate RPF performance in crowds with 20 people\cite{kastner2022human}. Although some scenarios involve many pedestrians, the people density of these scenarios is usually not provided. Besides, evaluations involving dense and dynamic crowds, like cafeterias and crossings, are limited.

Furthermore, directionality and patterns of pedestrian flows, including crossing\cite{kollmitz2015Time,leisiazar2023mcts,liu2019robot,chen2017person}, parallel passing\cite{gross2017roreas, kollmitz2015Time,liu2019robot,repiso2017line}, and getting stuck in traffic\cite{kollmitz2015Time,kastner2022human}, also impact the robot's following policy. For instance, the robot should leave a space for the target person to escape from getting stuck in a crowd \cite{kastner2022human}. Apart from the above factors, additional target-crowd interactions should be considered, e,g., accompanying the target to cross a perpendicular traffic\cite{kollmitz2015Time, wang2022metrics, chen2017person}, and joining or leaving a crowd\cite{tsoi2022sean2}.

\subsection{Evaluation Metrics}\label{sec:metrics}

\begin{figure}[t]
    \centering
    \includegraphics[width=0.8\linewidth]{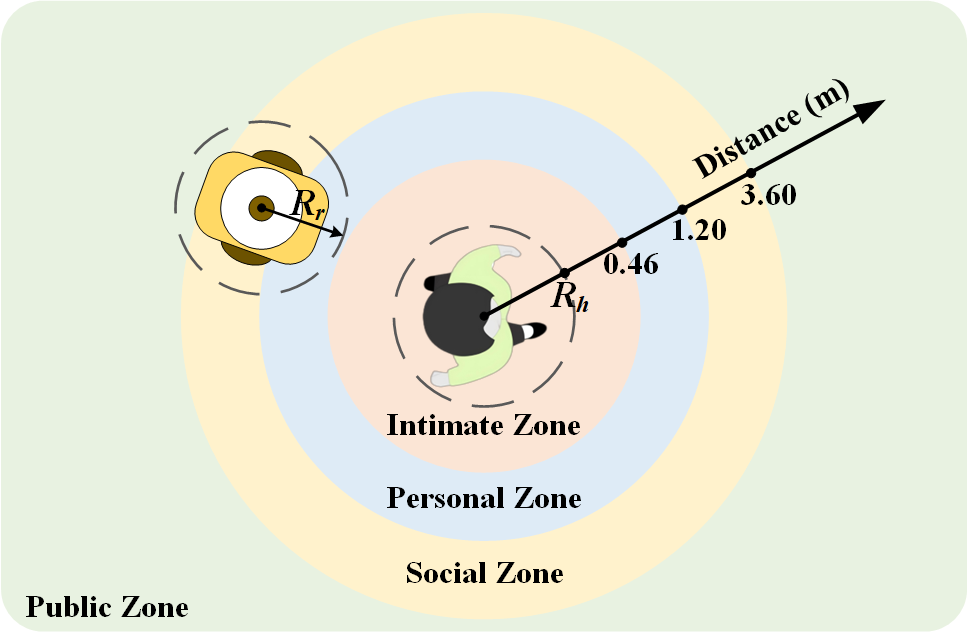}
    \caption{Social space definition. A desired following position is within the personal zone (0.45-1.2 meters) of the target, while avoiding the private zone (less than 0.45 meters) of surrounding pedestrians. The distance is measured as $|d_{rh}-R_h-R_r|$, where $d_{rh}$ is the distance between the robot and the target, $R_h$ is the radius of the target, and $R_r$ is the radius of the robot.}
    \label{fig:socialSpace}
    \vspace{-0.3cm}
\end{figure}

Safety and comfort are among the most frequently emphasized criteria in RPF research~\cite{francis2025principles,honig2018toward}. Consider a scenario in which a target person moves through an environment populated with static obstacles and surrounding pedestrians. An ideal person-following robot should (i) avoid collisions and maintain continuous observation of the target, (ii) move with smooth and stable motion, and (iii) preserve a socially acceptable distance from both the target and nearby people. The core evaluation metrics corresponding to these criteria are summarized in Table~\ref{tab:metrics}. Compared with the broadly reviewed metrics in prior surveys~\cite{honig2018toward,eirale2025human}, the metrics summarized here are streamlined to focus specifically on motion-planning-related aspects. Moreover, they are objectively quantifiable and categorized into two groups---safety-related and comfort-related metrics---to enable clearer capability analysis.

Regarding safety, a widely used metric is the avoidance success rate (ASR), which measures the percentage of trials completed without collisions~\cite{kastner2022human}. Another key safety-related metric is the search success rate (SSR), defined as the percentage of times the robot successfully re-finds the target within a limited time after occlusions~\cite{ye2025rpf,goldhoorn2018searching,goldhoorn2017searching}. To further assess search efficiency, the success weighted by path length (SPL) metric is often adopted~\cite{ye2025rpf}, jointly considering both SSR and the total path length taken during the search. Typically, ASR and SSR are combined into an overall success rate (SR), reflecting the overall completion performance of the RPF task.

Comfort is often assessed in terms of motion smoothness, typically evaluated through velocity, acceleration, and jerk \cite{kobayashi2015design, martini2024adaptive}. It also depends on the position between the robot and the target. Following proxemic theory \cite{hall1963system}, previous works \cite{peng2024dual, sekiguchi2021uncertainty} define a preferred region around the target, as illustrated in Fig.~\ref{fig:socialSpace}. Maintaining a position within the personal zone (0.45-1.2 meters) is generally considered desirable, and researchers often compute the duration the robot remains within this zone. In addition to distance preference, angle preference is sometimes also important (e.g., maintaining a side-following position). For instance, Repiso \textit{et.al}~\cite{repiso2017line} proposes a metric that measures the angular deviation from the preferred following angle. 

Moreover, the target visibility ratio (TVR) is another important comfort-related indicator, quantifying the proportion of time the robot successfully keeps the target within its field of view~\cite{koide2016reid,chen2022lopf,goldhoorn2018searching,goldhoorn2017searching}. The position between the robot and surrounding pedestrians also affects perceived comfort. Studies on social navigation~\cite{kastner2024arena,cai2022human,chen2017socially} suggest maximizing the minimum distance to nearby individuals. For example, Arena~\cite{kastner2024arena} introduces a metric that measures the duration the robot spends within the private zone (less than 0.45 m) of any pedestrian, where shorter durations indicate better compliance with social norms. Finally, path length is frequently employed as a proxy for energy efficiency~\cite{kastner2022human,kastner2024arena}. This view is consistent with early comfort-oriented following work such as Sun \textit{et al.}~\cite{sun2016ROBIO}, which combines motion, relative-position, and environmental factors into a comfort utility. In Follow-Bench, we adopt objectively measurable safety- and comfort-related metrics so that different planners can be compared under the same scenario and execution protocol.

It is important to note that proxemic-based quantitative metrics are not universal standards \cite{francis2025principles,honig2018toward}. Preferred following positions are highly context-dependent, varying with the specific RPF task (e.g., eldercare, photography, delivery), the robot's physical form factor, and even the social attributes of the involved humans (e.g., gender, age). Moreover, several studies argue that proxemic zones are better represented as egg-shaped \cite{barnaud2014proxemics}, asymmetrical \cite{cai2022human}, or even dynamic \cite{hayduk1994personal}, rather than fixed circular regions.

In addition to the physical indicators discussed above, user preference serves as a gold standard for comfort assessment. It is typically evaluated through user surveys or interviews, in which human participants experience and assess the interaction in real-world scenarios~\cite{murakami2014destination, jevtic2015comparison, shanee2016influence}. For example, Murakami \textit{et al.}~\cite{murakami2014destination} conducted interviews with 20 participants and asked them to rate the naturalness, perceived safety, and overall satisfaction of the interaction on a scale from one to seven. Since user preference is highly subjective and context-dependent\cite{honig2018toward}, it is recommended to conduct user studies in the specific application context to validate the performance of the methods.

\begin{table}[t]
    \centering
    \caption{Representative evaluation metrics for safety and comfort.}
    \scalebox{0.65}{  
    \begin{tabular}{l|l|c|l}
    \toprule
    \textbf{Category} & \textbf{Metric} & \textbf{Unit} & \textbf{Explanation}  \\
    \midrule
    \midrule
    \multirow{6}{*}{\textbf{Safety}} 
        & \makecell[l]{Avoidance Success Rate$\uparrow$ \\ (ASR)} & \% & Runs with no collision  \\
        & \makecell[l]{Search Success Rate$\uparrow$ \\ (SSR)} & \% & \makecell[l]{Runs with successful search within\\ limited time after target loss} \\
        & SPL$\uparrow$ & - & \makecell[l]{Search Success weighted by \\ path length during search process} \\
        & \makecell[l]{Success Rate (SR)$\uparrow$} & \% & \makecell[l]{Runs with successful search and\\ no collision} \\
        \midrule
        \multirow{11}{*}{\textbf{Comfort}}
        & \makecell[l]{Target Visibility Ratio$\uparrow$ \\ (TVR)} & \% & \makecell[l]{Ratio of target visible time\\ to the total time} \\
        & Path Length$\downarrow$ & [m] & Robot path length \\ 
        & Velocity (avg.) & [m/s] & Robot velocity \\ 
        & Acceleration (avg.) & [m/s$^2$] & Robot acceleration \\ 
        & Movement Jerk$\downarrow$ & [m/s$^3$] & Derivation of acceleration \\ 
        & \makecell[l]{Time Ratio in personal zone$\uparrow$ \\ (TinTPerson)} &\% & \makecell[l]{Time the robot in personal zone\\ of the target / Total time of a trial} \\
        & \makecell[l]{Time Ratio in private zone$\downarrow$ \\ (TinPrivate)} &\% & \makecell[l]{Time the robot in private zone\\ of the surrounding humans\\ / Total time of a trial} \\
        & Angle Deviation$\downarrow$ & [deg] & \makecell[l]{Deviation from the preferred\\ following angle} \\
        & User Preference$\uparrow$ & - & \makecell[l]{Preference score from user survey} \\
    \bottomrule
    \end{tabular}
    }
    \label{tab:metrics}
    \vspace{-0.3cm}
\end{table}

\subsection{RPF-oriented Motion Planning} \label{sec:methods}

\begin{table*}[t]
    \centering
    \caption{Representative reviewed methods. A \Checkmark \ indicates that the method incorporates the corresponding feature, while an \XSolidBrush \ indicates that it does not. \textbf{Obst. Avd.} denotes obstacle avoidance; \textbf{Proxemics} denotes consideration of social space when producing a person-following point/trajectory; \textbf{Active Occ. Avd.} denotes active occlusion avoidance; \textbf{Person Search} denotes re-finding and re-identifying the target after loss; and \textbf{Adap. F-P} denotes adaptive following position.}
    \scalebox{0.8}{  
    \begin{tabular}{l|c|c|c|c|c|c|c|c}
    \toprule
    \textbf{Reference} & \textbf{Year} & \textbf{Global Map} & \textbf{Obst. Avd.} & \textbf{Proxemics} & \textbf{Active Occ. Avd.} & \textbf{Person Search} & \textbf{Adap. F-P} & \textbf{Local planner} \\
    \midrule
    \midrule
    \textbf{Ferrer \textit{et al.}\cite{ferrer2017robot}} &2017 &\Checkmark &\Checkmark &\Checkmark &\XSolidBrush &\XSolidBrush &\Checkmark &SFM \\
    
    \textbf{Goldhoorn \textit{et al.}\cite{goldhoorn2017searching}} &2017 &\Checkmark &\Checkmark &\XSolidBrush &\XSolidBrush &\Checkmark &\XSolidBrush &Traj. Roll Out\cite{gerkey2008planning} \\
    
    \textbf{Montesdeoc \textit{et al.}\cite{bruckschen2020predicting}} &2020 &\XSolidBrush &\XSolidBrush &\Checkmark &\XSolidBrush &\XSolidBrush &\XSolidBrush &PID \\
    
    \textbf{Sekiguchi \textit{et al.}\cite{sekiguchi2021uncertainty}} &2021 &\XSolidBrush &\XSolidBrush &\Checkmark &\XSolidBrush &\XSolidBrush &\Checkmark &MPC \\
    
    \textbf{Dang \textit{et al.}\cite{van2022collision}} &2022 &\XSolidBrush &\Checkmark &\XSolidBrush &\XSolidBrush &\XSolidBrush &\XSolidBrush &DWA \\

    \textbf{Wang \textit{et al.}\cite{wang2024continuous}} &2024 &\XSolidBrush &\Checkmark &\XSolidBrush &\Checkmark &\XSolidBrush &\Checkmark &MPC \\

    \textbf{Leisiazar \textit{et al.}\cite{leisiazar2025adapting}} &2025 &\Checkmark &\Checkmark &\XSolidBrush &\XSolidBrush &\XSolidBrush &\XSolidBrush &RL \\

    \textbf{Ye \textit{et al.}\cite{ye2025rpf}} &2025 &\XSolidBrush &\Checkmark &\XSolidBrush &\XSolidBrush &\Checkmark &\XSolidBrush &MPC \\

    \textbf{Lyu \textit{et al.}\cite{lyu2025robust}}  &2025 &\XSolidBrush &\Checkmark &\XSolidBrush &\XSolidBrush &\XSolidBrush &\XSolidBrush &Traj. Opt. (B-spline) \\
    
    \bottomrule
    \end{tabular}
    }
    \label{tab:methods}
    \vspace{-0.3cm}
\end{table*}

\begin{figure}[t]
    \centering
    \includegraphics[width=\linewidth]{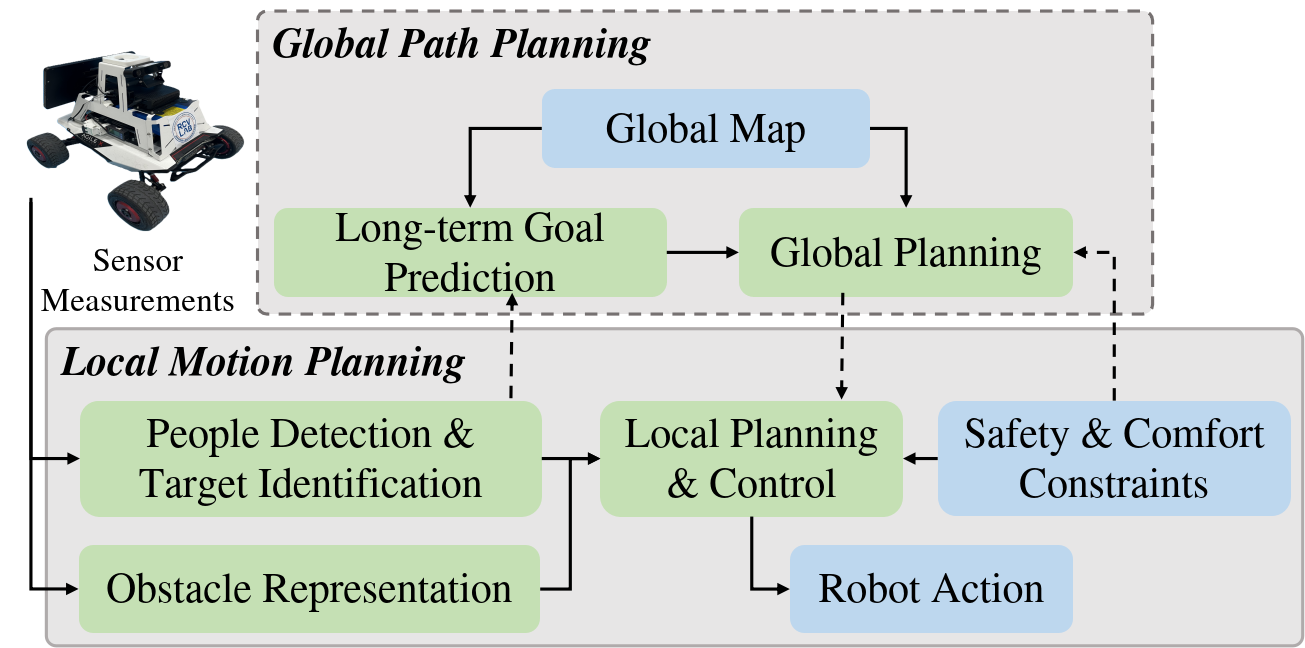}
    \caption{A general motion-planning framework for RPF emphasizing safety and comfort constraints (e.g., proxemics, collision avoidance, target visibility). The global planner uses an estimated long-term goal of the target person to generate a global path (optional in most existing methods), while the local planner is essential for producing local trajectories to follow the target person, ensuring safety and comfort concerns. In this paper, we mainly review and benchmark RPF-oriented local motion planners.}
    \label{fig:methodFramework}
    \vspace{-0.3cm}
\end{figure}

RPF-oriented motion planning must simultaneously account for safety and comfort. It generally follows the standard navigation pipeline—comprising global path planning and local motion planning—while incorporating additional human-aware functions or constraints. A representative framework is illustrated in Fig.~\ref{fig:methodFramework}, which includes both global and local planners.
Global planners typically rely on a prior map \cite{ferrer2014proactive,goldhoorn2017searching,repiso2017line,ferrer2017robot,karunarathne2018model,bayoumi2019speeding} to estimate the target's long-term destination and to compute a global path. These methods often emphasize safety, particularly in re-finding the target after occlusion, in order to maintain continuous visibility \cite{goldhoorn2017searching,bayoumi2019speeding}. However, constructing a reliable prior map is impractical in many real-world scenarios, especially in large-scale or dynamic environments.

Given the dynamic nature of RPF, most approaches therefore focus on local planners, which are responsible for generating safe and socially compliant person-following behaviors. For this reason, our review primarily concentrates on local planning methods. In contrast to previous surveys~\cite{islam2019person,eirale2025human} that mainly discuss collision avoidance and proxemic considerations, we further analyze how recent methods incorporate active occlusion avoidance, recovery after occlusion, and adaptive following position to enhance both safety and comfort. A summary of representative approaches and their attributes is presented in Table~\ref{tab:methods}. These references are selected as representative RPF-oriented planning studies rather than an exhaustive list. We include works that (i)~explicitly involve motion-planning decisions such as following-point/trajectory generation, collision avoidance, proxemic regulation, occlusion handling, or target search, and (ii)~collectively cover diverse planner paradigms, including SFM, trajectory rollout, PID, MPC, DWA, RL, and trajectory optimization. For simplicity, we hereafter refer to RPF-oriented local motion planners as \textbf{RPF planners}. 

Current RPF planners typically navigate to a goal based on the target's most recent position, which has been demonstrated more human-like and socially compliant than replicating the target's historical trajectory~\cite{gockley2007natural}. The most straightforward strategy is to predefine a relative position with respect to the target and use a PID controller to reach and maintain it~\cite{montesdeoca2022person,ye2023robot,ye2024reid,liu2023close}. For comfort consideration, some methods~\cite{montesdeoca2022person,karunarathne2018model} further incorporate proxemic constraints to regulate the robot's accompanying position. These constraints generally specify the desired distance and angle between the robot and the target person. The distance constraint~\cite{montesdeoca2022person} is usually defined as a comfortable following distance within the social space (Fig.~\ref{fig:socialSpace}), while angular preferences are commonly categorized into frontal~\cite{leisiazar2025adapting}, side-by-side~\cite{karunarathne2018model}, and back following~\cite{montesdeoca2022person}. Among these, back following is the most prevalent due to its unobtrusive nature. Frontal following is less common since it requires accurate prediction of the target's future movements while avoiding obstacles, whereas side-by-side following is typically adopted in socially interactive scenarios. 


Beyond the proxemic constraints considered for comfort, collision avoidance remains a fundamental safety requirement. To simultaneously satisfy both proxemic and collision-avoidance constraints, researchers have developed increasingly sophisticated local planning techniques, including dynamic window approach (DWA)\cite{jin2020access,eirale2022human, van2022collision}, SFM\cite{repiso2024adaptive,Ferrer2013sfm}, trajectory-optimization-based methods\cite{wu2026human,lyu2025robust,wang2022geometrically,lin2024safety,ji2022elastic}, model predictive control (MPC)\cite{Park2013Autonomous,sekiguchi2021uncertainty,peng2024dual,peng2023iros,yuan2025taes}, and reinforcement learning (RL)\cite{dewantara2016generation,leisiazar2023mcts,leisiazar2025adapting,kastner2022human}. For example, Dang \textit{et al.}\cite{van2022collision} enhance the DWA by incorporating the desired person-following direction into the velocity sampling process, along with collision constraints and smoothness considerations. SFM methods model the robot's movement using attractive forces toward the target (or their destination), repulsive forces to avoid obstacles, and social forces to respect interpersonal distance from pedestrians\cite{Ferrer2013sfm}. Furthermore, trajectory-optimization-based approaches formulate human-following and obstacle avoidance as a unified mathematical problem using parameterized models, such as time elastic bands (TEB)\cite{wu2026human}, B-spline curves\cite{lin2024safety,lyu2025robust}, or minimum control effort (MINCO) representations\cite{wang2022geometrically,ji2022elastic}. By explicitly minimizing tracking and control costs under strict spatial boundaries, these methods guarantee collision-free and kinematically feasible trajectories in complex environments.


Among these methods, MPC-based approaches have recently gained prominence due to their ability to optimize robot actions over a future time horizon while considering dynamic constraints. This enables proactive adjustments for maintaining safe and comfortable following behavior. MPC formulations can naturally integrate explicit constraints (e.g., speed limits, safety margins), and advances in solvers such as CasADi\cite{andersson2019casadi} make real-time optimization feasible. For instance, Sekiguchi \textit{et al.}\cite{sekiguchi2021uncertainty} introduce a nonlinear-MPC-based framework that predicts the target's future trajectory and integrates uncertainty as constraints, enabling proactive following within personal space without obstructing the target. In addition, they follow Hu \textit{et al.}\cite{hu2013design} to adaptively select the person-following position based on symmetry for avoiding obstruction to the target. 
In contrast, RL-based methods\cite{dewantara2016generation,leisiazar2023mcts,leisiazar2025adapting,kastner2022human} learn implicit policies from data using reward-driven training in simulation environments. These methods avoid manually defining constraints but are often limited by sim-to-real transfer issues and lack validation in large-scale real-world deployments.

Beyond collision avoidance, occlusion avoidance represents another critical dimension of RPF safety. Occlusions caused by static structures (e.g., walls or furniture) or dynamic obstacles (e.g., surrounding pedestrians) may lead to temporary loss of the target, resulting in discontinuous or uncomfortable following behaviors. Existing approaches typically address this challenge via either active planning\cite{Hoeller2007Accompanying, yao2021iros, wang2024continuous} or person search\cite{ye2025rpf,algabri2021target,lee2018icra,kim2018architecture,chen2017integrating} strategies. 

Active planning anticipates occlusion events and proactively adjusts the robot's position to maintain line-of-sight. For example, Yao \textit{et al.}\cite{yao2021iros} address corner occlusion by predicting the target's turn and switching from back-following to side-following via a PD controller. While effective in specific cases, this method lacks flexibility to diverse occlusion scenarios. Wang \textit{et al.}\cite{wang2024continuous} propose a more flexible MPC-based approach that introduces human-human and human-obstacle occlusion constraints. However, while maximizing visibility, it may compromise proxemic comfort, resulting in overly distant following behaviors.

Person search provides an alternative strategy when the target is lost. The most basic solution is to navigate the robot to the target's last known position\cite{chen2017integrating}. More advanced approaches leverage trajectory prediction techniques---such as linear regression\cite{algabri2021target}, Bayesian regression\cite{lee2018icra}, and SVM-based models\cite{kim2018architecture}---to estimate the target's future location. Despite improvements, these methods still struggle with occlusion in complex environments (e.g., U-turn corners). To address this, Ye \textit{et al.}\cite{ye2025rpf} represent the predicted trajectory probabilistically and construct a belief-guided field to incrementally update the likelihood of the target's presence. Furthermore, they introduce a fluid-following field for dynamically tracking occluders and an observation-based potential field to overtake them based on their motion patterns, thus addressing dynamic occlusion scenarios more robustly.

\section{Follow-Bench} \label{sec:bench}

\subsection{Overview}
To comprehensively evaluate the local motion planning capabilities of various RPF planners within a unified framework, we present Follow-Bench. This benchmark provides a diverse set of scenarios (Sec.~\ref{sec:benchScenarios}), covering different target trajectory patterns, crowd dynamics, and environmental layouts. In each scenario, the robot is required to follow a designated individual while avoiding obstacles and ensuring the safety and comfort of both the target and surrounding pedestrians. These scenarios are selected based on the studies reviewed in Sec.~\ref{sec:scenarios}, and represent typical yet challenging situations commonly encountered in RPF tasks. In addition, Follow-Bench includes a comprehensive reimplementation of representative RPF planners (Sec.~\ref{sec:benchBaselines}) and a set of evaluation metrics (Sec.~\ref{sec:benchMetrics}) with an emphasis on safety and comfort. The benchmark is designed to be highly extensible, allowing seamless integration of new maps, obstacles, agents, as well as customized agent trajectories and motion attributes. It should be noted that the benchmark evaluation considers only physical indicators. The real-world experiments with MPC and MPC w/ Traj. provide qualitative participant feedback on following behavior; this feedback does not constitute a validated assessment of perceived comfort.

Follow-Bench extends a Python-based 2D navigation simulator~\cite{han2023rda} by incorporating realistic interaction simulation and human visibility checking. Several of its design elements are inspired by existing social navigation simulators~\cite{chen2019crowd,mun2023icra}. Here, we adopt a lightweight simulator instead of high-fidelity simulators for its simplicity and compatibility with CPU-only execution. Moreover, it is worth noting that the interaction mechanisms in many high-fidelity simulators~\cite{tsoi2022sean2,kastner2024arena} are also fundamentally 2D-based: they first generate socially-aware 2D human trajectories and subsequently augment them with 3D animations for enhanced visualization.

As summarized in Table~\ref{tab:simComparison}, since few existing platforms directly target RPF motion-planning evaluation, we compare Follow-Bench with adjacent human-aware navigation and tracking benchmarks that share key simulation ingredients with RPF, such as pedestrian modeling, collision avoidance, and social interaction. Existing simulation platforms for human-aware robot navigation---such as CrowdNav~\cite{chen2019crowd}, SEAN~2.0~\cite{tsoi2022sean2}, and Arena~3.0~\cite{kastner2024arena}---were designed primarily for point-to-point social navigation and do not provide RPF-specific scenarios, planners, or metrics. The most closely related platform, EVT-Bench~\cite{wang2025trackvla}, targets end-to-end embodied visual tracking and evaluates vision-language-action models rather than motion planners. In contrast, Follow-Bench is, to our knowledge, the first benchmark that unifies RPF-specific scenario design, a comprehensive suite of re-implemented RPF planners spanning reactive, predictive, optimization-based, and learning-based paradigms, and dedicated RPF evaluation metrics within a single lightweight framework. Its CPU-only, state-based design deliberately decouples motion planning from perception, enabling controlled, large-scale statistical comparison of planning algorithms under systematically varied crowd dynamics, environmental layouts, and following configurations.

\begin{table*}[t]
\centering
\caption[Comparison of Follow-Bench with simulation platforms.]{Comparison of Follow-Bench with existing benchmarks for human-aware robot navigation and person following. \textit{Following position}: supported person-following geometries (back-following vs.\ side-following). \textit{RPF Scenarios}: whether the platform provides dedicated scenario suites (target trajectories, crowd dynamics, environmental layouts) for robot person-following evaluation. \textit{RPF Planners}: number of re-implemented RPF-specific motion planners included. \textit{RPF Metrics}: whether RPF-specific evaluation metrics (e.g., target proximity, visibility ratio, search success rate) are provided beyond standard navigation metrics.}
\label{tab:simComparison}
\begingroup
\footnotesize
\setlength{\tabcolsep}{2.5pt}
\renewcommand{\arraystretch}{0.95}
\resizebox{1.0\textwidth}{!}{%
\begin{tabular}{@{}lccccccccc@{}}
\toprule
\textbf{Platform}
  & \textbf{Dimension}
  & \textbf{Rendering}
  & \textbf{Compute}
  & \makecell{\textbf{Human}\\\textbf{model}}
  & \makecell{\textbf{Primary}\\\textbf{task}}
  & \makecell{\textbf{Following}\\\textbf{position}}
  & \makecell{\textbf{RPF}\\\textbf{scenarios}}
  & \makecell{\textbf{RPF}\\\textbf{planners}}
  & \makecell{\textbf{RPF}\\\textbf{metrics}} \\
\midrule
\midrule
CrowdNav~\cite{chen2019crowd} & 2D & None & CPU & ORCA & Social navigation & $\times$ & $\times$ & 0 & $\times$ \\
SEAN 2.0~\cite{tsoi2022sean2} & 3D & Unity & GPU & ORCA (2D core) & Social navigation & $\times$ & $\times$ & 0 & $\times$ \\
Arena 3.0~\cite{kastner2024arena} & 2D/3D & Flatland/Unity & CPU/GPU & ORCA/SFM (2D core) & Social navigation & $\times$ & $\times$ & 0 & $\times$ \\
EVT-Bench~\cite{wang2025trackvla} & 3D & Photorealistic & GPU & SMPL-X + ORCA & Embodied visual tracking & Back & Partial & 0 & Partial \\
\midrule
\textbf{Follow-Bench (Ours)} & 2D & None & CPU & ORCA / SFM & RPF motion planning & \makecell{Back/Side} & \Checkmark & 8 & \Checkmark \\
\bottomrule
\end{tabular}%
}%
\endgroup
\end{table*}


\subsection{Scenarios} \label{sec:benchScenarios}

\begin{figure}[t]
    \centering
    \includegraphics[width=\linewidth]{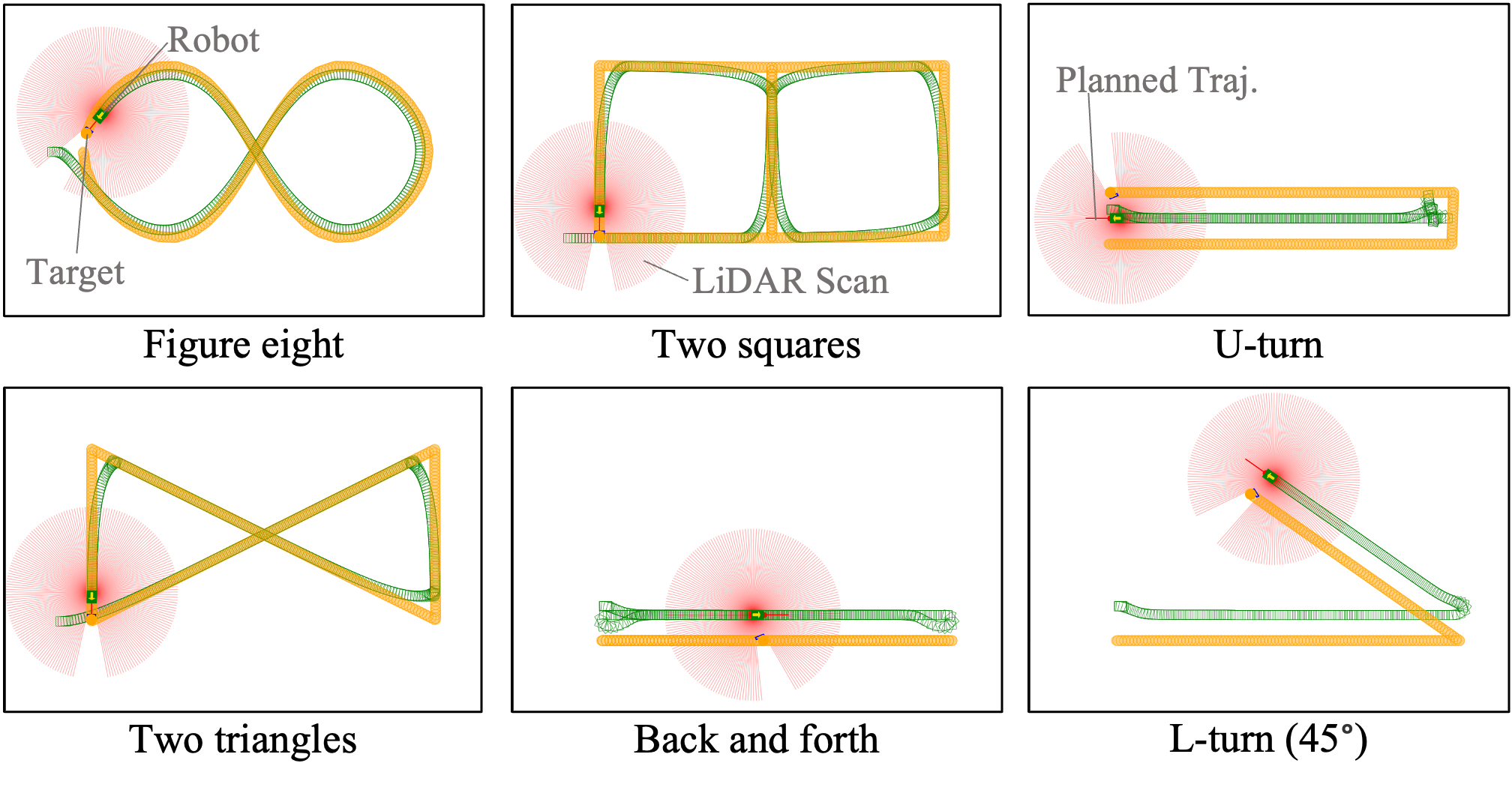}
    \caption{Scenarios of target trajectories. The scenarios include two-triangle, two-square, U-turn, figure-eight (`8'), L-turns with varying angles (30$^{\circ}$, 45$^{\circ}$, and 60$^{\circ}$), and back-and-forth walking. These scenarios are used to independently evaluate planners' agility and motion smoothness. The robot is initialized at its preferred following position relative to the target including back-following and side-following. The {\color[HTML]{287c1d} green rectangle} represents the robot, the {\color[HTML]{E87B1C} orange circle} represents the target, and the {\color[HTML]{EF1900} short red curve} denotes the desired following trajectory generated by the planner. The {\color[HTML]{F44336} long red lines around the robot} are simulated 2D LiDAR scans.}
    \label{fig:benchTargetScenarios}
\end{figure}

\begin{figure}[t]
    \centering
    \includegraphics[width=0.90\linewidth]{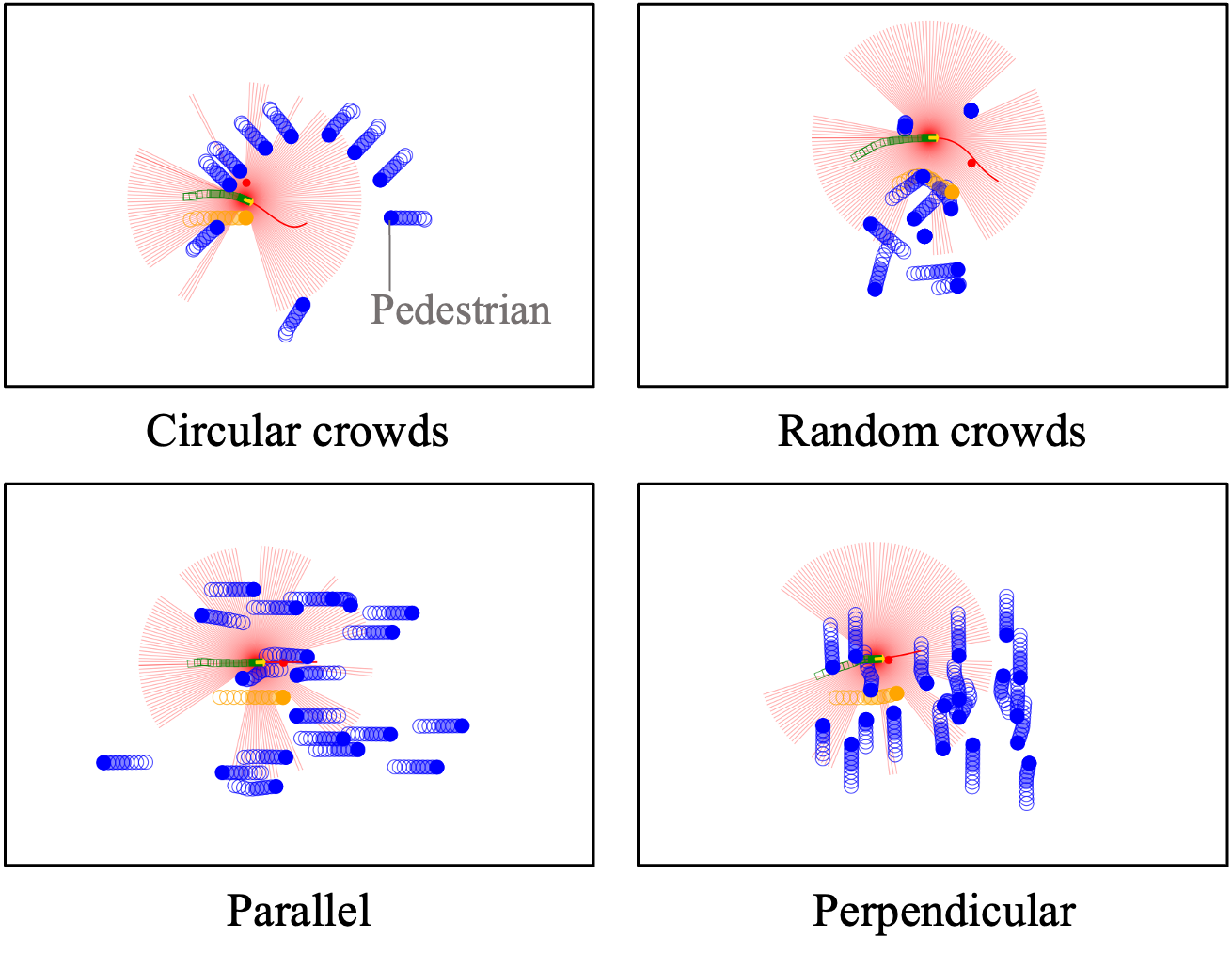}
    \caption{Scenarios of dynamic crowds. {\color[HTML]{0B00F8} Blue circles} denote dynamic pedestrians, and the orange circle denotes the target. In each scenario, the number of pedestrians is varied from 5 to 30 to create different crowd densities, enabling independent evaluation of planners' collision-avoidance and social-spacing capabilities in human-populated environments. The scenarios include circular crowds (flow around a central area, e.g., courtyards), random crowds (e.g., plazas or shopping malls), parallel crossing (e.g., pedestrians in crosswalks), and perpendicular crossing (e.g., intersections in open spaces).}
    \label{fig:benchDynamicCrowds}
\end{figure}

\begin{figure}[t]
    \centering
    \includegraphics[width=0.90\linewidth]{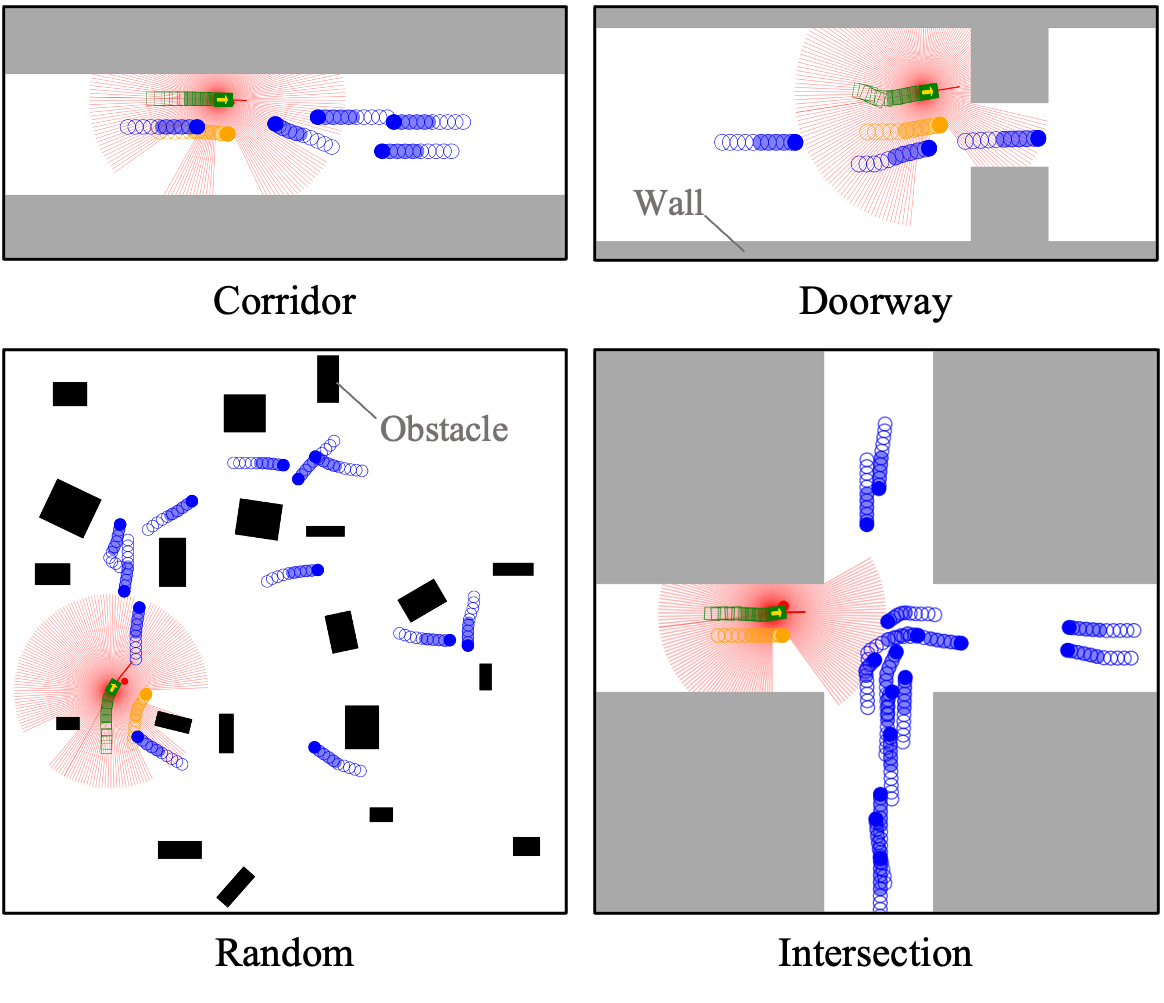}
    \caption{Scenarios of layouts with structured pedestrian flows. The scenarios include corridors, doorways, intersections, and cluttered spaces populated with dynamic pedestrians. For each scenario, we independently vary passage width (or obstacle density) and pedestrian number, enabling evaluation of planners' capabilities in navigating narrow or cluttered spaces while maintaining appropriate social distance from both the target and surrounding pedestrians.}
    \label{fig:benchTopographies}
    \vspace{-0.3cm}
\end{figure}

\begin{table*}[t]
\centering
\caption{Evaluation settings for planner performance under varying levels of complexity. The number of humans (H), occupancy conditions (O), and following configurations (F) are systematically varied across environmental layout and crowd-dynamics scenarios to characterize complexity factors such as collision risk, occlusion risk, and the maintenance of socially appropriate distances. Dynamic scenarios include circular, random, parallel-crossing, and perpendicular-crossing crowds. Each configuration is evaluated over 100 runs with distinct random seeds to capture the stochastic nature of pedestrian motion. Unless otherwise specified, all scenarios are tested under both back- and side-following configurations at a fixed following distance of 1.5 m.}
\label{tab:benchScenarios}
\scalebox{0.83}{
\begin{tabular}{llcc}
\toprule
\textbf{Experiment} & \textbf{Variable Type} & \textbf{Values} & \textbf{Fixed Conditions} \\
\midrule
\midrule
\textbf{Target Trajectory} & - & - & - \\
\midrule
\textbf{Dynamic (H)} & Human number & \{5, 10, 15, 20, 25, 30\} & -\\
\midrule
\multirow{4}{*}{\textbf{Layout (H)}} 
& Corridor humans & \{4, 12, 20, 28\} & 5.6 m width\\
& Intersection humans & \{2, 10, 18, 26\} & 4.8 m width \\
& Doorway humans & \{2, 6, 10, 14\} & 2.8 m width \\
& Cluttered space humans & \{10, 20, 30, 40\} & 30 obstacles \\
\midrule
\multirow{4}{*}{\textbf{Layout (O)}} 
& Corridor width & \{6.8, 6.2, 5.6, 5.0\} m & 20 humans \\
& Intersection width & \{6.0, 5.4, 4.8, 4.2\} m & 18 humans \\
& Doorway width & \{3.4, 3.1, 2.8, 2.5\} m & 10 humans \\
& Cluttered space obstacles & \{10, 20, 30, 40\} & 30 humans \\
\midrule
\textbf{Dynamic (F)} & Following configuration &Back / Side at \{1.0, 1.5, 2.0, 2.5\} m & 20 humans \\
\midrule
\multirow{4}{*}{\textbf{Layout (F)}}
& \multirow{4}{*}{Following configuration}
& \multirow{4}{*}{Back / Side at \{1.0, 1.5, 2.0, 2.5\} m}
& Corridor: 20 humans; 5.6 m width\\
& & & Intersection: 18 humans; 4.8 m width\\
& & & Doorway: 10 humans; 2.8 m width\\
& & & Cluttered space: 30 humans; 30 obstacles \\
\bottomrule
\end{tabular}
}
\end{table*}


Follow-Bench encompasses a diverse suite of 16 scenario types, systematically designed to capture variations in target trajectory patterns, crowd dynamics, and environmental layouts---drawing upon representative RPF scenarios summarized in Table~\ref{tab:scenarios}.
\begin{itemize}
    \item \textbf{Target trajectory patterns} include two-triangle\cite{sekiguchi2021uncertainty, peng2024dual}, two-square\cite{sekiguchi2021uncertainty, Kuderer2014Approach, Park2013Autonomous, hu2014tie, yao2021iros, montesdeoca2022person}, U-turn\cite{sekiguchi2021uncertainty, bayoumi2016learning, leisiazar2025adapting, nikdel2020lbgp, eirale2022human, karunarathne2018model}, figure-eight (`8') patterns\cite{sekiguchi2021uncertainty, bayoumi2016learning}, L-turns with varying angles (30$^{\circ}$, 45$^{\circ}$, and 60$^{\circ}$)\cite{sekiguchi2021uncertainty}, and back-and-forth walking\cite{sekiguchi2021uncertainty}. Representative examples are shown in Fig.~\ref{fig:benchTargetScenarios}. These scenarios involve abundant turning behaviors, designed to independently evaluate an RPF planner's agility, trajectory smoothness, and adaptability to target motion changes.
    \item \textbf{Crowd dynamics} cover parallel crossing\cite{gross2017roreas, kollmitz2015Time,liu2019robot,repiso2017line}, perpendicular crossing\cite{gross2017roreas, liu2019robot, repiso2017line}, circular crossing\cite{wang2022metrics}, and random crowds\cite{repisoadaptive2020,repiso2024adaptive}. Due to human-human interactions, pedestrian motions in these scenarios are highly dynamic and uncertain, often resulting in potential collisions and occlusions. They are therefore employed to evaluate the planner's capability in collision avoidance, occlusion recovery, and maintaining socially appropriate interpersonal distances in human-populated environments. Representative examples are shown in Fig.~\ref{fig:benchDynamicCrowds}.
    \item \textbf{Environmental layouts} comprise corridors\cite{Park2013Autonomous, eirale2022human, montesdeoca2022person, gross2017roreas, kollmitz2015Time}, doorways\cite{repisoadaptive2020, xiao2023Learning}, intersections\cite{yuan2018laser, kollmitz2015Time}, and cluttered spaces populated with dynamic pedestrians\cite{Kuderer2014Approach,xiao2023Learning}. These scenarios combine both dynamic and static obstacles, posing higher-level challenges where the robot must traverse constrained environments while ensuring collision-free motion and socially comfortable distances from both the target and nearby pedestrians. Representative cases are depicted in Fig.~\ref{fig:benchTopographies}.
\end{itemize}

Building upon the aforementioned scenarios, Follow-Bench further supports evaluation across different \textbf{following angles}, including back-following~\cite{ye2025rpf} and side-following~\cite{repiso2017line}, as illustrated in Fig.~\ref{fig:benchTargetScenarios}. To systematically quantify the complexity of RPF tasks, we adopt the characterization framework proposed by Stratton \textit{et al.}~\cite{stratton2024characterizing}, evaluating planner performance under varying \textbf{numbers of humans} and \textbf{environmental occupancy levels} (e.g., passageway widths and obstacle counts). Moreover, we investigate the impact of different \textbf{following distances} to analyze each planner's adaptability to diverse social preferences and proxemic constraints. These configuration variants within each scenario enable systematic assessment of a planner's adaptability to varying levels of complexity---including collision and occlusion risks, as well as the maintenance of socially appropriate interpersonal distances. The detailed configuration specifications are summarized in Table~\ref{tab:benchScenarios}.

In all scenarios, target and pedestrian trajectories are generated using an A* global planner over predefined waypoints, while local interactions and obstacle avoidance are governed by either ORCA\cite{van2011reciprocal} or SFM\cite{helbing1995social}. Further implementation details of the simulation environment are provided in the Appendix.

\subsection{RPF planners} \label{sec:benchBaselines}
We evaluate eight Follow-Bench implementations derived from representative RPF planners: \textbf{SFM}~\cite{ferrer2017robot}, \textbf{BSO-HFC}~\cite{lyu2025robust}, \textbf{MPC} (built on the RDA solver~\cite{han2023rda}), \textbf{DWA}~\cite{van2022collision}, and an \textbf{RL-based} planner~\cite{leisiazar2025adapting}, along with three composite variants (\textbf{MPC w/ Traj.}, \textbf{MPC w/ DS.}, and \textbf{DWA w/ Traj.}). Because the original systems rely on heterogeneous robot platforms, perception pipelines, map assumptions, and control interfaces, directly executing their full control stacks would confound motion planning with system-level differences. We therefore retain each method's core planning mechanism while standardizing the robot model, observation protocol, simulation timestep, and velocity interface. The Appendix details the source, retained core mechanism, and adaptations of each implementation.

To establish a unified RPF task specification, all implementations share a default 1.5\,m following distance, adaptive following-position generation~\cite{sekiguchi2021uncertainty,hu2013design}, simulator-level visibility checking, and a Kalman-filter-based fallback search module after target loss~\cite{algabri2021target,lee2018robust,do2017reliable}. In addition, we construct three composite variants within Follow-Bench: \textbf{MPC w/ Traj.} augments RDA with target-trajectory tracking and uncertainty-triggered braking~\cite{sekiguchi2021uncertainty}; \textbf{MPC w/ DS.} integrates the dynamic search field of~\cite{ye2025rpf} for post-occlusion recovery; and \textbf{DWA w/ Traj.} extends DWA with multi-step trajectory tracking. The \textbf{RL-based} implementation preserves the MCTS structure of~\cite{leisiazar2025adapting} while adding back- and side-following formulations and newly trained value networks.

All implementations focus on map-independent local planning and use a common observation protocol: ground-truth states of visible humans and 2D LiDAR scans for static obstacles, decoupling planning from perception errors. Although Follow-Bench extends the lightweight 2D simulator of~\cite{han2023rda}, all methods share paired scenario instances, differential-drive kinematics, $0.1$\,s control steps, geometric collision checking, and evaluation metrics. The simulator's state updates and collision checks are independent of RDA's optimization formulation. Follow-Bench adds interactive pedestrian flows, mixed static--dynamic layouts, and proxemic following objectives to evaluate RPF behavior. At each step, the shared observations are converted into each planner's native obstacle representation (polygonal clusters for MPC, cluster centers for SFM, point sets for DWA, and local occupancy grids for BSO-HFC and RL-based). For hyper-parameter selection, each baseline was initialized with reference to its cited paper and available implementation, calibrated during preliminary development runs on representative trajectory, crowd, and layout conditions, and then globally frozen before the final 100-trial batch evaluation without per-scenario re-tuning (see the hyper-parameter selection protocol and final planner configurations in the Appendix). Because both the shared wrapper and method-specific adaptations affect visibility and recovery, the reported results characterize the complete Follow-Bench implementations rather than the unmodified original systems.

\subsection{Evaluation Protocol and Metrics} \label{sec:benchMetrics}
Using the re-implemented RPF planners described above, we evaluate their performance across the benchmark scenarios introduced in Sec.~\ref{sec:benchScenarios}. The simulation timestep is fixed at 0.1 s. For each scenario, we first determine the minimum number of simulation steps, $N_{\text{min}}$, required for the target person to complete their trajectory in the absence of the robot, accounting for natural interactions with other pedestrians. Each scenario is then executed with the robot included for a total of $N_{\text{max}} = 2N_{\text{min}}$ steps, ensuring that the target has sufficient time to complete the trajectory even if temporary interruptions occur.

For each scenario, we conduct 100 independent trials. Across trials, the trajectories of surrounding pedestrians and the placement of static obstacles (when applicable) are randomized to introduce behavioral and environmental diversity. The robot is always initialized at its preferred following position relative to the target. During each trial, we compute the evaluation metrics summarized in Table~\ref{tab:metrics}. A search-success event is recorded when the robot successfully re-identifies the target within $0.5N_{\text{min}}$ steps after each loss of visual contact. A successful trial is defined as one in which no collisions occur and the robot resumes tracking of the target within $0.5N_{\text{min}}$ steps before the simulation terminates at $N_{\text{max}}$ steps.

For clarity, we formalize the primary metrics computed from the executed robot trajectory. Let $N = N_{\max}$ denote the maximum number of simulation steps per trial, $\Delta t = 0.1$\,s the fixed simulation timestep, $\mathbf{v}_k$ the robot's executed velocity at step $k$, $d_k = \|\mathbf{p}^{R}_k - \mathbf{p}^{T}_k\| - (r_R + r_T)$ the robot-to-target surface-to-surface distance (where $\mathbf{p}^{R}_k, \mathbf{p}^{T}_k$ are the robot and target positions and $r_R, r_T$ are their radii), and $M$ the number of trials per scenario. The discrete acceleration and jerk are computed as $\mathbf{a}_k = (\mathbf{v}_k - \mathbf{v}_{k-1})/\Delta t$ and $\mathbf{j}_k = (\mathbf{a}_k - \mathbf{a}_{k-1})/\Delta t = (\mathbf{v}_k - 2\mathbf{v}_{k-1} + \mathbf{v}_{k-2})/\Delta t^{2}$, respectively. The key metrics are defined as:
\begin{equation}
\mathrm{SR} = \frac{1}{M}\sum_{i=1}^{M} \mathbbm{1}[\text{trial }i\text{ succeeds}],
\end{equation}
\begin{equation}
\mathrm{TVR} = \frac{1}{N}\sum_{k=1}^{N} \mathbbm{1}[\text{target visible at step }k],
\end{equation}
\begin{equation}
\mathrm{TinTPerson} = \frac{1}{N}\sum_{k=1}^{N} \mathbbm{1}[d_k \in (0.46, 1.2)\,\text{m}],
\end{equation}
\begin{equation}
\mathrm{Jerk} = \frac{1}{N}\sum_{k=1}^{N} \|\mathbf{j}_k\|,
\end{equation}
where $\mathbbm{1}[\cdot]$ denotes the indicator function, and boundary jerk samples ($k<2$) are set to zero. Importantly, $\mathbf{v}_k$ denotes the robot's \emph{executed} velocity, sampled by the simulator at a fixed rate of $1/\Delta t = 10$\,Hz. Since Follow-Bench adopts a synchronous simulation protocol---where each planner is invoked exactly once per simulation step and the simulated clock advances only after the planner returns a velocity command---the executed velocity sequence is uniformly sampled at interval $\Delta t$ for all planners, regardless of their wall-clock planning time. Consequently, the Jerk metric is computed on a common and frequency-consistent time basis by construction, ensuring that differences in Jerk across planners reflect genuine differences in motion smoothness rather than artifacts of inconsistent sampling granularity. This per-step synchronous execution scheme is also widely adopted in established robot navigation and human-aware benchmarks, such as CrowdNav~\cite{chen2019crowd} and EVT-Bench~\cite{wang2025trackvla}. Our benchmark follows the same convention. To nevertheless provide a transparent comparison of computational cost, we additionally report the measured wall-clock planning time of each evaluated planner separately in Table~\ref{tab:planTime}.

\section{Analysis of RPF planners} \label{sec:simResults}
\begin{figure*}[t]
    \centering
    \includegraphics[width=\linewidth]{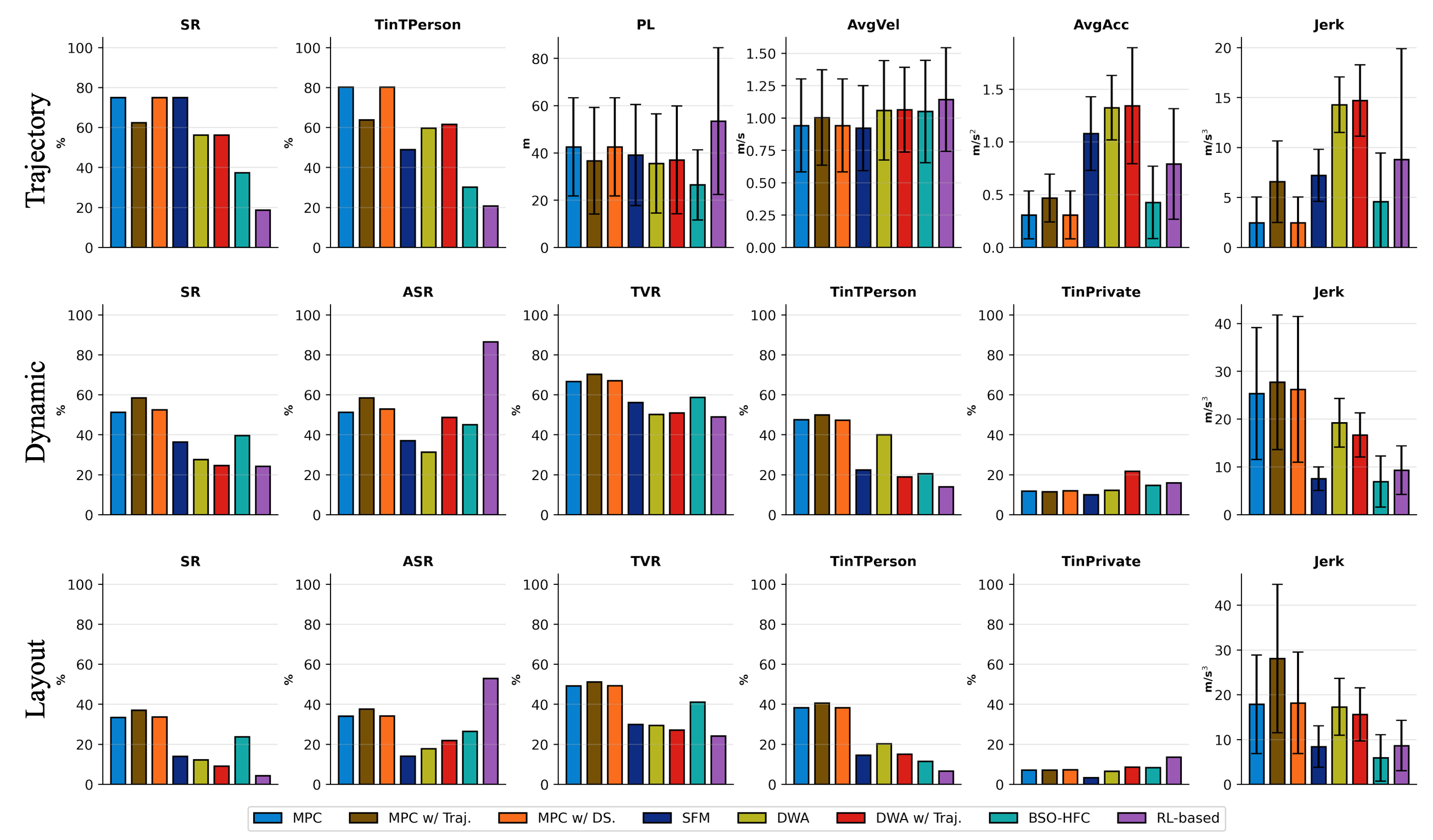}
    \caption{Average Performance on all tested scenarios including various target trajectory patterns, crowd dynamics and environmental layouts from top to down. Evaluated metrics include success rate (\textbf{SR$\uparrow$}) that a successful person-following represents no collision and the target is visible at the end of simulation, collision avoidance success rate (\textbf{ASR$\uparrow$}) target visibility ratio (\textbf{TVR$\uparrow$}), time ratio in target personal zone (\textbf{TinTPerson$\uparrow$}), time ratio in other pedestrians' private zone (\textbf{TinPrivate$\downarrow$}), path length (\textbf{PL$\downarrow$}), average velocity (\textbf{AvgVel}), average acceleration (\textbf{AvgAcc}) and \textbf{Jerk}. Since the target is always visible in scenarios of Trajectories, we do not show the metrics related to target visibility in the first row.}
    \label{fig:expAverage}
\end{figure*}

Based on the proposed benchmark, we conduct a series of experiments to evaluate the performance of re-implemented RPF planners. The experiments are designed to analyze how these planners perform under all scenarios (Sec.~\ref{sec:expAverage}), different person-following distances and angles, human densities and environmental occupancy conditions (Sec.~\ref{sec:expDensity}). Here, we simulate human-object and human-human interaction based on ORCA. For simplicity, we show part of the evaluation metrics here, detailed metric evaluation can be found in the released code. In addition, Appendix includes complete evaluation results of all experiments in each scenario.

\begin{figure}[t]
    \centering
    \includegraphics[width=\linewidth]{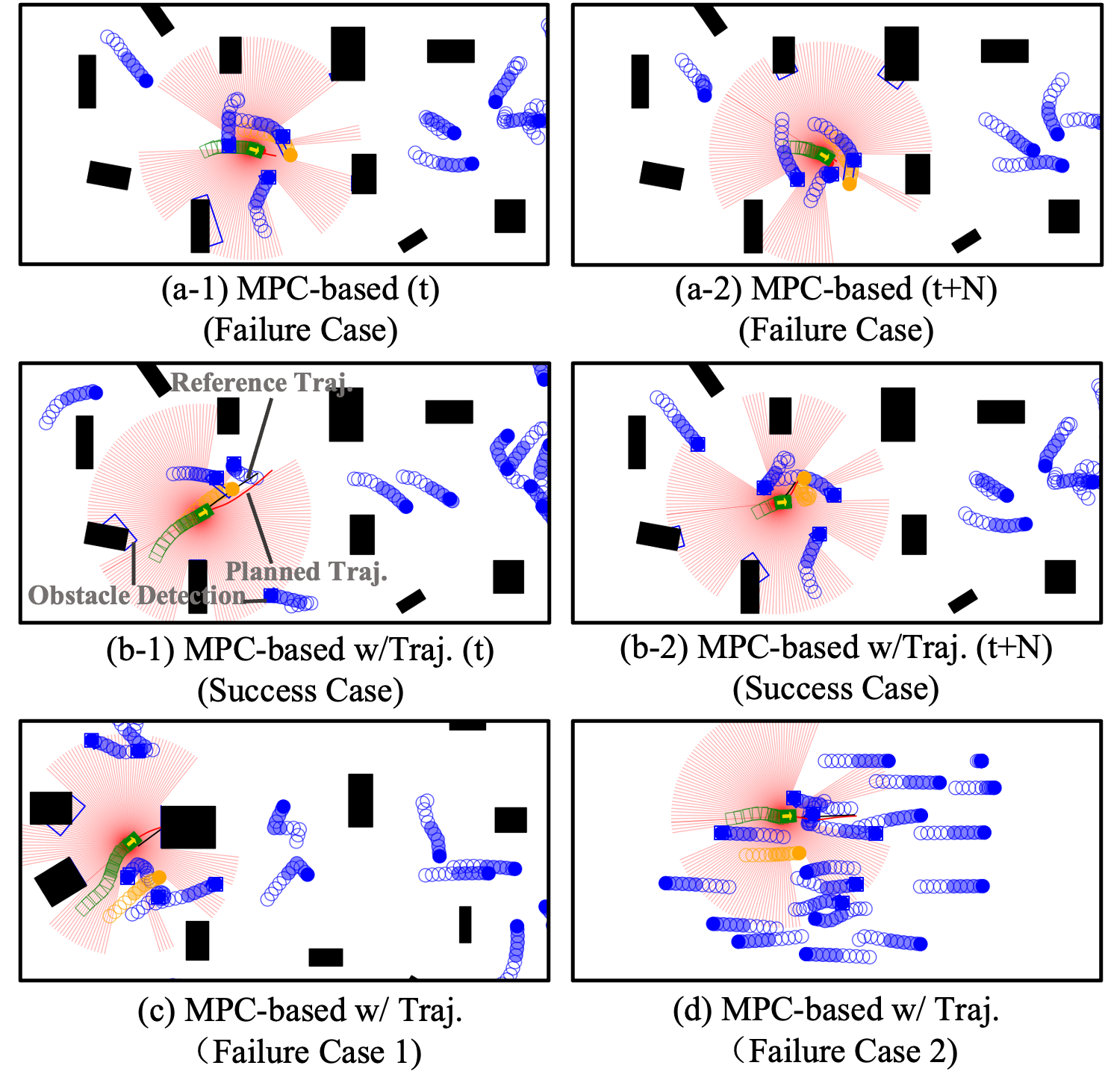}
    \caption{Visualization of baseline performance. Within robot's LiDAR scan region, \textcolor{blue}{blue polygons} denote obstacle detection results including static obstacles and dynamic pedestrians. Red trajectory is the planned following trajectory of the planners. \textbf{(a-1)} and \textbf{(a-2)} depict a failure case of MPC-based planners in a cluttered environment under the back-following configuration. The robot persistently tracks the designated following point, which leaves insufficient maneuvering space and ultimately results in a collision with pedestrians. \textbf{(b-1)} and \textbf{(b-2)} illustrate a success case of MPC w/ Traj. in the same scenario. By following a reference trajectory (\textbf{black curve}), the robot adapts its motion more effectively, enabling timely avoidance of pedestrians and mitigating potential collisions. \textbf{(c)} and \textbf{(d)} show failure cases of MPC-based w/ Traj. in the side-following configuration, where the robot collides with a static obstacle and a pedestrian, respectively. These failures arise from the inherent conflict between safety requirements and the comfort-oriented positioning of side-following.}
    \label{fig:expSimCases}
\end{figure}

\begin{figure*}[t]
    \centering
    \includegraphics[width=\linewidth]{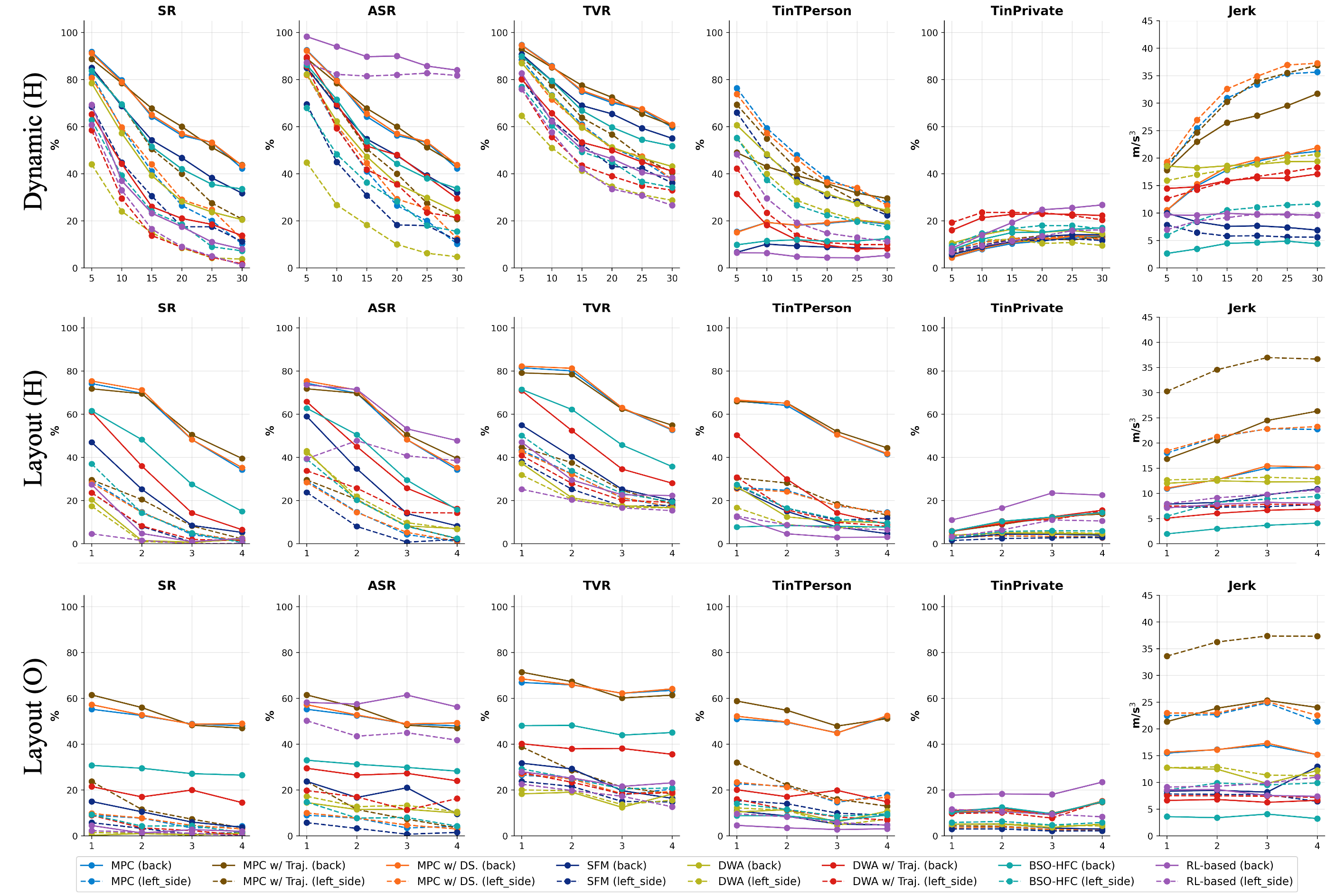}
    \caption{Performance under varying human densities and following angles. In \textbf{Dynamic (H)}, the number of humans ranges from 5 to 30. For \textbf{Layout (H)}, the number of humans varies across scenarios due to differing spatial constraints; to ensure comparability, we average the performance of all scenarios at the same number level, yielding four aggregated levels (1-4). In addition, \textbf{Layout (O)} captures varying occupancy conditions, represented by four levels (1-4) of passageway widths or obstacle numbers.}
    \label{fig:expHumanOccupancy}
\end{figure*}

\begin{figure*}[t]
    \centering
    \includegraphics[width=\linewidth]{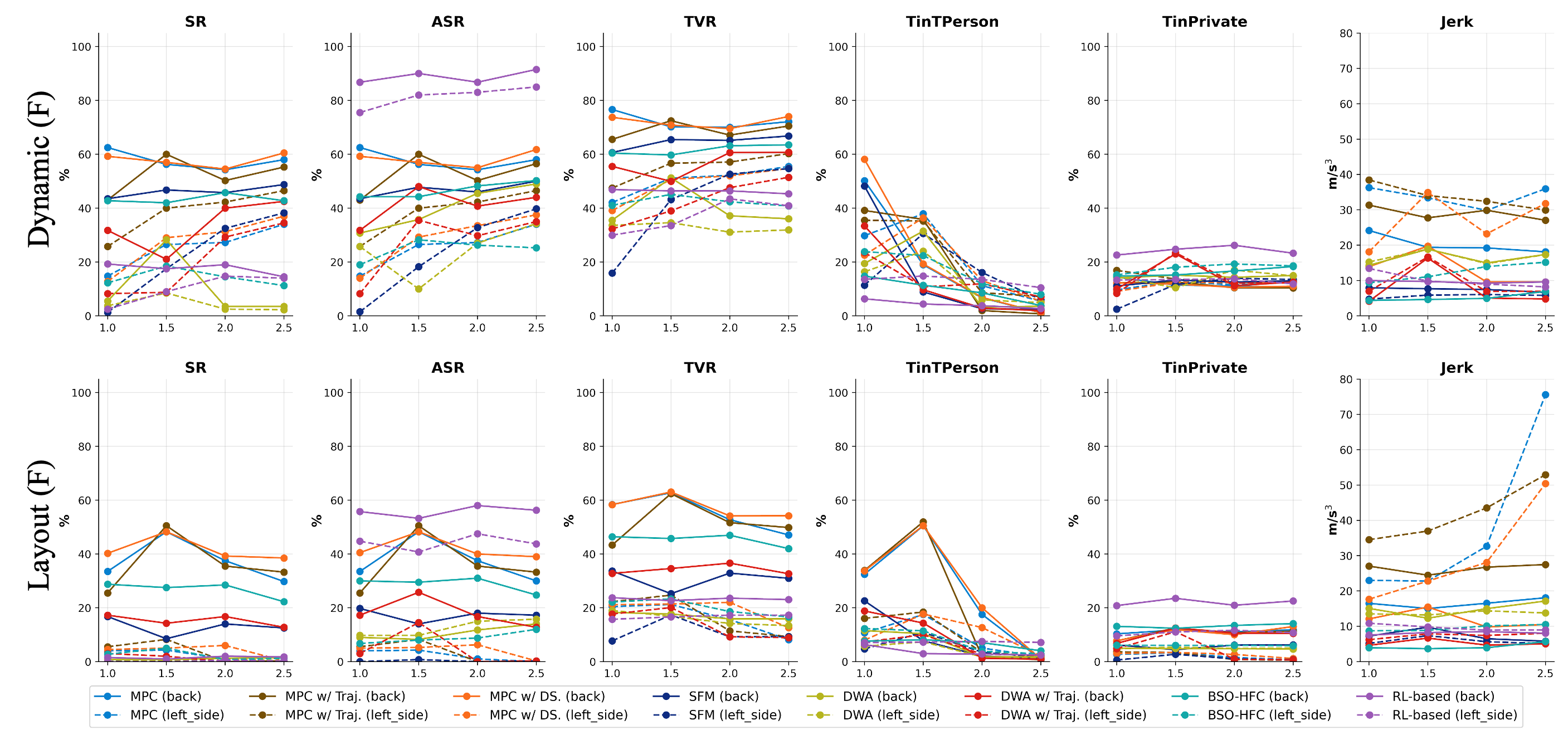}
    \caption{Performance under varying \textbf{following distances (F)}. The following distance varies from 1.0 m to 2.5 m to assess its impact on planner performance.}
    \label{fig:expFollowDist}
\end{figure*}

\subsection{Performance under Different Scenarios} \label{sec:expAverage}
MPC-based planners (MPC, MPC w/ Traj., and MPC w/ DS.) achieve higher observed aggregate SRs than the other methods (SFM, DWA, DWA w/ Traj., BSO-HFC, and RL-based) in the \textit{Dynamic} and \textit{Layout} categories (Fig.~\ref{fig:expAverage}). Their receding-horizon optimization under dynamic constraints and smooth biconvex reformulation of collision constraints~\cite{han2023rda} can enable more adaptive and anticipatory responses to dynamic environments. Such capability is particularly beneficial in person-following scenarios with continuously changing target positions, moving pedestrians, and cluttered or narrow spaces. The higher observed aggregate TinTPerson and TVR of MPC-based planners are also consistent with their ability to respond to target motion and remain near the target. These aggregate results are descriptive averages across the evaluated configurations and both following directions.

Among these, MPC w/ Traj. achieves the highest observed aggregate SR, ASR, and TVR in both the \textit{Dynamic} and \textit{Layout} categories, although its advantage varies with the setting. Its use of predicted target trajectories enables more flexible collision-avoidance maneuvers by tracking a trajectory rather than a single point. Representative examples are illustrated in Fig.~\ref{fig:expSimCases}(a) and (b). Another notable feature is its braking behavior~\cite{sekiguchi2021uncertainty} when the predicted target trajectory carries high uncertainty, which may help avoid collisions. However, this braking tendency reduces responsiveness in the \textit{Trajectories} category, where sharp turns (e.g., L-turns and two-triangle patterns) demand rapid adaptation to a suitable following position rather than stopping. This suggests a trade-off between cautious responses to uncertain predictions and responsiveness to abrupt target maneuvers.

In the two settings in Table~\ref{tab:ciStats}, MPC w/ Traj. shows higher mean TinTPerson but also higher mean Jerk, while the SR differences do not establish an ordering among the three MPC variants. It obtains 52\% SR and 45.7\% TinTPerson in \textit{Perpendicular Crossing}, and 52\% SR and 49.6\% TinTPerson in \textit{Cluttered Space}. Its mean Jerk reaches 28.96 and 25.39\,m/s$^3$, respectively, compared with approximately 20--24\,m/s$^3$ for MPC and MPC w/ DS., consistent with abrupt braking under trajectory uncertainty. The table reports Wilson 95\% SR confidence intervals and mean $\pm$ standard deviation for the continuous metrics over 100 trials per setting. For trials paired by their shared initial environment configurations, two-sided exact McNemar tests find no significant SR differences after Holm correction across the six comparisons (all adjusted $p=1.000$, $\alpha=0.05$). The large standard deviations further indicate substantial trial-to-trial variability. We therefore interpret the continuous metrics descriptively; the SR tests apply to these two settings and do not establish aggregate ranking significance.

MPC w/ DS. shows similar observed aggregate SRs to standalone MPC in the \textit{Dynamic} and \textit{Layout} categories, with limited improvement from the dynamic search field. For both planners, SR and ASR are nearly identical, indicating that most failures arise from collisions. Our simulated environments involve complex interactions in which pedestrians' future trajectories are difficult to predict due to frequent encounters. However, the dynamic search field~\cite{ye2025rpf} relies heavily on accurate trajectory prediction of occluders; when the predicted field is inaccurate, the resulting search actions may increase collision risk. Meanwhile, the underlying MPC planner~\cite{han2023rda} does not proactively plan maneuvers around dynamic occluders. Thus, once target occlusion occurs, the robot may have little room to maneuver for overtaking or following occluders, limiting the opportunity for the dynamic search field to demonstrate its advantages.

\begin{table}[t]
\centering

\caption{Performance summary of MPC-based planners across two representative scenarios under back-following at 1.5\,m. \textit{Dynamic: Perpendicular Crossing}: $N_\text{human}=20$ (100 trials). \textit{Layout: Cluttered Space}: 20 obstacles, 30 humans (100 trials). SR is reported with Wilson 95\% confidence intervals; continuous metrics are reported as mean {\scriptsize $\pm$ std}. Higher SR and TinTPerson are better ($\uparrow$); lower Jerk is better ($\downarrow$).}
\label{tab:ciStats}
\scalebox{0.73}{
\begin{tabular}{l ccc}
\toprule
\textbf{Planner} & \textbf{SR$\uparrow$ (\%, 95\% CI)} & \textbf{TinTPerson$\uparrow$ (\%)} & \textbf{Jerk$\downarrow$ (m/s$^3$)} \\
\midrule
\midrule
\multicolumn{4}{l}{\textit{Dynamic: Perpendicular Crossing} ($N_\text{human}=20$)} \\
\midrule
MPC~\cite{han2023rda}            & 47.0 {\scriptsize [37.5, 56.7]}                & 41.2 {\scriptsize $\pm$ 34.9} & 23.96 {\scriptsize $\pm$ 11.03} \\
MPC w/ Traj.~\cite{sekiguchi2021uncertainty}   & 52.0 {\scriptsize [42.3, 61.5]}       & 45.7 {\scriptsize $\pm$ 37.4} & 28.96 {\scriptsize $\pm$ 14.62} \\
MPC w/ DS.~\cite{ye2025rpf}     & 51.0 {\scriptsize [41.3, 60.6]}                & 44.1 {\scriptsize $\pm$ 34.8} & {23.66} {\scriptsize $\pm$ 11.71} \\
\midrule
\multicolumn{4}{l}{\textit{Layout: Cluttered Space} (20 obstacles, 30 humans)} \\
\midrule
MPC~\cite{han2023rda}            & 44.0 {\scriptsize [34.7, 53.8]}                & 36.0 {\scriptsize $\pm$ 23.8} & 19.63 {\scriptsize $\pm$ 6.51} \\
MPC w/ Traj.~\cite{sekiguchi2021uncertainty}   & 52.0 {\scriptsize [42.3, 61.5]}       & {49.6 {\scriptsize $\pm$ 31.1}} & 25.39 {\scriptsize $\pm$ 10.49} \\
MPC w/ DS.~\cite{ye2025rpf}     & 45.0 {\scriptsize [35.6, 54.8]}                & 36.0 {\scriptsize $\pm$ 24.1} & 19.94 {\scriptsize $\pm$ 7.16} \\
\bottomrule
\end{tabular}
}
\end{table}

\subsection{Performance under Different Environmental Settings} \label{sec:expDensity}
The results under different human densities and environmental conditions are shown in Fig.~\ref{fig:expHumanOccupancy}. In the evaluated \textit{Dynamic} and \textit{Layout} configurations, back-following generally shows higher observed SR, ASR, and TVR than side-following. Back-following reduces interference with pedestrian flows, thereby lowering collision risks and improving target visibility. In contrast, side-following is more prone to occlusions and conflicts with pedestrians or static obstacles, leading to reduced SR, ASR, and TVR. Illustrative examples are shown in Fig.~\ref{fig:expSimCases} (c) and (d), where side-following trajectories significantly interact with obstacles and pedestrian movements, ultimately leading to collisions. However, it is worth noting that side-following is often perceived as more comfortable and socially acceptable in real-world applications, as it allows for better interaction and communication with the target person \cite{repiso2017line,repiso2019people,ferrer2017robot}.

Planner performance degrades with increasing human density, as frequent pedestrian interactions produce less predictable trajectories and reduce available free space, thereby elevating the risk of collisions and occlusions. A similar trend is observed in \textit{Layout (O)}: narrower passageways or higher obstacle counts effectively reduce navigable space, analogous to the effect of increased human density. However, the degradation in \textit{Layout (O)} is less pronounced than in \textit{Layout (H)} under back-following. This can be attributed to structural constraints in environments such as corridors, doorways, and intersections, which limit the degrees of freedom in both target and pedestrian trajectories. As a result, back-following becomes nearly equivalent to tracking the target's trajectory directly, reducing the likelihood of interactions and occlusions with other pedestrians.
In contrast, side-following remains highly sensitive, as the follower is exposed to continuous interactions with nearby pedestrians and obstacles. In the plotted occupancy conditions, side-following shows lower observed SRs, with values below 30\%.

The results under different following distances are presented in Fig.~\ref{fig:expFollowDist}. Overall, both overly close (1.0 m) and overly distant (2.0 m or 2.5 m) following distances degrade the performance of the three MPC-based implementations (MPC, MPC w/ Traj., and MPC w/ DS.) across different environmental layouts (\textit{Layout (F)}). In contrast, this degradation is less pronounced in \textit{Dynamic (F)}; in some cases, the planners even perform better (e.g., MPC-based and MPC w/ DS.) because open spaces provide greater maneuvering room for collision avoidance. The additional room also facilitates person-search, though at farther distances the robot tends to lose track of the target more frequently. Conversely, in \textit{Layout (F)}, narrow environments constrain the available search space, resulting in more collisions. Many of these collisions arise from ambiguities in the preferred following distances relative to environmental structures. For example, in corridors or intersections, some preferred following positions may lie close to walls, thereby creating riskier situations.

Another notable observation from Fig.~\ref{fig:expFollowDist} is that MPC w/ Traj. performs worse than MPC-based and MPC w/ DS. when the following distance is 1.0, 2.0, or 2.5 m. We attribute this to its braking behavior: at short distances (e.g., 1.0 m), abrupt braking increases the likelihood of collisions with surrounding pedestrians. At longer distances (e.g., 2.0 m and 2.5 m), tracking a distant trajectory slows down responsiveness, whereas the point-tracking strategies of MPC-based and MPC w/ DS. allow quicker adjustments. Faster response is also advantageous for re-finding the target after occlusion. Between the two, MPC w/ DS. shows higher observed person-search performance than MPC, as the enlarged search space benefits the construction of the dynamic search field, thereby enabling safer and more robust recovery behaviors.

\begin{table}[t]
\centering
\caption{Planning time comparison across different RPF planners (mean $\pm$ variance, in ms). Tested on a PC with Intel®
Core™ i9-12900K CPU.}
\begin{tabular}{lc}
\toprule
\textbf{Planner} & \textbf{Time (ms)} \\
\midrule
\midrule
MPC~\cite{han2023rda}            & 48.4 {\scriptsize $\pm$ 37.1} \\
MPC w/ Traj.~\cite{sekiguchi2021uncertainty}   & 94.0 {\scriptsize $\pm$ 74.8} \\
MPC w/ DS.~\cite{ye2025rpf}     & 48.1 {\scriptsize $\pm$ 37.1} \\
SFM~\cite{ferrer2017robot}            & 0.1 {\scriptsize $\pm$ 0.0} \\
DWA~\cite{van2022collision}            & 3.3 {\scriptsize $\pm$ 2.1} \\
DWA w/ Traj.   & 11.5 {\scriptsize $\pm$ 7.1} \\
BSO-HFC~\cite{lyu2025robust}        & 19.76 {\scriptsize $\pm$ 32.3} \\
RL-based~\cite{leisiazar2025adapting}       & 73.3 {\scriptsize $\pm$ 75.4} \\
\bottomrule
\end{tabular}
\label{tab:planTime}
\vspace{-0.3cm}
\end{table}

\subsection{Planning Time Comparison}
The computational efficiency results (as shown in Table~\ref{tab:planTime}) indicate a clear trade-off between planning quality and runtime. The SFM and DWA-based planners exhibit extremely low computation times (below 15 ms), ensuring real-time feasibility. However, their limited optimization capability may compromise planning quality. In contrast, the MPC-based planners require substantially more computation (50-100 ms), with MPC w/ Traj. being the most time-consuming. Although this additional cost may hinder high-frequency control, it potentially provides stronger optimization-based guarantees in complex environments.

\begin{figure}[t]
    \centering
    \includegraphics[width=0.9\linewidth]{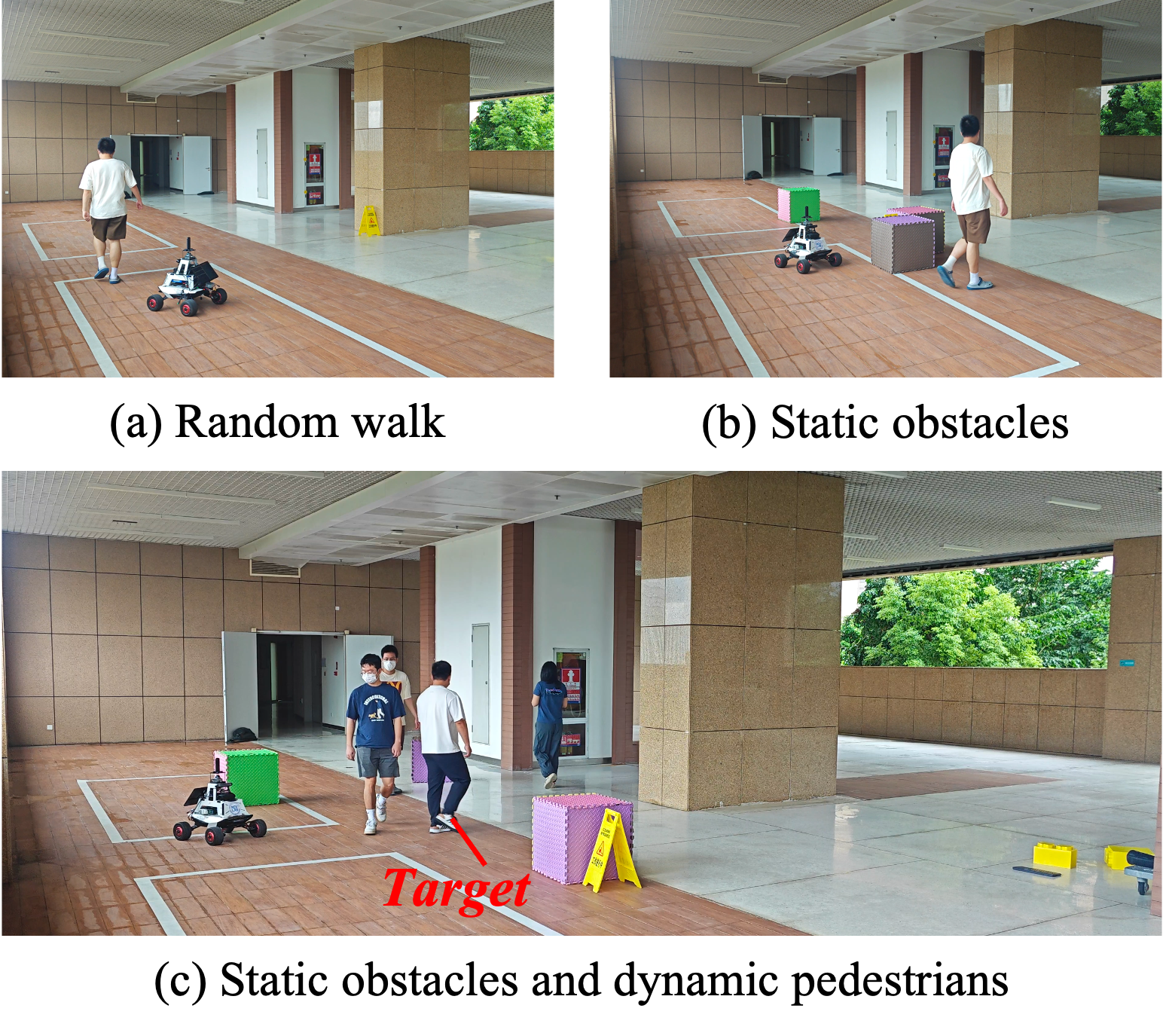}
    \caption{Real-world experiments of MPC-based and MPC w/ Traj. planners on a differential-drive robot across various scenarios: (a) random walks with multiple turns in open space, (b) walks in environments with static obstacles, and (c) walks in environments with both static obstacles and moving pedestrians.}
    \label{fig:realWorld}
\end{figure}

\subsection{Real-world Experiments} \label{sec:expRealWorld}

To examine deployment feasibility and practical following behaviors, we implemented MPC and MPC w/ Traj. on a differential-drive robot (Scout-mini) equipped with a ZED2 camera for pedestrian detection and target identification~\cite{ye2024reid}, as well as a MID-360 LiDAR for obstacle detection and localization via FAST-LIO~\cite{xu2021fast}. The onboard computer was an Intel NUC 11 mini PC with an Intel Core i7-1165G7 CPU. The evaluated scenarios (Fig.~\ref{fig:realWorld}) included: (i) a target walking randomly with abrupt turns in an open space; (ii) a target walking in an environment with static obstacles; and (iii) a target walking in an environment containing both static obstacles and moving pedestrians. The preferred following distance was set to 1.5~m, and the robot maintained a back-following configuration. Visualizations of the experiments are provided in the attached video. For each scenario, we conducted ten trials involving three participants (all male, with an average age of 25 years) to observe planner behavior and collect qualitative participant feedback.

The trials illustrate target-following and obstacle-avoidance behaviors in the tested settings, together with practical limitations. In the open-space scenario, both planners were observed to follow the target stably and respond to abrupt maneuvers. Participant interviews suggested smoother following and less aggressive braking with MPC w/ Traj. in this setting; these qualitative impressions do not establish lower physical jerk or a general comfort advantage. In the static-obstacle scenario, both planners were generally able to avoid collisions, but MPC occasionally became stuck when the target made sharp turns behind obstacles. MPC w/ Traj. generated wider-arc motions that reduced close encounters with obstacles, although such detours could lead to temporary target loss under a limited camera field of view. Similar behaviors were observed with moving pedestrians. Reversing or detouring maneuvers also avoided collisions in observed cases of sudden pedestrian interference. These responses remained reactive, since pedestrian trajectory prediction was not incorporated into either planner.




\section{Discussion} \label{sec:discussion}
We present our findings through experiments conducted on Follow-Bench and in real-world environments, and outline the challenges and future directions of RPF-driven planners (Sec. \ref{sec:challenges}). Furthermore, we discuss the limitations of our benchmark and outline potential future work (Sec. \ref{sec:limitations}).

\subsection{Challenges and Future Directions} \label{sec:challenges}

\subsubsection{\textbf{Balancing Trade-offs between Safety and Comfort}} \label{sec:discussionSafetyComfort}
Currently, most RPF planners consider comfort constraints by forcing robots to track a point or a trajectory that is in the social space of the target person. However, this reference point or trajectory is often occluded by surrounding pedestrians or static obstacles, which can lead to potential collisions. This is the contradiction between safety and comfort. Although collision-avoidance modules are typically incorporated, the conflicting objectives of maintaining proximity to the target and avoiding obstacles make it challenging for optimization-based planners to consistently generate feasible trajectories—particularly in dynamic environments where the target moves freely. Our experiments further confirm this difficulty: in cluttered settings such as \textit{Layout} and \textit{Dynamic}, collisions and occlusions occur more frequently, thereby reducing the overall success rate of person-following. The degradation is even more pronounced in side-following configurations (a position generally considered more comfortable), where the robot is more exposed to interactions with both pedestrians and obstacles.

Rather than following a fixed point or trajectory anchored to the target's latest position, a more promising direction is to actively estimate and pursue an adaptive trajectory that accounts for predicted human motions and surrounding static structures. Wang \textit{et al.} \cite{wang2024continuous} propose such an active-following approach by embedding both visibility maximization and collision avoidance as soft constraints within a nonlinear MPC framework. While effective in principle, this formulation introduces multiple coupled factors into the cost function, leading to highly non-convex optimization problems that are difficult to solve in real time. Moreover, the evaluated environments are relatively simple—containing only a few distractors and structural obstacles—and the codebase is not publicly available for straightforward benchmarking. Potential improvements include convexifying certain constraints (e.g., visibility and collision avoidance) or leveraging sampling-based MPC methods \cite{williams2018information} to mitigate computational overhead.

Another viable strategy is to decouple the RPF planning problem into hierarchical subproblems, thereby reducing computational complexity. For instance, RPF-Search \cite{ye2025rpf} addresses person-search in dynamic environments by: (i) sampling and assessing candidate waypoints around occluders, and (ii) selecting the best waypoint for overtaking or following, which is then executed by a local planner. Similarly, the active RPF problem can be structured into a high-level planner that generates candidate accompanied points around the target optimized for visibility and safety, based on predicted pedestrian trajectories and environmental context, and a low-level planner that selects the optimal waypoint and computes a collision-free trajectory towards it.

\subsubsection{\textbf{Improving Spatial-temporal Obstacle Representation, Prediction and Planning}} \label{sec:discussionObsRep}
The RDA-based MPC implementations represent static obstacles as polygons derived from DBSCAN clusters~\cite{ester1996density}, enabling efficient planning. This approach works well for discrete obstacles, such as cluttered objects (see Fig.~\ref{fig:benchTopographies}). However, we found that it is less accurate for continuous obstacles, such as walls and doorways. Moreover, it may incorrectly merge the target and nearby obstacles into a single large polygon, leading to suboptimal or even erroneous planning. In the Cluttered Space obstacle-representation sensitivity study reported in the Appendix, finer bounding boxes or minimum-enclosing circles ($\texttt{eps}=0.5$) yield higher observed MPC success rates than the default coarse boxes ($\texttt{eps}=2.0$), consistent with reduced obstacle merging. MPC has a higher observed SR than the non-MPC baselines under all three representations in this setting. Therefore, exploring a more robust and efficient obstacle representation that can handle environments with both static and dynamic obstacles remains a promising direction. In addition, the safe distance between the robot and obstacles should be adaptive to different human-robot coexistence scenarios—larger distances in cluttered environments and smaller distances in narrow spaces (e.g., corridors and doorways)---to balance target visibility and collision avoidance for successful target following.

MPC w/ Traj. achieves higher observed aggregate SR than standalone MPC in the Dynamic and Layout categories, suggesting benefits from target-trajectory forecasts; however, its higher mean Jerk in the two analyzed settings indicates a trade-off with motion smoothness. In this work, we only exploit the target's future trajectory. A natural extension would be to incorporate the predicted trajectories of surrounding pedestrians, which have demonstrated promising results for collision avoidance in independent social navigation \cite{ryu2024integrating}. For this purpose, instead of a simple Kalman filter, more advanced predictors that account for human-human interactions, such as Social-GAN \cite{gupta2018social}, may further improve performance. Moreover, within the RPF setting, it is important to consider not only collision avoidance but also target visibility maximization and proxemic relationships with both the target and surrounding pedestrians, guided by their predicted trajectories.

\subsubsection{\textbf{Discovering and Utilizing Environmental Patterns}}
In addition to proactive planning over a short-term horizon (e.g., several seconds for trajectory prediction and motion planning), leveraging long-term environmental patterns could further benefit RPF in complex environments. For instance, when accompanying a target across a crowded crosswalk (similar to parallel crossing in Fig. \ref{fig:benchDynamicCrowds}), rather than aggressively occupying a nearby position, the robot may temporarily follow the pedestrian flow \cite{dugas2022flowbot} until a safe following position becomes available, resulting in more comfortable and safer behavior. Similarly, when the target moves through a crowd (e.g., circular or random formations in Fig. \ref{fig:benchDynamicCrowds} or intersections in Fig. \ref{fig:benchTopographies}), proactively recognizing the underlying crowd pattern and selecting a less congested path~\cite{sun2025socially} to re-approach the target could improve both safety and comfort, even if it requires a temporary sacrifice of target visibility. Such a high-level planner that captures long-term environmental structures while guiding the low-level RPF planner represents a promising research direction.

\subsubsection{\textbf{Towards More Expressive Intermediate Representations}} 
Intermediate representations continue to be a fundamental challenge in RPF research. Our experiments reveal that point-based or trajectory-based references, coupled with obstacle-avoidance modules, often fail to maintain stable following behavior in crowded environments with frequent occlusions and obstructions. In addition to active following and hierarchical approaches (discussed in Sec. \ref{sec:discussionSafetyComfort}), an alternative solution is to develop more expressive intermediate representations that integrate multiple information sources and constraints to guide the robot's motion planning.

A common strategy is to use cost maps \cite{tan2025robust, zhang2023collision} or potential fields \cite{hoeller2007iros, yuan2018laser}, which provide continuous cost fields that encode environmental obstacles. These methods allow the robot to select follow points or plan paths within feasible regions, rather than relying on isolated instantaneous target points or trajectories. However, they primarily encode obstacle presence and not explicit distances, and updating these maps in real time can be computationally expensive. Euclidean Signed Distance Fields (ESDF) offer a complementary approach by providing a continuous, differentiable distance field that explicitly encodes obstacle distances. ESDFs enable the robot to plan smooth and safe trajectories for efficient target tracking while supporting real-time updates in dynamic environments \cite{ji2022elastic, zhu2024efficient}. However, updating ESDFs in highly dynamic environments can still be computationally demanding.

Beyond collision avoidance, effective intermediate representations must also incorporate additional constraints to improve RPF planning, such as target visibility and social compliance. However, fulfilling all these constraints can result in computational burdens and potentially conflicting objectives. Future research could explore more efficient, adaptive, and expressive representations, for example, by leveraging learning-based methods \cite{liu2023icra} to encode complex constraints and interactions in a compact form.

\subsection{Limitations and Future Work} \label{sec:limitations}
\subsubsection{\textbf{More Context-rich Environments}}
Currently, Follow-Bench encompasses the most representative scenarios in RPF research, including various target trajectory patterns, crowd dynamics, and environmental layouts. Nevertheless, it primarily evaluates basic motion-planning capabilities in terms of safety and comfort. Future extensions could introduce more context-rich environments, such as hospitals, factories, museums, households, and offices. In these settings, the target person may exhibit application-specific behaviors and interactions with the environment (e.g., visiting a patient, collaborating with colleagues, or shopping), while the robot could be assigned complementary task-oriented goals during following (e.g., delivering medicine, carrying luggage, or guiding to a destination). Such scenarios would demand more sophisticated reasoning and decision-making from RPF planners.

Furthermore, the current benchmark is implemented in a 2D simulator to facilitate efficient evaluation and development of RPF planners due to its simplicity and CPU-only compatibility. It is worth noting that the interaction mechanisms in many high-fidelity simulators~\cite{kastner2024arena,tsoi2022sean2} remain fundamentally 2D-based---they first generate socially aware 2D human trajectories and subsequently render them with 3D animations for enhanced visualization. Nevertheless, adopting a more realistic 3D simulator (e.g., IsaacSim~\cite{NVIDIAIsaacSim}) would enable richer evaluations of RPF systems under perceptual challenges such as human-trajectory prediction, visual occlusion~\cite{ye2025rpf}, visual localization on quadruped platforms~\cite{zhan2025monocular}, and semantic understanding~\cite{Bruckschen2020Human}. Moreover, photorealistic 3D simulation would create broader opportunities for the evaluation and advancement of learning-based methods, such as vision-language-action models~\cite{wang2025trackvla,liu2025trackvla++}, which unify visual perception, language understanding, and motion planning. For example, ReferTrack~\cite{ye2026refertrack} identifies a language-described target among image-space bounding boxes and then predicts tracking waypoints conditioned on that selection, using a single forward-facing camera. 

The current 2D, state-based setting should therefore be viewed as a controlled benchmark for RPF motion-planning capability, rather than a substitute for future 3D embodied evaluation. By assuming relatively complete and low-noise state observations, Follow-Bench isolates the influence of planner design from upstream perception errors. In contrast, a 3D simulator would introduce additional uncertainty from target tracking, visual occlusion, depth estimation, semantic perception, and 3D obstacle representation. These factors may further amplify failures observed in the 2D setting, while also revealing different robustness characteristics among planners with otherwise similar performance, especially for optimization-based methods that directly depend on target and obstacle states. In this sense, Follow-Bench can provide useful prior evidence for future 3D RPF benchmarks: its re-implemented planners can be extended with target-tracking and 3D obstacle-representation modules, and the resulting planner-level comparisons can help identify promising candidates for real-robot deployment.

\subsubsection{\textbf{More Realistic and Diverse Interaction Simulation}}
Currently, Follow-Bench incorporates human-human and human-object interactions through the ORCA and SFM models, which are commonly used in social navigation research. Nevertheless, these models fall short in capturing the full complexity of real-world interactions, such as heterogeneous walking speeds, diverse interaction patterns, and varied obstacle-avoidance behaviors. Future extensions could exploit real-world datasets~\cite{ye2025tpt,pellegrini2009you} to derive more realistic interaction dynamics, or employ more sophisticated interaction models~\cite{chandra2025multi} that better approximate human behavior in dynamic environments. Such enhancements would substantially improve the benchmark's capacity to evaluate RPF planners under challenging and realistic conditions.

\subsubsection{\textbf{Limited Real-world and Comfort Evaluation}}
The real-world trials provide limited feasibility evidence for MPC and MPC w/ Traj. Only these two implementations were deployed, with three male participants and ten trials per scenario, restricting both method coverage and participant diversity. The reported behaviors and interview feedback are qualitative and cannot validate the relative ranking of all eight planners or establish general safety and perceived-comfort benefits. Moreover, the benchmark's interpersonal-distance and motion-smoothness indicators describe physical behavior without directly measuring perceived comfort. Broader deployment comparisons and controlled user studies with more diverse participants and validated comfort measures are needed to assess these aspects.

\section{Conclusions} \label{sec:conclusion}
Robot person following (RPF) is a fundamental capability in human-robot interaction, with applications spanning personal assistance, security, service, and industrial robotics. This paper provides a comprehensive and structured survey of existing RPF research, with a particular emphasis on two widely recognized and critical criteria: safety and comfort. Guided by these objectives, we review representative scenarios, evaluation metrics, and motion planners. To enable systematic benchmarking of reviewed planners, we introduce Follow-Bench, a unified benchmark implemented in a lightweight simulator that models socially aware interactions, including diverse target trajectory patterns, crowd dynamics, and environmental layouts. To the best of our knowledge, this is the first benchmark to explicitly highlight two essential requirements for RPF: (i) continuously and naturally following a target person, and (ii) proactively avoiding obstacles and pedestrians in socially interactive environments. Follow-Bench allows researchers to develop and evaluate new RPF planners, with flexible support for adding maps, obstacles, agents, metrics, and following strategies. All code and deployment scripts are publicly released.

Using Follow-Bench, we evaluate eight implementations derived from representative RPF formulations under a unified safety--comfort protocol; the resulting findings characterize these Follow-Bench implementations rather than the unmodified original systems. Real-world trials with MPC and MPC w/ Traj. demonstrate deployment feasibility and illustrate practical following behaviors in the tested settings. Our results show that the MPC-based implementations achieve higher observed aggregate success rates across the evaluated Dynamic and Layout scenarios. MPC w/ Traj. attains the highest observed aggregate SR and improved target proximity in several settings, but its advantage is scenario-dependent and accompanied by higher jerk; paired tests do not establish an SR ordering among the MPC variants in the two analyzed settings. Nevertheless, all planners show reduced performance in high-density scenarios, particularly in side-following configurations with frequent occlusions and complex interactions.

Finally, we discuss open challenges and future research directions for RPF-driven planners, alongside the current limitations of our benchmark. We believe that the comprehensive survey and unified evaluation framework presented in this paper will provide a solid foundation for future research, particularly toward developing socially aware RPF planners that can robustly balance safety and comfort in real-world environments.





\bibliographystyle{./bibliography/IEEEtran}
\bibliography{./bibliography/ref}

@STRING{ICRA = "Proc. {IEEE} Int. Conf. Robot. Autom."}

@STRING{IROS = "Proc. {IEEE/RJS} Int. Conf. Intell. Robots Syst."}

@STRING{HRI = "Proc. {ACM/IEEE} Int. Conf. Human-Robot Interact."}

@STRING{CoRL = "Proc. {IEEE} Conf. Rob. Learn."}

@STRING{ECMR = "Proc. Eur. Conf. Mobile Robots (ECMR)"}

@STRING{CogInfoCom = "Proc. {IEEE} Int. Conf. Cogn. Infocommun. (CogInfoCom)"}

@STRING{Humanoids = "Proc. {IEEE/RAS} Int. Conf. Humanoid Robots (Humanoids)"}

@STRING{CYBER = "Proc. {IEEE} Int. Conf. CYBER Technol. Autom., Control, Intell. Syst. (CYBER)"}

@STRING{ARSO = "Proc. {IEEE} Int. Conf. Adv. Robot. Social Impacts (ARSO)"}

@STRING{RO-MAN = "Proc. {IEEE} Int. Symp. Robot and Human Interactive Commun. (RO-MAN)"}

@STRING{CVPR = "Proc. {IEEE} Conf. Comput. Vis. Pattern Recognit."}

@STRING{ICCV = "Proc. {IEEE} Int. Conf. Comput. Vis."}

@STRING{ICMA = "Proc. IEEE Int. Conf. Mechatronics Autom. (ICMA)"}

@STRING{IES = "Proc. {IEEE} Int. Electron. Symp. (IES)"}

@STRING{IJRR = "Int. J. Robot. Res. (IJRR)"}

@STRING{TRO = "{IEEE} Trans. Robot."}

@STRING{TIM = "{IEEE} Trans. Instrum. and Meas."}

@STRING{RAL = "{IEEE} Robot. Autom. Lett."}

@STRING{RAS = "Robot. and Auton. Syst."}

@STRING{ICCVS = "Int. Conf. Comput. Vis. Syst."}

@STRING{CRV = "Conf. Comput. Rob. Vis. (CRV)"}

@STRING{IJSR = "Int. J. Social Robot. (IJSR)"}

@STRING{HUMANOIDS = "{IEEE} Int. Conf. Humanoid Robots"}

@STRING{TSMCS = "{IEEE} Trans. Syst., Man, Cybern., Syst."}

@STRING{TASE = "{IEEE} Trans. Autom. Sci. Eng."}

@STRING{TAES = "{IEEE} Trans. Aerosp. Electron. Syst."}

@STRING{AR = "Autonomous Robots"}

@STRING{URAI = "{IEEE} Int. Conf. Ubiquitous Robots and Ambient Intelligence (URAI)"}

@STRING{TMECH = "{IEEE/ASME} Trans. Mechatronics"}

@STRING{TCDS = "{IEEE} Trans. Trans. Cogn. Dev. Syst."}

@STRING{THRI = "{ACM} Trans. Hum.-Robot Interact."}

@STRING{THMS = "{IEEE} Trans. Hum.-Mach. Syst."}

@article{islam2019person,
  title={Person-following by autonomous robots: A categorical overview},
  author={Islam, Md Jahidul and Hong, Jungseok and Sattar, Junaed},
  journal=IJRR,
  volume={38},
  number={14},
  pages={1581--1618},
  year={2019},
  publisher={SAGE Publications Sage UK: London, England}
}

@article{leisiazar2025adapting,
  title={Adapting to Frequent Human Direction Changes in Autonomous Frontal Following Robots},
  author={Leisiazar, Sahar and Rohani, Seyed Roozbeh Razavi and Park, Edward J and Lim, Angelica and Chen, Mo},
  journal=RAL,
  year={2025},
  publisher={IEEE}
}

@article{repiso2024adaptive,
  title={Adaptive social planner to accompany people in real-life dynamic environments},
  author={Repiso, Ely and Garrell, Ana{\'\i}s and Sanfeliu, Alberto},
  journal=IJSR,
  volume={16},
  number={6},
  pages={1189--1221},
  year={2024},
  publisher={Springer}
}

@inproceedings{peng2024dual,
  title={A Dual Closed-Loop Control Strategy for Human-Following Robots Respecting Social Space},
  author={Peng, Jianwei and Liao, Zhelin and Su, Zefan and Yao, Hanchen and Zeng, Yadan and Dai, Houde},
  booktitle=ICRA,
  pages={11252--11258},
  year={2024},
  organization={IEEE}
}

@inproceedings{sekiguchi2021uncertainty,
  title={Uncertainty-aware non-linear model predictive control for human-following companion robot},
  author={Sekiguchi, Shunichi and Yorozu, Ayanori and Kuno, Kazuhiro and Okada, Masaki and Watanabe, Yutaka and Takahashi, Masaki},
  booktitle=ICRA,
  pages={8316--8322},
  year={2021}
}

@article{montesdeoca2022person,
  title={Person-following controller with socially acceptable robot motion},
  author={Montesdeoca, Julio and Toibero, J Marcos and Jordan, Julian and Zell, Andreas and Carelli, Ricardo},
  journal=RAS,
  volume={153},
  pages={104075},
  year={2022},
  publisher={Elsevier}
}

@article{van2022collision,
  title={Collision-free navigation in human-following task using a cognitive robotic system on differential drive vehicles},
  author={Van Dang, Chien and Ahn, Heungju and Kim, Jong-Wook and Lee, Sang C},
  journal=TCDS,
  volume={15},
  number={1},
  pages={78--87},
  year={2022},
  publisher={IEEE}
}

@article{eirale2022human,
  title={Human-centered navigation and person-following with omnidirectional robot for indoor assistance and monitoring},
  author={Eirale, Andrea and Martini, Mauro and Chiaberge, Marcello},
  journal={Robotics},
  volume={11},
  number={5},
  pages={108},
  year={2022},
  publisher={MDPI}
}

@ARTICLE{ye2025rpf,
  author={Ye, Hanjing and Cai, Kuanqi and Zhan, Yu and Xia, Bingyi and Ajoudani, Arash and Zhang, Hong},
  journal=TMECH, 
  title={RPF-Search: Field-Based Search for Robot Person Following in Unknown Dynamic Environments}, 
  year={2025},
  volume={},
  number={},
  pages={1-12}
}

@article{ye2025tpt,
  title={TPT-Bench: A Large-Scale, Long-Term and Robot-Egocentric Dataset for Benchmarking Target Person Tracking},
  author={Ye, Hanjing and Zhan, Yu and Situ, Weixi and Chen, Guangcheng and Yu, Jingwen and Cai, Kuanqi and Zhang, Hong},
  journal={arXiv preprint arXiv:2503.02188},
  year={2025}
}

@inproceedings{ye2023robot,
  title={Robot person following under partial occlusion},
  author={Ye, Hanjing and Zhao, Jieting and Pan, Yaling and Cherr, Weinan and He, Li and Zhang, Hong},
  booktitle=ICRA,
  pages={7591--7597},
  year={2023},
  organization={IEEE}
}

@inproceedings{leisiazar2023mcts,
  title={An MCTS-DRL based obstacle and occlusion avoidance methodology in robotic follow-ahead applications},
  author={Leisiazar, Sahar and Park, Edward J and Lim, Angelica and Chen, Mo},
  booktitle=IROS,
  pages={221--228},
  year={2023},
  organization={IEEE}
}

@inproceedings{zhao2024human,
  title={Human Orientation Estimation Under Partial Observation},
  author={Zhao, Jieting and Ye, Hanjing and Zhan, Yu and Luan, Hao and Zhang, Hong},
  booktitle=IROS,
  pages={11544--11551},
  year={2024},
  organization={IEEE}
}

@article{yuan2018laser,
  title={Laser-based intersection-aware human following with a mobile robot in indoor environments},
  author={Yuan, Jing and Zhang, Shengming and Sun, Qinxuan and Liu, Gangdun and Cai, Jingxin},
  journal=TSMCS,
  volume={51},
  number={1},
  pages={354--369},
  year={2018},
  publisher={IEEE}
}

@INPROCEEDINGS{Ferrer2013sfm,
  author={Ferrer, Gonzalo and Garrell, Anaís and Sanfeliu, Alberto},
  booktitle=IROS, 
  title={Robot companion: A social-force based approach with human awareness-navigation in crowded environments}, 
  year={2013},
  pages={1688-1694},
  doi={10.1109/IROS.2013.6696576}
}

@article{algabri2021target,
  title={Target recovery for robust deep learning-based person following in mobile robots: Online trajectory prediction},
  author={Algabri, Redhwan and Choi, Mun-Taek},
  journal={Applied Sciences},
  volume={11},
  number={9},
  pages={4165},
  year={2021},
  publisher={MDPI}
}

@INPROCEEDINGS{Hoeller2007Accompanying,
  author={Hoeller, Frank and Schulz, Dirk and Moors, Mark and Schneider, Frank E.},
  booktitle=IROS, 
  title={Accompanying persons with a mobile robot using motion prediction and probabilistic roadmaps}, 
  year={2007},
  pages={1260-1265},
  doi={10.1109/IROS.2007.4399194}
}

@article{repisoadaptive2020,
	title = {Adaptive {Side}-by-{Side} {Social} {Robot} {Navigation} to {Approach} and {Interact} with {People}},
	volume = {12},
	issn = {1875-4791, 1875-4805},
	doi = {10.1007/s12369-019-00559-2},
	language = {en},
	number = {4},
	urldate = {2023-08-30},
	journal = IJSR,
	author = {Repiso, Ely and Garrell, Anaís and Sanfeliu, Alberto},
	month = aug,
	year = {2020},
	note = {Number: 4},
	pages = {909--930},
}

@article{hu2014tie,
	title = {Design of {Sensing} {System} and {Anticipative} {Behavior} for {Human} {Following} of {Mobile} {Robots}},
	volume = {61},
	issn = {0278-0046, 1557-9948},
	doi = {10.1109/TIE.2013.2262758},
	language = {en},
	number = {4},
	urldate = {2023-07-30},
	journal = TIE,
	author = {Hu, Jwu-Sheng and Wang, Jyun-Ji and Ho, Daniel Minare},
	month = apr,
	year = {2014},
	pages = {1916--1927},
}

@inproceedings{bayoumi2016learning,
	address = {Stockholm, Sweden},
	title = {Learning optimal navigation actions for foresighted robot behavior during assistance tasks},
	isbn = {978-1-4673-8026-3},
	doi = {10.1109/ICRA.2016.7487135},
	urldate = {2024-04-24},
	booktitle = ICRA,
	publisher = {IEEE},
	author = {Bayoumi, AbdElMoniem and Bennewitz, Maren},
	month = may,
	year = {2016},
	pages = {207--212},
}

@INPROCEEDINGS{nikdel2020lbgp,
  author={Nikdel, Payam and Vaughan, Richard and Chen, Mo},
  booktitle=ICRA, 
  title={LBGP: Learning Based Goal Planning for Autonomous Following in Front}, 
  year={2021},
  pages={3140-3146},
  doi={10.1109/ICRA48506.2021.9560914}}

@InProceedings{Kuderer2014Approach,
author="Kuderer, Markus
and Burgard, Wolfram",
editor="Beetz, Michael
and Johnston, Benjamin
and Williams, Mary-Anne",
title="An Approach to Socially Compliant Leader Following for Mobile Robots",
booktitle="Social Robotics",
year="2014",
publisher="Springer International Publishing",
address="Cham",
pages="239--248",
}

@INPROCEEDINGS{Park2013Autonomous,
  author={Park, Jong Jin and Kuipers, Benjamin},
  booktitle=ICRA, 
  title={Autonomous person pacing and following with Model Predictive Equilibrium Point Control}, 
  year={2013},
  pages={1060-1067},
  doi={10.1109/ICRA.2013.6630704}
}

@INPROCEEDINGS{Morales2012How,
  author={Morales, Yoichi and Satake, Satoru and Huq, Rajibul and Glas, Dylan and Kanda, Takayuki and Hagita, Norihiro},
  booktitle=HRI, 
  title={How do people walk side-by-side? — Using a computational model of human behavior for a social robot}, 
  year={2012},
  pages={301-308},
  doi={}}

@INPROCEEDINGS{Bruckschen2020Human,
  author={Bruckschen, Lilli and Bungert, Kira and Dengler, Nils and Bennewitz, Maren},
  booktitle=IROS, 
  title={Human-Aware Robot Navigation by Long-Term Movement Prediction}, 
  year={2020},
  pages={11032-11037},
  doi={10.1109/IROS45743.2020.9340776}
}

@InProceedings{Xiao2023Learning,
  title = 	 {Learning Model Predictive Controllers with Real-Time Attention for Real-World Navigation},
  author =       {Xiao, Xuesu and Zhang, Tingnan and Choromanski, Krzysztof Marcin and Lee, Tsang-Wei Edward and Francis, Anthony and Varley, Jake and Tu, Stephen and Singh, Sumeet and Xu, Peng and Xia, Fei and Persson, Sven Mikael and Kalashnikov, Dmitry and Takayama, Leila and Frostig, Roy and Tan, Jie and Parada, Carolina and Sindhwani, Vikas},
  booktitle = 	 CoRL,
  pages = 	 {1708--1721},
  year = 	 {2023},
  editor = 	 {Liu, Karen and Kulic, Dana and Ichnowski, Jeff},
  volume = 	 {205},
  series = 	 {Proceedings of Machine Learning Research},
  month = 	 {14--18 Dec},
  publisher =    {PMLR},
}

@article{karunarathne2018model,
	title = {Model of {Side}-by-{Side} {Walking} {Without} the {Robot} {Knowing} the {Goal}},
	volume = {10},
	issn = {1875-4791, 1875-4805},
	doi = {10.1007/s12369-017-0443-6},
	language = {en},
	number = {4},
	urldate = {2023-08-29},
	journal = IJSR,
	author = {Karunarathne, Deneth and Morales, Yoichi and Kanda, Takayuki and Ishiguro, Hiroshi},
	month = sep,
	year = {2018},
	note = {Number: 4},
	pages = {401--420},
}

@article{gross2017roreas,
	title = {{ROREAS}: robot coach for walking and orientation training in clinical post-stroke rehabilitation{\textemdash}prototype implementation and evaluation in field trials},
	volume = {41},
	issn = {0929-5593, 1573-7527},
	shorttitle = {{ROREAS}},
	doi = {10.1007/s10514-016-9552-6},
	language = {en},
	number = {3},
	urldate = {2024-04-10},
	journal = AR,
	author = {Gross, Horst-Michael and Scheidig, Andrea and Debes, Klaus and Einhorn, Erik and Eisenbach, Markus and Mueller, Steffen and Schmiedel, Thomas and Trinh, Thanh Q. and Weinrich, Christoph and Wengefeld, Tim and Bley, Andreas and Martin, Christian},
	month = mar,
	year = {2017},
	note = {Number: 3},
	pages = {679--698},
}

@INPROCEEDINGS{kollmitz2015Time,
  author={Kollmitz, Marina and Hsiao, Kaijen and Gaa, Johannes and Burgard, Wolfram},
  booktitle=ECMR, 
  title={Time dependent planning on a layered social cost map for human-aware robot navigation}, 
  year={2015},
  volume={},
  number={},
  pages={1-6},
  doi={10.1109/ECMR.2015.7324184}
}

@ARTICLE{ye2024reid,
  author={Ye, Hanjing and Zhao, Jieting and Zhan, Yu and Chen, Weinan and He, Li and Zhang, Hong},
  journal=RAL, 
  title={Person Re-Identification for Robot Person Following With Online Continual Learning}, 
  year={2024},
  volume={9},
  number={11},
  pages={9151-9158},
  doi={10.1109/LRA.2024.3438042}
}

@article{socialSurvey2024ijrr,
	title = {A {Survey} on {Socially} {Aware} {Robot} {Navigation}: {Taxonomy} and {Future} {Challenges}},
	issn = {0278-3649, 1741-3176},
	shorttitle = {A {Survey} on {Socially} {Aware} {Robot} {Navigation}},
	doi = {10.1177/02783649241230562},
	language = {en},
	urldate = {2024-04-11},
	journal = IJRR,
	author = {Singamaneni, Phani Teja and Bachiller-Burgos, Pilar and Manso, Luis J. and Garrell, Anaís and Sanfeliu, Alberto and Spalanzani, Anne and Alami, Rachid},
	month = feb,
	year = {2024},
	pages = {02783649241230562},
}

@article{francis2025principles,
  title={Principles and guidelines for evaluating social robot navigation algorithms},
  author={Francis, Anthony and P{\'e}rez-d’Arpino, Claudia and Li, Chengshu and Xia, Fei and Alahi, Alexandre and Alami, Rachid and Bera, Aniket and Biswas, Abhijat and Biswas, Joydeep and Chandra, Rohan and others},
  journal=THRI,
  volume={14},
  number={2},
  pages={1--65},
  year={2025},
  publisher={ACM New York, NY}
}

@inproceedings{hoeller2007iros,
	address = {San Diego, CA, USA},
	title = {Accompanying persons with a mobile robot using motion prediction and probabilistic roadmaps},
	doi = {10.1109/IROS.2007.4399194},
	language = {en},
	urldate = {2023-09-29},
	booktitle = IROS,
	publisher = {IEEE},
	author = {Hoeller, Frank and Schulz, Dirk and Moors, Mark and Schneider, Frank E.},
	month = oct,
	year = {2007},
	pages = {1260--1265},
}

@ARTICLE{han2023rda,
  author={Han, Ruihua and Wang, Shuai and Wang, Shuaijun and Zhang, Zeqing and Zhang, Qianru and Eldar, Yonina C. and Hao, Qi and Pan, Jia},
  journal=RAL, 
  title={RDA: An Accelerated Collision Free Motion Planner for Autonomous Navigation in Cluttered Environments}, 
  year={2023},
  volume={8},
  number={3},
  pages={1715-1722},
  doi={10.1109/LRA.2023.3242138}
}

@inproceedings{chen2017person,
  title={Person following robot using selected online ada-boosting with stereo camera},
  author={Chen, Bao Xin and Sahdev, Raghavender and Tsotsos, John K},
  booktitle={2017 14th conference on computer and robot vision (CRV)},
  pages={48--55},
  year={2017},
  organization={IEEE}
}

@article{zhang2024uni,
  title={Uni-NaVid: A Video-based Vision-Language-Action Model for Unifying Embodied Navigation Tasks},
  author={Zhang, Jiazhao and Wang, Kunyu and Wang, Shaoan and Li, Minghan and Liu, Haoran and Wei, Songlin and Wang, Zhongyuan and Zhang, Zhizheng and Wang, He},
  journal={arXiv preprint arXiv:2412.06224},
  year={2024}
}

@article{puig2023habitat,
  title={Habitat 3.0: A co-habitat for humans, avatars and robots},
  author={Puig, Xavier and Undersander, Eric and Szot, Andrew and Cote, Mikael Dallaire and Yang, Tsung-Yen and Partsey, Ruslan and Desai, Ruta and Clegg, Alexander William and Hlavac, Michal and Min, So Yeon and others},
  journal={arXiv preprint arXiv:2310.13724},
  year={2023}
}

@article{stratton2024characterizing,
  title={Characterizing the Complexity of Social Robot Navigation Scenarios},
  author={Stratton, Andrew and Hauser, Kris and Mavrogiannis, Christoforos},
  journal=RAL,
  year={2024},
  publisher={IEEE}
}

@article{tsoi2022sean2, 
author={Tsoi, Nathan and Xiang, Alec and Yu, Peter and Sohn, Samuel S. and Schwartz, Greg and Ramesh, Subashri and Hussein, Mohamed and Gupta, Anjali W. and Kapadia, Mubbasir and Vázquez, Marynel},
journal=RAL,
title={SEAN 2.0: Formalizing and Generating Social Situations for Robot Navigation},
year={2022},
pages={1-8},
doi={10.1109/LRA.2022.3196783}
}

@article{helbing1995social,
  title={Social force model for pedestrian dynamics},
  author={Helbing, Dirk and Molnar, Peter},
  journal={Physical review E},
  volume={51},
  number={5},
  pages={4282},
  year={1995},
  publisher={APS}
}

@inproceedings{van2011reciprocal,
  title={Reciprocal n-body collision avoidance},
  author={Van Den Berg, Jur and Guy, Stephen J and Lin, Ming and Manocha, Dinesh},
  booktitle={Robotics Research: The 14th International Symposium ISRR},
  pages={3--19},
  year={2011},
  organization={Springer}
}

@inproceedings{gockley2007natural,
  title={Natural person-following behavior for social robots},
  author={Gockley, Rachel and Forlizzi, Jodi and Simmons, Reid},
  booktitle=HRI,
  pages={17--24},
  year={2007}
}

@article{honig2018toward,
  title={Toward socially aware person-following robots},
  author={Honig, Shanee S and Oron-Gilad, Tal and Zaichyk, Hanan and Sarne-Fleischmann, Vardit and Olatunji, Samuel and Edan, Yael},
  journal=TCDS,
  volume={10},
  number={4},
  pages={936--954},
  year={2018},
  publisher={IEEE}
}

@inproceedings{kobayashi2015design,
  title={Design of personal mobility motion based on cooperative movement with a companion},
  author={Kobayashi, Tsukasa and Chugo, Daisuke and Yokota, Sho and Muramatsu, Satoshi and Hashimoto, Hiroshi},
  booktitle=CogInfoCom,
  pages={165--170},
  year={2015},
  organization={IEEE}
}

@inproceedings{martini2024adaptive,
  title={Adaptive Social Force Window Planner with Reinforcement Learning},
  author={Martini, Mauro and P{\'e}rez-Higueras, No{\'e} and Ostuni, Andrea and Chiaberge, Marcello and Caballero, Fernando and Merino, Luis},
  booktitle=IROS,
  pages={4816--4822},
  year={2024},
  organization={IEEE}
}

@inproceedings{repiso2017line,
  title={On-line adaptive side-by-side human robot companion in dynamic urban environments},
  author={Repiso, Ely and Ferrer, Gonzalo and Sanfeliu, Alberto},
  booktitle=IROS,
  pages={872--877},
  year={2017},
  organization={IEEE}
}

@inproceedings{kastner2022human,
  title={Human-following and-guiding in crowded environments using semantic deep-reinforcement-learning for mobile service robots},
  author={K{\"a}stner, Linh and Fatloun, Bassel and Shen, Zhengcheng and Gawrisch, Daniel and Lambrecht, Jens},
  booktitle=ICRA,
  pages={833--839},
  year={2022},
  organization={IEEE}
}

@article{kastner2024arena,
  title={Arena 3.0: Advancing social navigation in collaborative and highly dynamic environments},
  author={K{\"a}stner, Linh and Shcherbyna, Volodymyir and Zeng, Huajian and Le, Tuan Anh and Schreff, Maximilian Ho-Kyoung and Osmaev, Halid and Tran, Nam Truong and Diaz, Diego and Golebiowski, Jan and Soh, Harold and others},
  journal={arXiv preprint arXiv:2406.00837},
  year={2024}
}

@article{chen2022lopf,
  title={LoPF: An online LiDAR-only person-following framework},
  author={Chen, Xiangyu and Liu, Jinhao and Wu, Jin and Wang, Chaoqun and Song, Rui},
  journal=TIM,
  volume={71},
  pages={1--13},
  year={2022},
  publisher={IEEE}
}

@inproceedings{goldhoorn2014continuous,
  title={Continuous real time POMCP to find-and-follow people by a humanoid service robot},
  author={Goldhoorn, Alex and Garrell, Ana{\'\i}s and Alqu{\'e}zar, Ren{\'e} and Sanfeliu, Alberto},
  booktitle=Humanoids,
  pages={741--747},
  year={2014},
  organization={IEEE}
}

@inproceedings{liu2019robot,
  title={Robot comfort following based on extended social force model in dynamic environment},
  author={Liu, Meng and Zhou, Lei and Sun, Yue and Liu, Jingtai},
  booktitle=CYBER,
  pages={30--35},
  year={2019},
  organization={IEEE}
}

@inproceedings{ferrer2014proactive,
  title={Proactive kinodynamic planning using the extended social force model and human motion prediction in urban environments},
  author={Ferrer, Gonzalo and Sanfeliu, Alberto},
  booktitle=IROS,
  pages={1730--1735},
  year={2014},
  organization={IEEE}
}

@article{samavi2024sicnav,
  title={SICNav: Safe and Interactive Crowd Navigation using Model Predictive Control and Bilevel Optimization},
  author={Samavi, Sepehr and Han, James R and Shkurti, Florian and Schoellig, Angela P},
  journal=TRO,
  year={2024},
  publisher={IEEE}
}

@article{eirale2025human,
  title={Human following and guidance by autonomous mobile robots: A comprehensive review},
  author={Eirale, Andrea and Martini, Mauro and Chiaberge, Marcello},
  journal={IEEE Access},
  year={2025},
  publisher={IEEE}
}

@article{hall1963system,
  title={A system for the notation of proxemic behavior},
  author={Hall, Edward T},
  journal={American anthropologist},
  volume={65},
  number={5},
  pages={1003--1026},
  year={1963},
  publisher={JSTOR}
}

@inproceedings{repiso2019people,
  title={People's v-formation and side-by-side model adapted to accompany groups of people by social robots},
  author={Repiso, Ely and Zanlungo, Francesco and Kanda, Takayuki and Garrell, Ana{\'\i}s and Sanfeliu, Alberto},
  booktitle=IROS,
  pages={2082--2088},
  year={2019},
  organization={IEEE}
}

@article{koide2016reid,
title = {Identification of a specific person using color, height, and gait features for a person following robot},
journal = RAS,
volume = {84},
pages = {76-87},
year = {2016},
issn = {0921-8890},
author = {K. Koide and J. Miura},
}

@article{cai2022human,
  title={Human-aware path planning with improved virtual doppler method in highly dynamic environments},
  author={Cai, Kuanqi and Chen, Weinan and Wang, Chaoqun and Song, Shuang and Meng, Max Q-H},
  journal=TASE,
  volume={20},
  number={2},
  pages={1304--1321},
  year={2022},
  publisher={IEEE}
}

@inproceedings{chen2017socially,
  title={Socially aware motion planning with deep reinforcement learning},
  author={Chen, Yu Fan and Everett, Michael and Liu, Miao and How, Jonathan P},
  booktitle=IROS,
  pages={1343--1350},
  year={2017},
  organization={IEEE}
}

@article{goldhoorn2018searching,
  title={Searching and tracking people with cooperative mobile robots},
  author={Goldhoorn, Alex and Garrell, Ana{\'\i}s and Alqu{\'e}zar, Ren{\'e} and Sanfeliu, Alberto},
  journal=AR,
  volume={42},
  number={4},
  pages={739--759},
  year={2018},
  publisher={Springer}
}

@article{goldhoorn2017searching,
  title={Searching and tracking people in urban environments with static and dynamic obstacles},
  author={Goldhoorn, Alex and Garrell, Ana{\'\i}s and Alqu{\'e}zar, Ren{\'e} and Sanfeliu, Alberto},
  journal=RAS,
  volume={98},
  pages={147--157},
  year={2017},
  publisher={Elsevier}
}

@inproceedings{barnaud2014proxemics,
  title={Proxemics models for human-aware navigation in robotics: Grounding interaction and personal space models in experimental data from psychology},
  author={Barnaud, Marie-Lou and Morgado, Nicolas and Palluel-Germain, Richard and Diard, Julien and Spalanzani, Anne},
  booktitle={Proc. 3rd IROS Workshop Assist. Serv. Robot. Human Environ. (IROS Workshop)},
  year={2014}
}

@article{hayduk1994personal,
  title={Personal space: Understanding the simplex model},
  author={Hayduk, Leslie A},
  journal={Journal of Nonverbal Behavior},
  volume={18},
  number={3},
  pages={245--260},
  year={1994},
  publisher={Springer}
}

@article{ferrer2017robot,
  title={Robot social-aware navigation framework to accompany people walking side-by-side},
  author={Ferrer, Gonzalo and Zulueta, Ana{\'\i}s Garrell and Cotarelo, Fernando Herrero and Sanfeliu, Alberto},
  journal=AR,
  volume={41},
  number={4},
  pages={775--793},
  year={2017},
  publisher={Springer}
}

@article{bayoumi2019speeding,
  title={Speeding up person finding using hidden Markov models},
  author={Bayoumi, AbdElMoniem and Karkowski, Philipp and Bennewitz, Maren},
  journal=RAS,
  volume={115},
  pages={40--48},
  year={2019},
  publisher={Elsevier}
}

@article{bruckschen2020predicting,
  title={Predicting human navigation goals based on Bayesian inference and activity regions},
  author={Bruckschen, Lilli and Bungert, Kira and Dengler, Nils and Bennewitz, Maren},
  journal=RAS,
  volume={134},
  pages={103664},
  year={2020},
  publisher={Elsevier}
}

@article{liu2023close,
  title={Close-range human following control on a cane-type robot with multi-camera fusion},
  author={Liu, Haowen and Wu, Fengxian and Zhong, Bin and Zhao, Yijun and Zhang, Jiatong and Niu, Wenxin and Zhang, Mingming},
  journal=RAL,
  volume={8},
  number={10},
  pages={6443--6450},
  year={2023},
  publisher={IEEE}
}

@inproceedings{wang2024continuous,
  title={Continuous Robotic Tracking of Dynamic Targets in Complex Environments Based on Detectability},
  author={Wang, Zhihao and Huang, Shixing and Li, Minghang and Ouyang, Junyuan and Wang, Yu and Chen, Haoyao},
  booktitle=ICRA,
  pages={16338--16344},
  year={2024},
  organization={IEEE}
}

@ARTICLE{yuan2025taes,
  author={Yuan, Shengze and Gao, Feiyu and Wang, Yiyu and Fabiani, Filippo and Yuan, Shuai},
  journal=TAES, 
  title={Path Planning for Lunar Surface Person Following Robot via Flatness-Based Safe-MPC with Virtual Disturbances}, 
  year={2025},
  pages={1-12},
  doi={10.1109/TAES.2025.3575050}
}

@article{andersson2019casadi,
  title={CasADi: a software framework for nonlinear optimization and optimal control},
  author={Andersson, Joel AE and Gillis, Joris and Horn, Greg and Rawlings, James B and Diehl, Moritz},
  journal={Math. Program. Comput.},
  volume={11},
  number={1},
  pages={1--36},
  year={2019},
  publisher={Springer}
}

@ARTICLE{jin2020access,
  author={Jin, Daping and Fang, Zheng and Zeng, Jiexin},
  journal={IEEE Access}, 
  title={A Robust Autonomous Following Method for Mobile Robots in Dynamic Environments}, 
  year={2020},
  volume={8},
  pages={150311-150325},
  doi={10.1109/ACCESS.2020.3016472}
}

@inproceedings{dewantara2016generation,
  title={Generation of a socially aware behavior of a guide robot using reinforcement learning},
  author={Dewantara, Bima Sena Bayu and Miura, Jun},
  booktitle=IES,
  pages={105--110},
  year={2016},
  organization={IEEE}
}

@INPROCEEDINGS{yao2021iros,
  author={Yao, Hanchen and Dai, Houde and Zhao, Enhao and Liu, Penghua and Zhao, Ran},
  booktitle=IROS, 
  title={Laser-Based Side-by-Side Following for Human-Following Robots}, 
  year={2021},
  pages={2651-2656},
  doi={10.1109/IROS51168.2021.9636458}
}

@inproceedings{chen2017integrating,
  title={Integrating stereo vision with a CNN tracker for a person-following robot},
  author={Chen, Bao Xin and Sahdev, Raghavender and Tsotsos, John K},
  booktitle=ICCVS,
  pages={300--313},
  year={2017},
  organization={Springer}
}

@INPROCEEDINGS{lee2018icra,
  author={Lee, Beom-Jin and Choi, Jinyoung and Baek, Christina and Zhang, Byoung-Tak},
  booktitle=ICRA, 
  title={Robust Human Following by Deep Bayesian Trajectory Prediction for Home Service Robots}, 
  year={2018},
  pages={7189-7195},
  doi={10.1109/ICRA.2018.8462969}
}

@article{kim2018architecture,
  title={An architecture for person-following using active target search},
  author={Kim, Minkyu and Arduengo, Miguel and Walker, Nick and Jiang, Yuqian and Hart, Justin W and Stone, Peter and Sentis, Luis},
  journal={arXiv preprint arXiv:1809.08793},
  year={2018}
}

@article{hu2013design,
  title={Design of sensing system and anticipative behavior for human following of mobile robots},
  author={Hu, Jwu-Sheng and Wang, Jyun-Ji and Ho, Daniel Minare},
  journal=TIE,
  volume={61},
  number={4},
  pages={1916--1927},
  year={2013},
  publisher={IEEE}
}

@inproceedings{gerkey2008planning,
  title={Planning and control in unstructured terrain},
  author={Gerkey, Brian P and Konolige, Kurt},
  booktitle={Proc. ICRA Workshop Path Planning on Costmaps},
  year={2008},
  organization={Citeseer}
}

@inproceedings{chen2019crowd,
  title={Crowd-robot interaction: Crowd-aware robot navigation with attention-based deep reinforcement learning},
  author={Chen, Changan and Liu, Yuejiang and Kreiss, Sven and Alahi, Alexandre},
  booktitle=ICRA,
  pages={6015--6022},
  year={2019},
  organization={IEEE}
}

@INPROCEEDINGS{mun2023icra,
  author={Mun, Ye-Ji and Itkina, Masha and Liu, Shuijing and Driggs-Campbell, Katherine},
  booktitle=ICRA, 
  title={Occlusion-Aware Crowd Navigation Using People as Sensors}, 
  year={2023},
  volume={},
  number={},
  pages={12031-12037},
  doi={10.1109/ICRA48891.2023.10160715}
}

@inproceedings{lee2018robust,
  title={Robust human following by deep bayesian trajectory prediction for home service robots},
  author={Lee, Beom-Jin and Choi, Jinyoung and Baek, Christina and Zhang, Byoung-Tak},
  booktitle=ICRA,
  pages={7189--7195},
  year={2018},
  organization={IEEE}
}

@inproceedings{do2017reliable,
  title={The reliable recovery mechanism for person-following robot in case of missing target},
  author={Do Hoang, Minh and Yun, Sang-Seok and Choi, Jong-Suk},
  booktitle=URAI,
  pages={800--803},
  year={2017},
  organization={IEEE}
}

@article{williams2018information,
  title={Information-theoretic model predictive control: Theory and applications to autonomous driving},
  author={Williams, Grady and Drews, Paul and Goldfain, Brian and Rehg, James M and Theodorou, Evangelos A},
  journal=TRO,
  volume={34},
  number={6},
  pages={1603--1622},
  year={2018},
  publisher={IEEE}
}

@inproceedings{ester1996density,
  title={A density-based algorithm for discovering clusters in large spatial databases with noise},
  author={Ester, Martin and Kriegel, Hans-Peter and Sander, J{\"o}rg and Xu, Xiaowei and others},
  booktitle={Proc. ACM SIGKDD Int. Conf. Knowl. Discov. Data Min. (KDD)},
  volume={96},
  number={34},
  pages={226--231},
  year={1996}
}

@inproceedings{ryu2024integrating,
  title={Integrating predictive motion uncertainties with distributionally robust risk-aware control for safe robot navigation in crowds},
  author={Ryu, Kanghyun and Mehr, Negar},
  booktitle=ICRA,
  pages={2410--2417},
  year={2024},
  organization={IEEE}
}

@inproceedings{gupta2018social,
  title={Social gan: Socially acceptable trajectories with generative adversarial networks},
  author={Gupta, Agrim and Johnson, Justin and Fei-Fei, Li and Savarese, Silvio and Alahi, Alexandre},
  booktitle=CVPR,
  pages={2255--2264},
  year={2018}
}

@inproceedings{dugas2022flowbot,
  title={FlowBot: Flow-based modeling for robot navigation},
  author={Dugas, Daniel and Cai, Kuanqi and Andersson, Olov and Lawrance, Nicholas and Siegwart, Roland and Chung, Jen Jen},
  booktitle=IROS,
  pages={8799--8805},
  year={2022},
  organization={IEEE}
}

@article{sun2025socially,
  title={Socially Aware Robot Crowd Navigation via Online Uncertainty-Driven Risk Adaptation},
  author={Sun, Zhirui and Diao, Xingrong and Wang, Yao and Zhu, Bi-Ke and Wang, Jiankun},
  journal={arXiv preprint arXiv:2506.14305},
  year={2025}
}

@software{NVIDIAIsaacSim,
author = {{NVIDIA}},
license = {Apache-2.0},
title = {{Isaac Sim}},
url = {https://github.com/isaac-sim/IsaacSim},
version = {5.0.0}
}

@article{zhan2025monocular,
  title={Monocular Person Localization under Camera Ego-motion},
  author={Zhan, Yu and Ye, Hanjing and Zhang, Hong},
  journal={arXiv preprint arXiv:2503.02916},
  year={2025}
}

@INPROCEEDINGS{sun2016ROBIO,
  author={Sun, Yue and Sun, Lei and Liu, Jingtai},
  booktitle={2016 IEEE International Conference on Robotics and Biomimetics (ROBIO)}, 
  title={Human comfort following behavior for service robots}, 
  year={2016},
  volume={},
  number={},
  pages={649-654},
  doi={10.1109/ROBIO.2016.7866396}
}

@inproceedings{pellegrini2009you,
  title={You'll never walk alone: Modeling social behavior for multi-target tracking},
  author={Pellegrini, Stefano and Ess, Andreas and Schindler, Konrad and Van Gool, Luc},
  booktitle=ICCV,
  pages={261--268},
  year={2009},
  organization={IEEE}
}

@article{chandra2025multi,
  title={Multi-Robot Navigation in Social Mini-Games: Definitions, Taxonomy, and Algorithms},
  author={Chandra, Rohan and Singh, Shubham and Jha, Abhishek and Andrade, Dannon and Sainathuni, Hriday and Sycara, Katia},
  journal={arXiv preprint arXiv:2508.13459},
  year={2025}
}

@article{scheidemann2024obstacle,
  title={Obstacle-avoidant leader following with a quadruped robot},
  author={Scheidemann, Carmen and Werner, Lennart and Reijgwart, Victor and Cramariuc, Andrei and Chomarat, Joris and Chiu, Jia-Ruei and Siegwart, Roland and Hutter, Marco},
  journal={arXiv preprint arXiv:2410.00572},
  year={2024}
}

@article{xu2021fast,
  title={Fast-lio: A fast, robust lidar-inertial odometry package by tightly-coupled iterated kalman filter},
  author={Xu, Wei and Zhang, Fu},
  journal=RAL,
  volume={6},
  number={2},
  pages={3317--3324},
  year={2021},
  publisher={IEEE}
}

@INPROCEEDINGS{peng2023iros,
  author={Peng, Jianwei and Liao, Zhelin and Yao, Hanchen and Su, Zefan and Zeng, Yadan and Dai, Houde},
  booktitle=IROS, 
  title={MPC-Based Human-Accompanying Control Strategy for Improving the Motion Coordination Between the Target Person and the Robot}, 
  year={2023},
  volume={},
  number={},
  pages={7969-7975}
}

@article{tan2025robust,
  title={Robust side following robotic wheelchair by using homotopy class of human intention},
  author={Tan, Kuan Yuee and Garg, Neha P and Ramanathan, Manoj and Ang, Wei Tech},
  journal=RAL,
  year={2025},
  publisher={IEEE}
}

@inproceedings{zhang2023collision,
  title={A Collision-Free Pedestrian-Following System for Mobile Robots Base on Path Planning},
  author={Zhang, Hong and Wang, Songyan and Huo, Ju and Chao, Tao},
  booktitle=ICMA,
  pages={1853--1858},
  year={2023},
  organization={IEEE}
}

@inproceedings{ji2022elastic,
  title={Elastic tracker: A spatio-temporal trajectory planner for flexible aerial tracking},
  author={Ji, Jialin and Pan, Neng and Xu, Chao and Gao, Fei},
  booktitle=ICRA,
  pages={47--53},
  year={2022},
  organization={IEEE}
}

@article{zhu2024efficient,
  title={An efficient and robust system for human following scenario using differential robot},
  author={Zhu, Jiangchao and Ma, Changjia and Xu, Chao and Gao, Fei},
  journal={IET Cyber-Syst. Robot.},
  volume={6},
  number={1},
  pages={e12108},
  year={2024},
  publisher={Wiley Online Library}
}

@INPROCEEDINGS{liu2023icra,
  author={Liu, Shuijing and Chang, Peixin and Huang, Zhe and Chakraborty, Neeloy and Hong, Kaiwen and Liang, Weihang and McPherson, D. Livingston and Geng, Junyi and Driggs-Campbell, Katherine},
  booktitle=ICRA, 
  title={Intention Aware Robot Crowd Navigation with Attention-Based Interaction Graph}, 
  year={2023},
  volume={},
  number={},
  pages={12015-12021}
}

@inproceedings{wang2022metrics,
  title={Metrics for evaluating social conformity of crowd navigation algorithms},
  author={Wang, Junxian and Chan, Wesley P and Carreno-Medrano, Pamela and Cosgun, Akansel and Croft, Elizabeth},
  booktitle=ARSO,
  pages={1--6},
  year={2022},
  organization={IEEE}
}

@article{lyu2025robust,
  title={A robust human-following system for autonomous mobile robot in unknown environments},
  author={Lyu, Haojie and Wu, Wei},
  journal={IEEE Sensors Journal},
  year={2025},
  publisher={IEEE}
}

@article{wu2026human,
  title={Human-following control of an autonomous agricultural vehicle using an adaptive navigation point time elastic band method},
  author={Wu, Yihao and He, Yong and Cui, Bingbo and Li, Feng and Hu, Anguo and Feng, Lei and Liu, Yufei},
  journal={Information Processing in Agriculture},
  year={2026},
  publisher={Elsevier}
}

@article{wang2022geometrically,
  title={Geometrically constrained trajectory optimization for multicopters},
  author={Wang, Zhepei and Zhou, Xin and Xu, Chao and Gao, Fei},
  journal=TRO,
  volume={38},
  number={5},
  pages={3259--3278},
  year={2022},
  publisher={IEEE}
}

@inproceedings{lin2024safety,
  title={Safety-first tracker: A trajectory planning framework for omnidirectional robot tracking},
  author={Lin, Yue and Liu, Yang and Zhang, Pingping and Chen, Xin and Wang, Dong and Lu, Huchuan},
  booktitle=IROS,
  pages={5416--5423},
  year={2024},
  organization={IEEE}
}

@inproceedings{murakami2014destination,
  title={Destination unknown: walking side-by-side without knowing the goal},
  author={Murakami, Ryo and Morales Saiki, Luis Yoichi and Satake, Satoru and Kanda, Takayuki and Ishiguro, Hiroshi},
  booktitle=HRI,
  pages={471--478},
  year={2014}
}

@article{jevtic2015comparison,
  title={Comparison of interaction modalities for mobile indoor robot guidance: Direct physical interaction, person following, and pointing control},
  author={Jevti{\'c}, Aleksandar and Doisy, Guillaume and Parmet, Yisrael and Edan, Yael},
  journal=THMS,
  volume={45},
  number={6},
  pages={653--663},
  year={2015},
  publisher={IEEE}
}

@inproceedings{shanee2016influence,
  title={The influence of following angle on performance metrics of a human-following robot},
  author={Shanee, Honig S and Dror, Katz and Tal, Oron-Gilad and Yael, Edan},
  booktitle=RO-MAN,
  pages={593--598},
  year={2016},
  organization={IEEE}
}

@article{liu2025trackvla++,
  title={Trackvla++: Unleashing reasoning and memory capabilities in vla models for embodied visual tracking},
  author={Liu, Jiahang and Qi, Yunpeng and Zhang, Jiazhao and Li, Minghan and Wang, Shaoan and Wu, Kui and Ye, Hanjing and Zhang, Hong and Chen, Zhibo and Zhong, Fangwei and others},
  journal={arXiv preprint arXiv:2510.07134},
  year={2025}
}

@inproceedings{wang2025trackvla,
  title={TrackVLA: Embodied Visual Tracking in the Wild},
  author={Wang, Shaoan and Zhang, Jiazhao and Li, Minghan and Liu, Jiahang and Li, Anqi and Wu, Kui and Zhong, Fangwei and Yu, Junzhi and Zhang, Zhizheng and Wang, He},
  booktitle={Conference on Robot Learning},
  pages={4139--4164},
  year={2025},
  organization={PMLR}
}

@article{ye2026refertrack,
  title={{ReferTrack}: Referring Then Tracking for Embodied Visual Tracking},
  author={Ye, Hanjing and Zeng, Tianle and Zhang, Jiazhao and Wang, Shaoan and Zhang, Zibo and Situ, Weixi and Zhou, Yuchen and Ling, Yonggen and Zhang, Hong},
  journal={arXiv preprint arXiv:2607.20061},
  year={2026},
  url={https://arxiv.org/abs/2607.20061}
}
\clearpage
\section*{APPENDIX}
\subsection{Baselines and Simulation Settings}
Follow-Bench evaluates eight implementations derived from five core planners and three composite variants under a unified RPF wrapper (1.5\,m default following distance, adaptive following-position generation, simulator-level visibility checking, and Kalman-filter-based fallback search). Table~\ref{tab:plannerProvenance} records the cited source and retained core mechanism of each implementation, together with benchmark-level additions and method-specific adaptations. Table~\ref{tab:appendixSimulator} summarizes the shared simulator and pedestrian-interaction settings (ORCA~\cite{van2011reciprocal} and SFM~\cite{helbing1995social}), while Table~\ref{tab:appendixHyperparams} and the released scripts and YAML files record the complete final Follow-Bench configurations. Because most RPF baselines do not provide complete open-source implementations and several low-level parameters are either omitted or coupled to different robot platforms, we treat each re-implementation as a complete Follow-Bench configuration rather than an exact reproduction of every original numerical setting.

\subsubsection{Hyper-parameter Selection Protocol}
Platform and task constants ($\delta t=0.1$\,s, robot footprint and kinematic bounds, $5.0$\,m LiDAR range, and default $1.5$\,m following distance) were fixed uniformly. Planner-specific parameters were initialized from the cited papers and available implementations whenever possible; unavailable parameters and those requiring adaptation to the unified interface were manually calibrated in preliminary development runs covering representative obstacle-free target trajectories, dynamic-crowd interactions, and static environmental layouts. Stable target tracking, collision avoidance, and execution within the benchmark control cycle were the common selection criteria. The final 100-trial evaluation outcomes were not used for parameter selection: one configuration per planner was frozen beforehand and then applied unchanged across all 16 scenario types, densities, occupancy levels, following distances, and following angles, without per-scenario re-tuning.

\begin{table*}[t]
    
    \centering
    \caption{Provenance of the evaluated Follow-Bench implementations. All methods share the unified RPF wrapper in Sec.~\ref{sec:benchBaselines} (preferred distance, adaptive following position, visibility check, and KF fallback search).}
    \label{tab:plannerProvenance}
    \scriptsize
    \renewcommand{\arraystretch}{1.15}
    \begin{tabular}{p{2.3cm} p{5.2cm} p{8.4cm}}
    \toprule
    \textbf{Implementation} & \textbf{Retained core/source} & \textbf{Follow-Bench adaptation and scope} \\
    \midrule
    \textbf{SFM} &
    Social-force local planner~\cite{ferrer2017robot}. &
    Re-implemented under the unified interface with EDT-based obstacle repulsion. \\
    \midrule
    \textbf{MPC} &
    RDA receding-horizon solver and collision formulation~\cite{han2023rda}. &
    Uses RDA as the local planning core under the shared RPF wrapper. \\
    \midrule
    \textbf{MPC w/ Traj.} &
    RDA backbone~\cite{han2023rda}; trajectory tracking and uncertainty braking inspired by~\cite{sekiguchi2021uncertainty}. &
    Composite variant integrating CVKF target-trajectory prediction and uncertainty-triggered braking. \\
    \midrule
    \textbf{MPC w/ DS.} &
    RDA backbone~\cite{han2023rda} and dynamic search field~\cite{ye2025rpf}. &
    Composite variant invoking overtaking or fluid-following search after occlusion. \\
    \midrule
    \textbf{DWA} &
    Dynamic-window velocity sampling~\cite{van2022collision}. &
    Re-implemented under the unified interface using point-cloud obstacle inputs. \\
    \midrule
    \textbf{DWA w/ Traj.} &
    DWA core~\cite{van2022collision} with the CVKF predictor. &
    Composite variant adding multi-step rollout and predicted-trajectory tracking costs. \\
    \midrule
    \textbf{BSO-HFC} &
    Hybrid-A*, B-spline optimization, and MPC tracking~\cite{lyu2025robust}. &
    Re-implemented with a local occupancy grid under the shared RPF wrapper. \\
    \midrule
    \textbf{RL-based} &
    Human-action-aware MCTS planner~\cite{leisiazar2025adapting}. &
    Extended to back- and side-following via a human-centered formulation and newly trained A2C value networks. \\
    \bottomrule
    \end{tabular}
\end{table*}

\begin{table*}[t]
    \centering
    \caption{Common simulator settings and pedestrian-simulation parameters used in Follow-Bench scenario generation.}
    \label{tab:appendixSimulator}
    \scalebox{0.96}{
    \begin{tabular}{p{3.2cm} p{13.2cm}}
    \toprule
    \textbf{Component} & \textbf{Settings (Follow-Bench)} \\
    \midrule
    \textbf{Common simulator} &
    World and control step: $0.1$\,s. Differential-drive robot footprint: $0.7 \times 0.55$\,m rectangle with wheelbase $0.5$\,m.
    2D LiDAR range: $5.0$\,m, with $200$ beams over $360^\circ$. Occupancy-map resolution: $0.1$\,m.
    Humans and the target are modeled as omni-directional disks of radius $0.3$\,m with $|v_x|, |v_y| \le 1.5$\,m/s.
    Target search falls back to the Kalman-predicted target position with search timeout $0.5N_{\text{min}}$.
    Static obstacles are extracted from 2D LiDAR using DBSCAN clustering and minimum-area rectangle fitting. \\
    \midrule
    \textbf{Pedestrian simulation (ORCA)} &
    $v_x^{\max}=v_y^{\max}=1.5$\,m/s, with acceleration limit $1.0$\,m/s$^2$.
    Desired \texttt{safe\_dist}: $=0.4$ m for layout generators and $0.5$ m for dynamic-crowd scenarios. \\
    \midrule
    \textbf{Pedestrian simulation (SFM)} &
    Same speed and acceleration limits as ORCA.
    Crowd-side interaction gains: $\eta_{\mathrm{trk}}=1.5$, $\eta_{\mathrm{sa}}=5.0$, $\eta_{\mathrm{sb}}=0.25$,
    $\eta_{\mathrm{ha}}=7.0$, $\eta_{\mathrm{hb}}=0.3$, $\eta_{\mathrm{cut}}=5.0$\,m, and
    $\eta_{\mathrm{safe}}=0.5$\,m. Symbol meanings are aligned with Alg.~\ref{alg:sfm}. \\
    \bottomrule
    \end{tabular}
    }
\end{table*}

\begin{table*}[t]
    \centering
    \caption{Final planner hyper-parameters corresponding to Algorithms~\ref{alg:sfm}--\ref{alg:rl} in the released Follow-Bench scripts.}
    \label{tab:appendixHyperparams}
    \scalebox{0.96}{
    \begin{tabular}{p{3.2cm} p{13.2cm}}
    \toprule
    \textbf{Planner (Alg.)} & \textbf{Final Follow-Bench hyper-parameters} \\
    \midrule
    \textbf{SFM (Alg.~\ref{alg:sfm})} &
    $\delta t=0.1$\,s; $v_{\max}=3.0$\,m/s; $\eta_{\mathrm{trk}}=1.3$; $\eta_{\mathrm{sa}}=10$, $\eta_{\mathrm{sb}}=0.5$; $\eta_{\mathrm{ha}}=7$, $\eta_{\mathrm{hb}}=0.5$; $\eta_{\mathrm{cut}}=5.0$\,m; $\eta_{\mathrm{safe}}=0.3$\,m; and $\eta_{\mathrm{wa}}=\eta_{\mathrm{wr}}=1.0$. \\
    \midrule
    \textbf{MPC (Alg.~\ref{alg:mpc})} &
    $N_{\mathrm{hor}}=10$ steps at $\delta t=0.1$\,s; $d_{\mathrm{safe}}\in[0.1,1.0]$\,m (nominal $0.5$\,m); inner residual tolerance $\epsilon=0.2$; maximum inner iterations $K=4$; state weight $w_s=10.0$; velocity weight $w_u=1.0$; and slack penalty gain $10$. \\
    \midrule
    \textbf{MPC w/ Traj. (Alg.~\ref{alg:mpcTraj})} &
    $d_{\mathrm{safe}}$, $\epsilon$, and $K$ as in Alg.~\ref{alg:mpc}; CVKF horizon $2.0$\,s ($N_{\mathrm{hor}}=20$ at $\delta t=0.1$\,s), history length $1$, and \texttt{num\_samples}$=20$; NIS braking threshold $\gamma=4.0$; $w_s=10.0$, $w_u=2.0$, $w_o=1.0$, and $\sigma_o=0.5$. \\
    \midrule
    \textbf{MPC w/ DS (Alg.~\ref{alg:mpcDS})} &
    $N_{\mathrm{hor}}$, $d_{\mathrm{safe}}$, $\epsilon$, and $K$ as in Alg.~\ref{alg:mpc}; prediction horizon $2.0$\,s; fluid-mode $\gamma=0.1$ and density kernel $\sigma=1.0$; switching threshold $\delta_c=3.5$; occlusion IoU threshold $0.3$; and $20/30$ overtaking/fluid candidate samples. \\
    \midrule
    \textbf{DWA (Alg.~\ref{alg:dwa})} &
    One-step \textsf{ForwardSim} at $\delta t=0.1$\,s; obstacle inflation radius $0.6$\,m; velocity sampling resolution scale $(0.01,0.01)$; and heading, obstacle, and velocity weights $(3.0,0.1,1.0)$. \\
    \midrule
    \textbf{DWA w/ Traj. (Alg.~\ref{alg:dwaTraj})} &
    $2.0$\,s rollout and CVKF horizon ($N_{\mathrm{hor}}=20$ at $\delta t=0.1$\,s); obstacle inflation radius $0.3$\,m; sampling resolution scale $(0.02,0.02)$; and heading, trajectory-distance, obstacle, and velocity weights $(3.0,2.0,1.0,0.5)$. \\
    \midrule
    \textbf{BSO-HFC (Alg.~\ref{alg:bsohfc})} &
    Shared $1.5$\,m following distance and $0.1$\,s control step; $0.1$\,m local LiDAR grid; cubic B-spline ($p=3$), $10$ control points, and $0.15$\,m minimum spacing; Hybrid A* step $0.2$\,m, heading resolution $15^\circ$, and max expansions $900$; L-BFGS-B max iterations $60$; $d_{\mathrm{thr}}=4.0$\,m; $(\omega_c,\omega_g,\omega_s,\omega_d)=(40.0,1.5,20.0,0.02)$; $v_{\max}=2.0$\,m/s, $a_{\max}=1.0$\,m/s$^2$; PID gains $(16,8,2)$; and MPC horizon $N_{\mathrm{hor}}=20$. \\
    \midrule
    \textbf{RL-based (Alg.~\ref{alg:rl})} &
    MCTS simulation step $\delta t=0.2$\,s; search budget $\tau=0.15$\,s per replan; discount $\gamma=0.9$; robot/human turn angles $45^\circ/10^\circ$; nominal robot/human speed $0.6$\,m/s; fast-action multiplier $1.5$; and A2C value models trained for back- and side-following. \\
    \bottomrule
    \end{tabular}
    }
\end{table*}

\begin{algorithm}[h]
    \DontPrintSemicolon
    \SetNoFillComment
    \footnotesize
    \KwIn{2D LiDAR scan $\mathcal{S}$, Target person's historical trajectory $\mathcal{H}_t$, Robot pose $\mathbf{p}^{R}_{t}$, velocity $\mathbf{v}^{R}_{t}$, Target position $\mathbf{p}^{T}_{t}$, Other people's positions $\{\mathbf{p}^H_{i,t}\}^N_{i=1}$, Preferred person-following orientation $\theta$ and distance $d$, Maximum speed $v_{\text{max}}$, Step $\delta{t}$, Hyper-parameters $\eta_{\mathrm{trk}}$, $\eta_{\mathrm{sa}}$, $\eta_{\mathrm{sb}}$, $\eta_{\mathrm{ha}}$, $\eta_{\mathrm{hb}}$, $\eta_{\mathrm{cut}}$, $\eta_{\mathrm{safe}}$, $\eta_{\mathrm{wa}}$, $\eta_{\mathrm{wr}}$}
    \KwOut{Velocity $\mathbf{v}^{*}_t$}
    \If{Target visible}{
        $\mathbf{p}^{G}_{t}=\mathbf{p}^{T}_{t}-d[\cos(\theta), \sin(\theta)]$;\\
    }
    \Else{
        $\mathbf{p}^{G}_{t} = KalmanPredict(\mathcal{H}_t)$;\\
    }
    Obstacles $\mathcal{O}_t=ObstacleDetection(\mathcal{S})$;\\
    Attractive force $\mathbf{F}_a = CalAttractive(\mathbf{p}^{G}_{t}, \mathbf{p}^{R}_{t}, \mathbf{v}^{R}_{t}, v_{\text{max}}, \eta_{\mathrm{trk}})$;\\
    Interaction force $\mathbf{F}_i = CalInteraction(\{\mathbf{p}^H_{i,t}\}^N_{i=1}, \mathbf{p}^{R}_{t}, \eta_{\mathrm{ha}}, \eta_{\mathrm{hb}}, \eta_{\mathrm{safe}})$;\\
    Repulsive force $\mathbf{F}_r = CalRepulsive(\mathcal{O}_t, \mathbf{p}^{R}_{t}, \eta_{\mathrm{sa}}, \eta_{\mathrm{sb}}, \eta_{\mathrm{cut}}, \eta_{\mathrm{safe}})$;\\

    $\mathbf{v}^{*}_t=\big(\eta_{\mathrm{wa}} \mathbf{F}_a + \eta_{\mathrm{wr}} (\mathbf{F}_i + \mathbf{F}_r)\big) * \delta{t}$

    $\mathbf{v}^{*}_t=v_{\text{max}}*\mathbf{v}^{*}/||\mathbf{v}^{*}||$

    \textbf{Return} $\mathbf{v}^{*}_t$
\caption{SFM-based planner}
\label{alg:sfm}
\end{algorithm}

\begin{algorithm}[h]
    \DontPrintSemicolon
    \SetNoFillComment
    \footnotesize
    \KwIn{2D LiDAR scan $\mathcal{S}$, Target person's historical trajectory $\mathcal{H}_t$, Robot state $\mathbf{s}_{t}$, Target position $\mathbf{p}^{T}_{t}$, Other people's positions $\{\mathbf{p}^H_{i,t}\}^N_{i=1}$, Preferred person-following orientation $\theta$ and distance $d$, Robot maximum velocity $v_{\text{max}}$, Time step $\delta{t}$, MPC Horizon $N_{\text{hor}}$, Safe distance $d_{\text{safe}}$, Stopping criteria value $\epsilon$}
    \KwOut{Velocity $\mathbf{v}^{*}_t$}
    \If{Target visible}{
        $\mathbf{p}^{G}_{t}=\mathbf{p}^{T}_{t}-d[\cos(\theta), \sin(\theta)]$;\\
    }
    \Else{
        $\mathbf{p}^{G}_{t} = KalmanPredict(\mathcal{H}_t)$;\\
    }
    Reference path $\mathcal{P}_t = \begin{bmatrix}
                                        \mathbf{p}^G_t & \mathbf{p}^G_t & \cdots & \mathbf{p}^G_t
                                        \end{bmatrix}
                                        \in \mathbb{R}^{2 \times N_{\text{hor}}}$;\\
    Obstacles $\mathcal{O}_t=ObstacleDetection(\mathcal{S})$;\\
    $C_{\text{track}}=TrackCost(\{\mathbf{v}_t, \mathbf{s}_t\}, \mathcal{P}_t, v_{\text{max}})$;\\
    $C_{\text{collision}}=CollisionCost(\{\mathbf{s}_t\}, \mathcal{O}_t, \{\mathbf{p}^H_{i,t}\}^N_{i=1},d_{\text{safe}})$;\\
    Bi-convex optimization $\mathcal{L}=Formulate(C_{\text{track}}, C_{\text{collision}})$;\\
    \For{iteration $k=\{1,2,\cdots,K\}$}{
        $\mathbf{v}^{*}_t, \textit{residual}= Solve(\mathcal{L})$;\\
        \If{$\textit{residual} < \epsilon$}{
            \textbf{break};
        }
    }
    \textbf{Return} $\mathbf{v}^{*}_t$
\caption{MPC-based planner}
\label{alg:mpc}
\end{algorithm}

\begin{algorithm}[t]
    \DontPrintSemicolon
    \SetNoFillComment
    \footnotesize
    \KwIn{2D LiDAR scan $\mathcal{S}$, Target person's historical trajectory $\mathcal{H}_t$, Robot state $\mathbf{s}_{t}$, Target position $\mathbf{p}^{T}_{t}$, Other people's positions $\{\mathbf{p}^H_{i,t}\}^N_{i=1}$, Preferred person-following orientation $\theta$ and distance $d$, Robot maximum velocity $v_{\text{max}}$, Time step $\delta{t}$, MPC Horizon $N_{\text{hor}}$, Safe distance $d_{\text{safe}}$, Stopping criteria value $\epsilon$, Uncertainty threshold $\gamma$}
    \KwOut{Velocity $\mathbf{v}^{*}_t$}
    \If{Target visible}{
        Predicted target trajectory $\{\mathbf{p}^{T}_{t+k}\}^{N_{\text{hor}}}_{k=1}, \textit{uncertainty}=TrajectoryPredict(\mathcal{H}_t)$;\\
        Reference path $\mathcal{P}_t = \bigl\{\,\mathbf{p}^{T}_{t+k} - d[\cos(\theta), \sin(\theta)] \,\bigr\}_{k=1}^{N_{\text{hor}}} \in \mathbb{R}^{2 \times N_{\text{hor}}}$;\\
        \If{\textit{uncertainty} $> \gamma$}{
            \textbf{Return} $\mathbf{0}$
        }
    }
    \Else{
        $\mathbf{p}^{G}_{t} = KalmanPredict(\mathcal{H}_t)$;\\
        Reference path $\mathcal{P}_t = \begin{bmatrix}
                                        \mathbf{p}^G_t & \mathbf{p}^G_t & \cdots & \mathbf{p}^G_t
                                        \end{bmatrix}
                                        \in \mathbb{R}^{2 \times N_{\text{hor}}}$;\\
    }
    Obstacles $\mathcal{O}_t=ObstacleDetection(\mathcal{S})$;\\
    $C_{\text{track}}=TrackCost(\{\mathbf{v}_t, \mathbf{s}_t\}, \mathcal{P}_t, v_{\text{max}})$;\\
    $C_{\text{collision}}=CollisionCost(\{\mathbf{s}_t\}, \mathcal{O}_t, \{\mathbf{p}^H_{i,t}\}^N_{i=1},d_{\text{safe}})$;\\
    Bi-convex optimization $\mathcal{L}=Formulate(C_{\text{track}}, C_{\text{collision}})$;\\
    \For{iteration $k=\{1,2,\cdots,K\}$}{
        $\mathbf{v}^{*}_t, \textit{residual}= Solve(\mathcal{L})$;\\
        \If{$\textit{residual} < \epsilon$}{
            \textbf{break};
        }
    }
    \textbf{Return} $\mathbf{v}^{*}_t$
\caption{MPC-based planner with Traj.}
\label{alg:mpcTraj}
\end{algorithm}

\begin{algorithm}[t]
    \DontPrintSemicolon
    \SetNoFillComment
    \footnotesize
    \KwIn{2D LiDAR scan $\mathcal{S}$, Target person's historical trajectory $\mathcal{H}_t$,  Occluders historical trajectory $\{\mathcal{H}_{{\rm occ},i}\}_t$, Robot state $\mathbf{s}_{t}$, Target position $\mathbf{p}^{T}_{t}$, Other people's positions $\{\mathbf{p}^H_{i,t}\}^N_{i=1}$, Preferred person-following orientation $\theta$ and distance $d$, Robot maximum velocity $v_{\text{max}}$, Time step $\delta{t}$, MPC Horizon $N_{\text{hor}}$, Safe distance $d_{\text{safe}}$, Stopping criteria value $\epsilon$, overtaking minimal cost threshold $\delta_c$, `Viscosity' of the fluid flow $\gamma$}
    \KwOut{Velocity $\mathbf{v}^{*}_t$}
    \If{Target visible}{
        $\mathbf{p}^{G}_{t}=\mathbf{p}^{T}_{t}-d[\cos(\theta), \sin(\theta)]$;\\
    }
    \Else{
        \textcolor{blue}{\# Observation-based potential field overtaking} \\
        $\{\bar{\mathbf{p}}_{i,j}\}^{N_{\text{hor}}}_{j=1}, \_=TrajectoryPredict(\{\mathcal{H}_{{\rm occ},i}\}_t)$;\\
        $\mathbf{F}_{\text{rep}}(\cdot)=CalRepulsiveForce(\{\bar{\mathbf{p}}_{i,j}\}^{N_{\text{hor}}}_{j=1})$;\\
        $\mathbf{F}_{\text{att}}(\cdot)=CalAttractiveForce(\{\bar{\mathbf{p}}_{i,j}\}^{N_{\text{hor}}}_{j=1}, \mathbf{s}_{t})$;\\
        $\left( \mathbf{x}^* , C^* \right) = \left( \mathop{\arg\min}_{\{\mathbf{x}_i\}}, \min \right) \left( \|\mathbf{F}_{\text{att}}(\mathbf{x}_i)\| + \|\mathbf{F}_{\text{rep}}(\mathbf{x}_i)\| \right)$;\\
        \If{$C^* > \delta_c$}{
            \textcolor{blue}{\# Fluid field-based following} \\
            $\mathrm{v}(\cdot)=CalVelocityField(\{v_{{\rm occ},i}\}_t, \{\mathcal{H}_{{\rm occ},i}\}_t)$;\\
            $\rho_(\cdot)=CalDensityField(\{\mathcal{H}_{{\rm occ},i}\}_t)$;\\
            $\mathbf{x}^* = \mathop{\arg\min}_{\{\mathbf{x}_i\}} (\gamma \rho(\mathbf{x}_i) \cdot (\mathrm{v}_{\text{robot}} - \mathrm{v}(\mathbf{x}_i)) \cdot ||\mathbf{x}_i-\mathbf{s}_{t}||_2)$;\\
            }
        $\mathbf{p}^{G}_{t} = \mathbf{x}^*$;\\
    }
    Reference path $\mathcal{P}_t = \begin{bmatrix}
                                        \mathbf{p}^G_t & \mathbf{p}^G_t & \cdots & \mathbf{p}^G_t
                                        \end{bmatrix}
                                        \in \mathbb{R}^{2 \times N_{\text{hor}}}$;\\
    Obstacles $\mathcal{O}_t=ObstacleDetection(\mathcal{S})$;\\
    $C_{\text{track}}=TrackCost(\{\mathbf{v}_t, \mathbf{s}_t\}, \mathcal{P}_t, v_{\text{max}})$;\\
    $C_{\text{collision}}=CollisionCost(\{\mathbf{s}_t\}, \mathcal{O}_t, \{\mathbf{p}^H_{i,t}\}^N_{i=1},d_{\text{safe}})$;\\
    Bi-convex optimization $\mathcal{L}=Formulate(C_{\text{track}}, C_{\text{collision}})$;\\
    \For{iteration $k=\{1,2,\cdots,K\}$}{
        $\mathbf{v}^{*}_t, \textit{residual}= Solve(\mathcal{L})$;\\
        \If{$\textit{residual} < \epsilon$}{
            \textbf{break};
        }
    }
    \textbf{Return} $\mathbf{v}^{*}_t$
\caption{MPC-based planner with DS.}
\label{alg:mpcDS}
\end{algorithm}

\begin{algorithm}[t]
    \DontPrintSemicolon
    \SetNoFillComment
    \footnotesize
    \KwIn{2D LiDAR scan $\mathcal{S}$, Robot state $\mathbf{s}_{t}$, Target position $\mathbf{p}^{T}_{t}$, Other people's positions $\{\mathbf{p}^H_{i,t}\}^N_{i=1}$, Preferred person-following orientation $\theta$ and distance $d$, Robot maximum velocity $v_{\text{max}}$, Time step $\delta{t}$, Safe distance $d_{\text{safe}}$}
    \KwOut{Velocity $\mathbf{v}^{*}_t$}
    \If{Target visible}{
        $\mathbf{p}^{G}_{t}=\mathbf{p}^{T}_{t}-d[\cos(\theta), \sin(\theta)]$;\\
    }
    \Else{
        $\mathbf{p}^{G}_{t} = KalmanPredict(\mathcal{H}_t)$;\\
    }
    Obstacles $\mathcal{O}_t=ObstacleDetection(\mathcal{S})$;\\
    Admissible velocity $\mathcal{V}_t = CalDynamicWindow(\mathbf{v}_{t})$; \\
    \For{each $\mathbf{v}$ in $\mathcal{V}_t$}{
        \For{each $\mathbf{w}$ in $\mathcal{V}_t$}{
            $\mathbf{s}_{t+1}, \mathbf{v}_{t+1} = ForwardSim(\{ \mathbf{v}, \mathbf{w} \}, \mathbf{s}_{t}, \delta{t})$; \\
            $C_{\text{heading}} = HeadingCost(\mathbf{s}_{t+1}, \mathbf{p}^{G}_{t})$; \\
            $C_{\text{obs}} = ObstacleCost(\mathbf{s}_{t+1}, \mathcal{O}_t, \{\mathbf{p}^H_{i,t}\}^N_{i=1}, d_{\text{safe}})$; \\
            $C_{\text{vel}} = VelocityCost(\mathbf{v}_{t+1}, v_{\text{max}})$; \\
            $\textit{Cost} = CostFunction(C_{\text{heading}}, C_{\text{obs}}, C_{\text{vel}})$; \\
            \If{$\textit{Cost} < \textit{MiniCost}$}{
                $\mathbf{v}^{*}_t$ = $\{ \mathbf{v}, \mathbf{w} \}$; \\
                $\textit{MiniCost} = \textit{Cost}$;
            }
        }
    }
    \textbf{Return} $\mathbf{v}^{*}_t$
\caption{DWA-based planner}
\label{alg:dwa}
\end{algorithm}

\begin{algorithm}[t]
    \DontPrintSemicolon
    \SetNoFillComment
    \footnotesize
    \KwIn{2D LiDAR scan $\mathcal{S}$, Robot state $\mathbf{s}_{t}$, Target position $\mathbf{p}^{T}_{t}$, Other people's positions $\{\mathbf{p}^H_{i,t}\}^N_{i=1}$, Preferred person-following orientation $\theta$ and distance $d$, Robot maximum velocity $v_{\text{max}}$, Time step $\delta{t}$, Predict horizon $N_{\text{hor}}$, Safe distance $d_{\text{safe}}$}
    \KwOut{Velocity $\mathbf{v}^{*}_t$}
    \If{Target visible}{
        Predicted target trajectory $\{\mathbf{p}^{T}_{t+k}\}^{N_{\text{hor}}}_{k=1}, \_=TrajectoryPredict(\mathcal{H}_t)$;\\
        Reference path $\mathcal{P}_t = \bigl\{\,\mathbf{p}^{T}_{t+k} - d[\cos(\theta), \sin(\theta)] \,\bigr\}_{k=1}^{N_{\text{hor}}} \in \mathbb{R}^{2 \times N_{\text{hor}}}$;\\
    }
    \Else{
        $\mathbf{p}^{G}_{t} = KalmanPredict(\mathcal{H}_t)$;\\
        Reference path $\mathcal{P}_t = \begin{bmatrix}
                                        \mathbf{p}^G_t & \mathbf{p}^G_t & \cdots & \mathbf{p}^G_t
                                        \end{bmatrix}
                                        \in \mathbb{R}^{2 \times N_{\text{hor}}}$;\\
    }
    Obstacles $\mathcal{O}_t = ObstacleDetection(\mathcal{S})$;\\
    Admissible velocity $\mathcal{V}_t = CalDynamicWindow(\mathbf{v}_{t})$; \\
    \For{each $\mathbf{v}$ in $\mathcal{V}_t$}{
        \For{each $\mathbf{w}$ in $\mathcal{V}_t$}{
            $SimTraj = ForwardSim(\{ \mathbf{v}, \mathbf{w} \}, \mathbf{s}_{t}, N_{\text{hor}}, \delta{t})$; \\
            $C_{\text{heading}} = HeadingCost(SimTraj, \mathcal{P}_t)$; \\
            $C_{\text{dist}} = DistanceCost(SimTraj, \mathcal{P}_t)$; \\
            $C_{\text{obs}} = ObstacleCost(SimTraj, \mathcal{O}_t, \{\mathbf{p}^H_{i,t}\}^N_{i=1}, d_{\text{safe}})$; \\
            $C_{\text{vel}} = VelocityCost(SimTraj, v_{\text{max}})$; \\
            $\textit{Cost} = CostFunction(C_{\text{heading}}, C_{\text{dist}}, C_{\text{obs}}, C_{\text{vel}})$; \\
            \If{$\textit{Cost} < \textit{MiniCost}$}{
                $\mathbf{v}^{*}_t$ = $\{ \mathbf{v}, \mathbf{w} \}$; \\
                $\textit{MiniCost} = \textit{Cost}$;
            }
        }
    }
    \textbf{Return} $\mathbf{v}^{*}_t$
\caption{DWA-based planner with Traj.}
\label{alg:dwaTraj}
\end{algorithm}

\begin{algorithm}[t]
    \DontPrintSemicolon
    \SetNoFillComment
    \footnotesize
    \KwIn{2D LiDAR scan $\mathcal{S}$, Robot state $\mathbf{s}_{t}$, Robot velocity $\mathbf{v}_{t}$, Target person's position $\mathbf{p}^{T}_{t}$, Target trajectory history $\mathcal{H}_t$, Desired following distance $d$, B-spline order $p$, Control-point spacing $d_j$, Collision threshold $d_{\text{thr}}$, Dynamic limits $(v_{\max}, a_{\max})$, Cost weights $(\omega_c, \omega_g, \omega_s, \omega_d)$, MPC horizon $N_{\text{hor}}$}
    \KwOut{Velocity $\mathbf{v}^{*}_t$}

    \For{each time step $t$}{
        \If{Target visible}{
            $\mathbf{p}^{G}_{t} = \mathbf{p}^{T}_{t}$;\\
            Guidance path $\mathcal{G}_t = BuildGuidancePath(\mathcal{H}_t)$;\\
        }
        \Else{
            $\mathbf{p}^{G}_{t} = SearchReferencePoint(\mathcal{H}_t)$;\\
            Guidance path $\mathcal{G}_t = BuildGuidancePath(\mathcal{H}_t, \mathbf{p}^{G}_{t})$;\\
        }

        Local map $\mathcal{M}_t = BuildLocalMap(\mathcal{S}, \mathbf{s}_{t})$;\\
        Front-end goal $\mathbf{p}^{F}_{t} = CalFrontendGoal(\mathbf{p}^{G}_{t}, \mathcal{M}_t)$;\\
        Initial path $\mathcal{P}^{A}_{t} = HybridAStar(\mathbf{s}_{t}, \mathbf{p}^{F}_{t}, \mathcal{M}_t)$;\\

        Control points $\{\mathbf{p}_i\}_{i=0}^{N} = InitControlPoints(\mathcal{P}^{A}_{t}, p)$;\\
        Reference points $\{\mathbf{g}_i\}_{i=0}^{N} = UniformSample(\mathcal{G}_t, N+1)$;\\

        $d_t = CalFollowDistance(\mathbf{s}_{t}, \mathbf{p}^{G}_{t})$;\\
        $v_{\text{ref}} = IncrementalPID(d_t - d)$;\\
        $\Delta t = d_j / v_{\text{ref}}$;\\

        $J = (\omega_c, \omega_g, \omega_s, \omega_d, d_{\text{thr}}, v_{\max}, a_{\max})$;\\
        $\mathcal{P}^{B}_{t} = BSplineOptimize(\{\mathbf{p}_i\}_{i=0}^{N}, \{\mathbf{g}_i\}_{i=0}^{N}, \mathcal{M}_t, \Delta t, J)$;\\

        $\mathbf{v}^{*}_t = MPCTrack(\mathcal{P}^{B}_{t}, \mathbf{s}_{t}, \mathbf{v}_{t}, N_{\text{hor}})$;\\
    }
    \textbf{Return} $\mathbf{v}^{*}_t$
\caption{BSO-HFC planner}
\label{alg:bsohfc}
\end{algorithm}

\begin{algorithm}[t]
    \DontPrintSemicolon
    \SetNoFillComment
    \footnotesize
    \KwIn{2D LiDAR scan $\mathcal{S}$, Robot state $\mathbf{s}_{t}$, Target person's historical trajectory $\mathcal{H}_t$, Target position $\mathbf{p}^{T}_{t}$ if visible, Follow mode $m$, Desired distance $d$, Time step $\delta{t}$, Search budget $\tau$, Discount factor $\gamma$, RL value model $V$}
    \KwOut{Velocity $\mathbf{v}^{*}_t$}
    \If{Target visible}{
        $\hat{\mathbf{p}}^{T}_{t} = \mathbf{p}^{T}_{t}$;\\
    }
    \Else{
        $\hat{\mathbf{p}}^{T}_{t} = TargetStatePredict(\hat{\mathbf{p}}^{T}_{t-1},\mathcal{H}_t,\delta{t})$;\\
    }
    Obstacles $\mathcal{O}_t = ObstacleDetection(\mathcal{S})$;\\
    $\pi^{H}_t = HumanActionPredict(\mathcal{H}_t)$;\\
    Root node $n_0 = (\mathbf{s}_{t},\hat{\mathbf{p}}^{T}_{t})$;\\
    \While{time $< \tau$}{
        $n \gets SelectNodeByUCB(n_0,\pi^{H}_t)$;\\
        $c \gets Expand(n,\mathcal{O}_t)$;\\
        \If{$c = \varnothing$}{
            \textbf{continue};\\
        }
        $\mathbf{x}_{t+1} \gets StateTransition(c,\delta{t})$;\\
        $R_{\mathrm{imm}} = Reward(\mathbf{x}_{t+1},m,d)$;\\
        $R_{\mathrm{val}} = V(\mathbf{x}_{t+1},m,d)$;\\
        $R = R_{\mathrm{imm}} + \gamma R_{\mathrm{val}}$;\\
        $BackPropagate(c,R)$;\\
    }
    $a^{*}_t = MostVisitedChild(n_0)$;\\
    $\mathbf{v}^{*}_t = ActionToVelocity(a^{*}_t)$;\\
    \textbf{Return} $\mathbf{v}^{*}_t$
\caption{RL-based planner}
\label{alg:rl}
\end{algorithm}

The \textbf{SFM}-based planner~\cite{ferrer2017robot} follows Algorithm~\ref{alg:sfm}, with hyper-parameters $\eta_{\mathrm{trk}}$, $\eta_{\mathrm{sa}}$, $\eta_{\mathrm{sb}}$, $\eta_{\mathrm{ha}}$, $\eta_{\mathrm{hb}}$, $\eta_{\mathrm{cut}}$, $\eta_{\mathrm{safe}}$, $\eta_{\mathrm{wa}}$, and $\eta_{\mathrm{wr}}$ passed into \textsf{CalAttractive}, \textsf{CalRepulsive}, and \textsf{CalInteraction}; numerical values are in Table~\ref{tab:appendixHyperparams}.
Classical social-force interpretations are discussed in~\cite{ferrer2017robot,helbing1995social}.
The \textbf{MPC}-based planner utilizes the RDA-planner~\cite{han2023rda} to trace the navigation point while avoiding collisions with static obstacles and pedestrians. We chose the RDA-planner for its efficient collision constraint formulation, which takes the form of a smooth bi-convex dual, facilitating subsequent parallelization. The detailed algorithm is shown in Algorithm~\ref{alg:mpc}. The \textbf{MPC w/ Traj.} Follow-Bench variant (Algorithm~\ref{alg:mpcTraj}) combines this RDA backbone with reference-path tracking based on the target's predicted future trajectory. Inspired by~\cite{sekiguchi2021uncertainty}, it also commands the robot to brake when prediction uncertainty exceeds a threshold. Specifically, we employ a Kalman filter to predict the target's future trajectory and measure uncertainty using the normalized innovation squared value. This composite implementation is therefore not presented as an unchanged reproduction of~\cite{sekiguchi2021uncertainty}.

The \textbf{MPC w/ DS.} Follow-Bench variant (Algorithm~\ref{alg:mpcDS}) combines the RDA backbone with the dynamic search field of~\cite{ye2025rpf} for person-search after occlusion. The field consists of two modes: overtaking and fluid-following. In overtaking mode, repulsive forces penalize proximity to occluders, while attractive forces encourage minimal travel cost and improved visibility of the target. When overtaking costs exceed a predefined threshold, the planner switches to fluid-following mode, where the crowd is modeled as a flow field. Velocity and density fields are constructed from pedestrian trajectory histories, and the robot's goal is selected by minimizing a cost function that balances robot velocity, local crowd velocity, and density. This encourages the robot to follow regions of lower density and align with pedestrian flow, thereby improving safety and comfort. Results for MPC w/ DS. consequently characterize this composite implementation rather than either component in isolation.

The \textbf{DWA}-based planner \cite{van2022collision} samples admissible linear and angular velocities within the dynamic window and performs forward simulation for each candidate. Trajectories are evaluated using a cost function that combines goal alignment, obstacle clearance, and velocity preference, and the optimal velocity is selected as detailed in Algorithm~\ref{alg:dwa}. However, this method considers only the robot's immediate next state, failing to fully exploit the predictive capability of forward simulation and often resulting in shortsighted behaviors. \textbf{DWA w/ Traj.} (Algorithm~\ref{alg:dwaTraj}) is a Follow-Bench extension that adds multi-step rollout, tracking of the target's predicted trajectory, and a refined cost function to the DWA core. In particular, the heading cost integrates short-term and long-term orientation objectives, while the distance cost ensures real-time target tracking. The trajectory prediction module is identical to that used in the MPC w/ Traj. planner.

The \textbf{BSO-HFC} planner~\cite{lyu2025robust} employs a hierarchical planning-and-tracking framework that begins with a lightweight Hybrid A* search within the local map to generate an initial collision-free path. This front-end path is subsequently refined by a back-end uniform B-spline optimizer, which jointly penalizes obstacle proximity, dynamic infeasibility, and deviations from a guidance path derived from the target's trajectory history. To maintain the desired following distance, an incremental PID controller adaptively regulates the reference speed, intrinsically dictating the B-spline time allocation. Ultimately, an MPC tracker executes this optimized trajectory to produce precise robot control commands. The detailed procedure is outlined in Algorithm~\ref{alg:bsohfc}.

The \textbf{RL-based} Follow-Bench implementation extends the original work~\cite{leisiazar2025adapting} by additionally supporting back- and side-following. To achieve this, we reformulate the follow task in a human-centered local frame and define mode-dependent desired relative poses for different follow behaviors. To support these new follow modes, we construct a dedicated follow-task training environment and train separate A2C-based value networks for different follow modes and desired distances. During planning, the target state is directly updated from observation when the target is visible. Otherwise, it is propagated from motion history using an LSTM-based human-action prior. The resulting robot-target state is then used in Monte Carlo Tree Search (MCTS). Each candidate branch is scored based on an immediate follow reward and the learned value function. The final control command is selected from the most visited branch, as detailed in Algorithm~\ref{alg:rl}. The implementation thus preserves the original human-action-aware search structure while extending its task formulation and learned value models; the reported results apply to this extended implementation.

In all simulated scenarios, the baseline configuration models each human as an omni-directional dynamic cylinder (radius: $0.3 \ \mathrm{m}$, max speed: $\pm 1.5 \ \mathrm{m/s}$) and the robot as a differential-drive platform ($0.5 \ \mathrm{m} \times 0.7 \ \mathrm{m}$) equipped with a 2D LiDAR ($5.0 \ \mathrm{m}$ range), subject to $\pm 1.5 \ \mathrm{m/s}$ translational velocity and $\pm 2.0 \ \mathrm{m/s}^2$ acceleration limits. As for varied human densities and environmental occupancy conditions, the maximum pedestrian densities, obstacle numbers, and minimum passage widths are determined empirically by observing conditions under which simulated pedestrians become prone to congestion.

\subsection{Detailed Experimental Results} \label{sec:appendixResults}
\begin{figure*}[t]
    \centering
    \includegraphics[width=\linewidth]{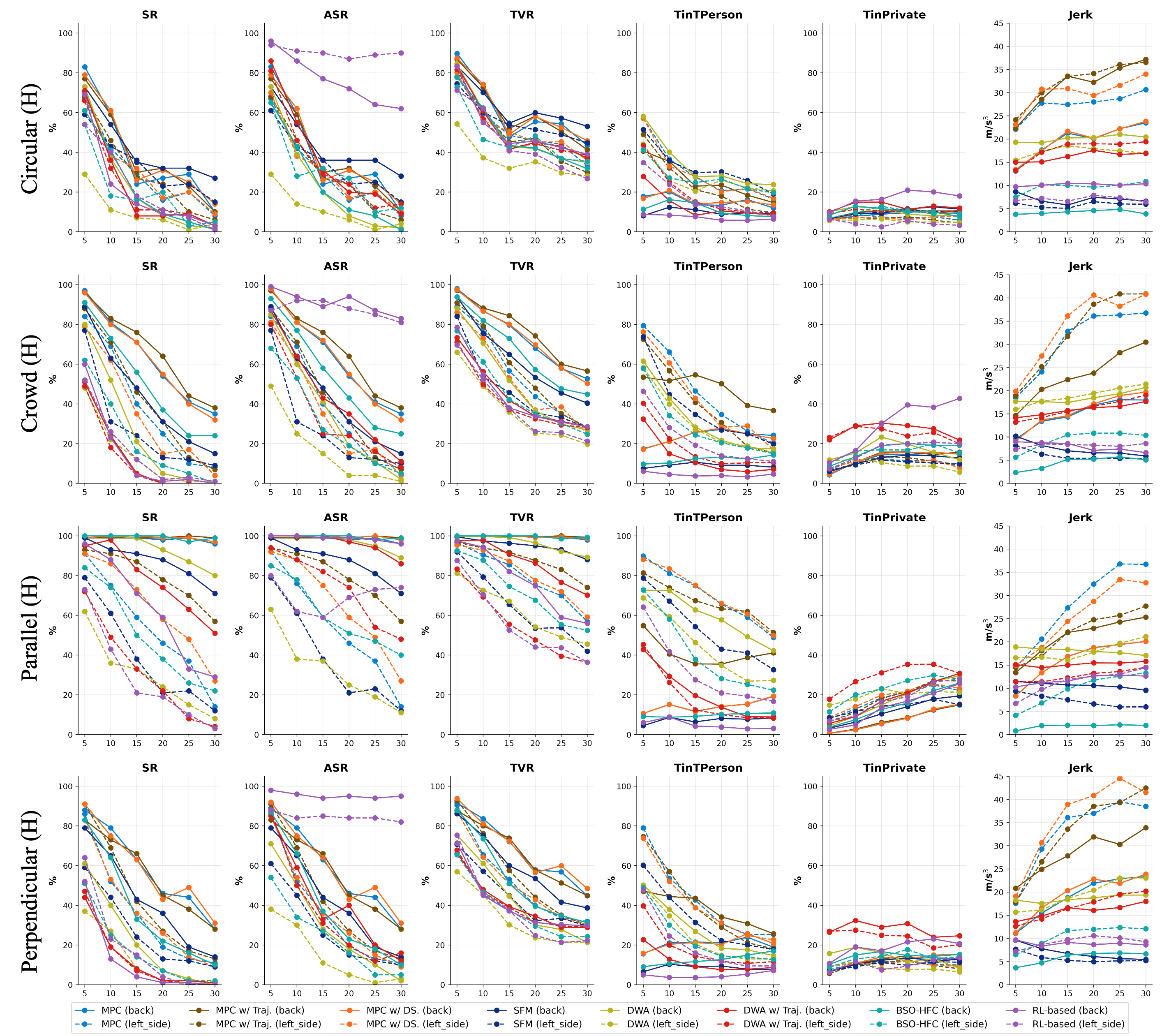}
    \caption{Performance under varying numbers of humans in dynamic crowd flows. The number ranges from 5 to 30. In parallel flows, MPC-based RPF planners with back-following consistently achieve stable performance, as their back-following strategy aligns with the parallel movement patterns and reduces occlusion.}
    \label{fig:expDynamicHuman}
    \vspace{-0.3cm}
\end{figure*}

\begin{figure*}[t]
    \centering
    \includegraphics[width=\linewidth]{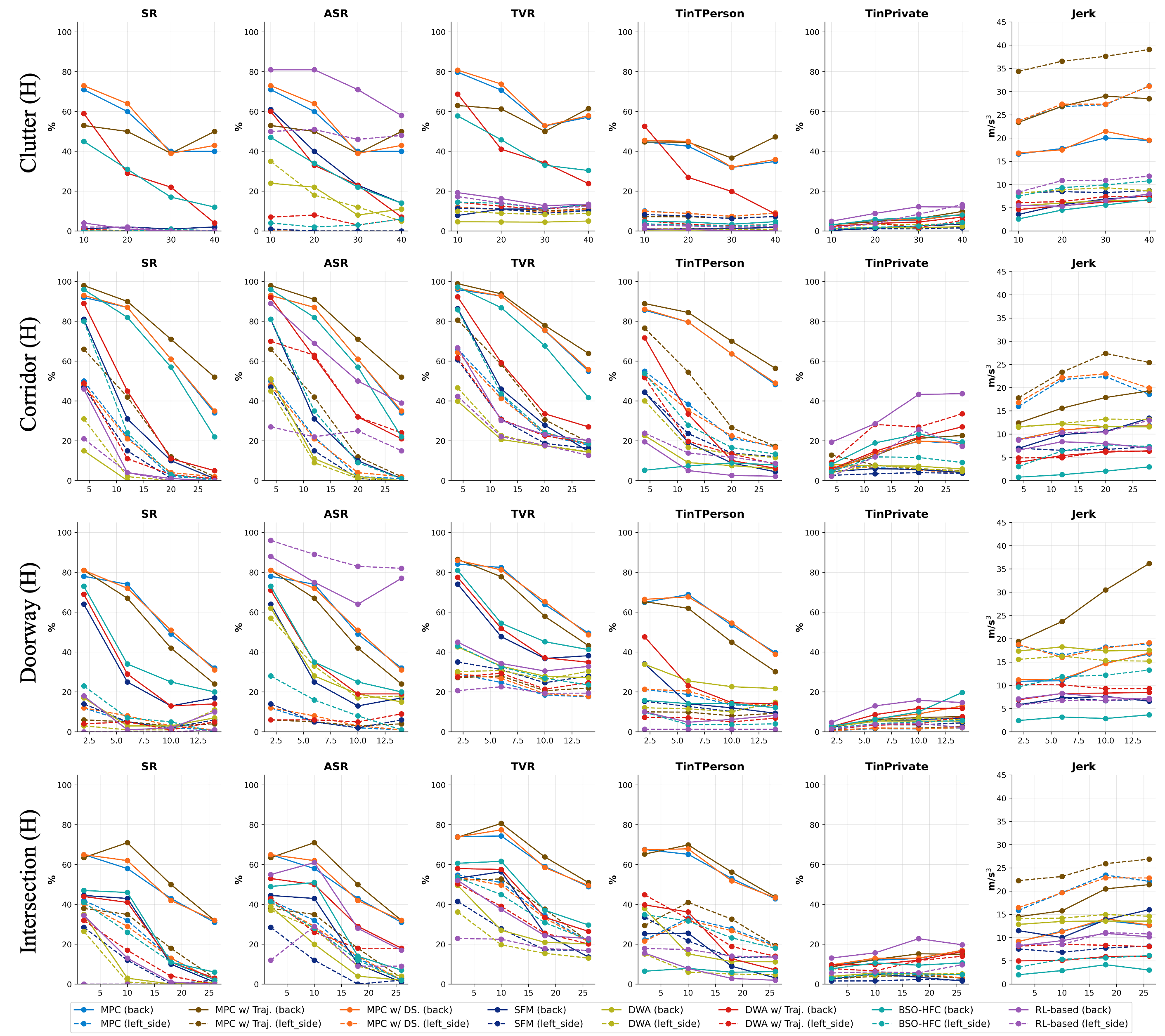}
    \caption{Performance under varying numbers of humans in environmental layouts. The number ranges from 5 to 30. In constrained structural scenarios, denser crowds trigger more frequent interactions, leading to unpredictable human movements and reduced free space. These conditions pose significant challenges for RPF planners, resulting in higher discomfort and an elevated risk of collisions.}
    \label{fig:expLayoutHuman}
    \vspace{-0.3cm}
\end{figure*}

\begin{figure*}[t]
    \centering
    \includegraphics[width=\linewidth]{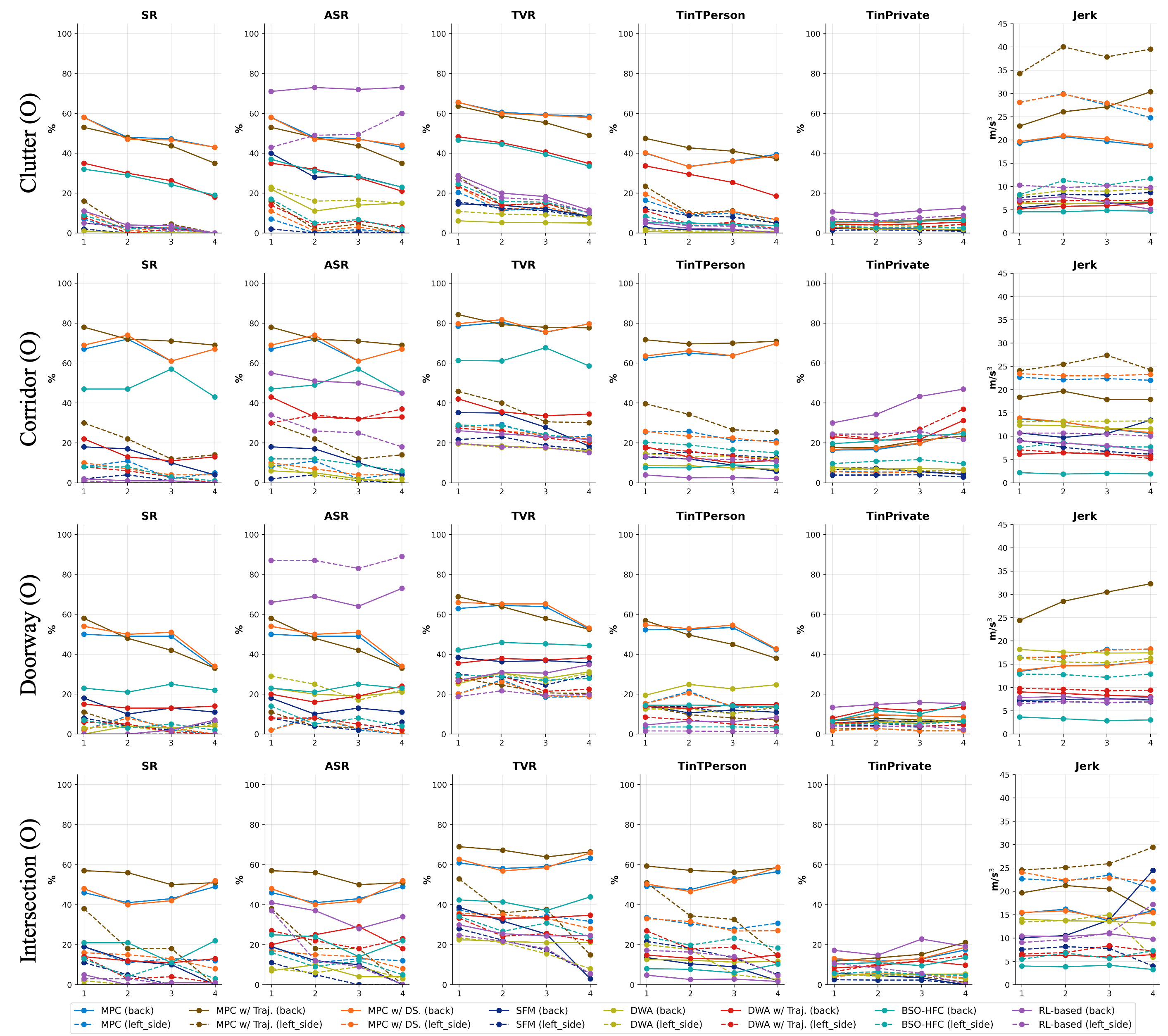}
    \caption{Performance under different environmental occupied conditions in environmental layouts. Their occupied conditions are: clutter (10, 20, 30, 40 obstacles), corridor (6.8, 6.2, 5.6, 5.0 m width), intersection (6.0, 5.4, 4.8, 4.2 m width), and doorway (3.4, 3.1, 2.8, 2.5 m width).}
    \label{fig:expLayoutOccupancy}
    \vspace{-0.3cm}
\end{figure*}

\begin{figure*}[t]
    \centering
    \includegraphics[width=\linewidth]{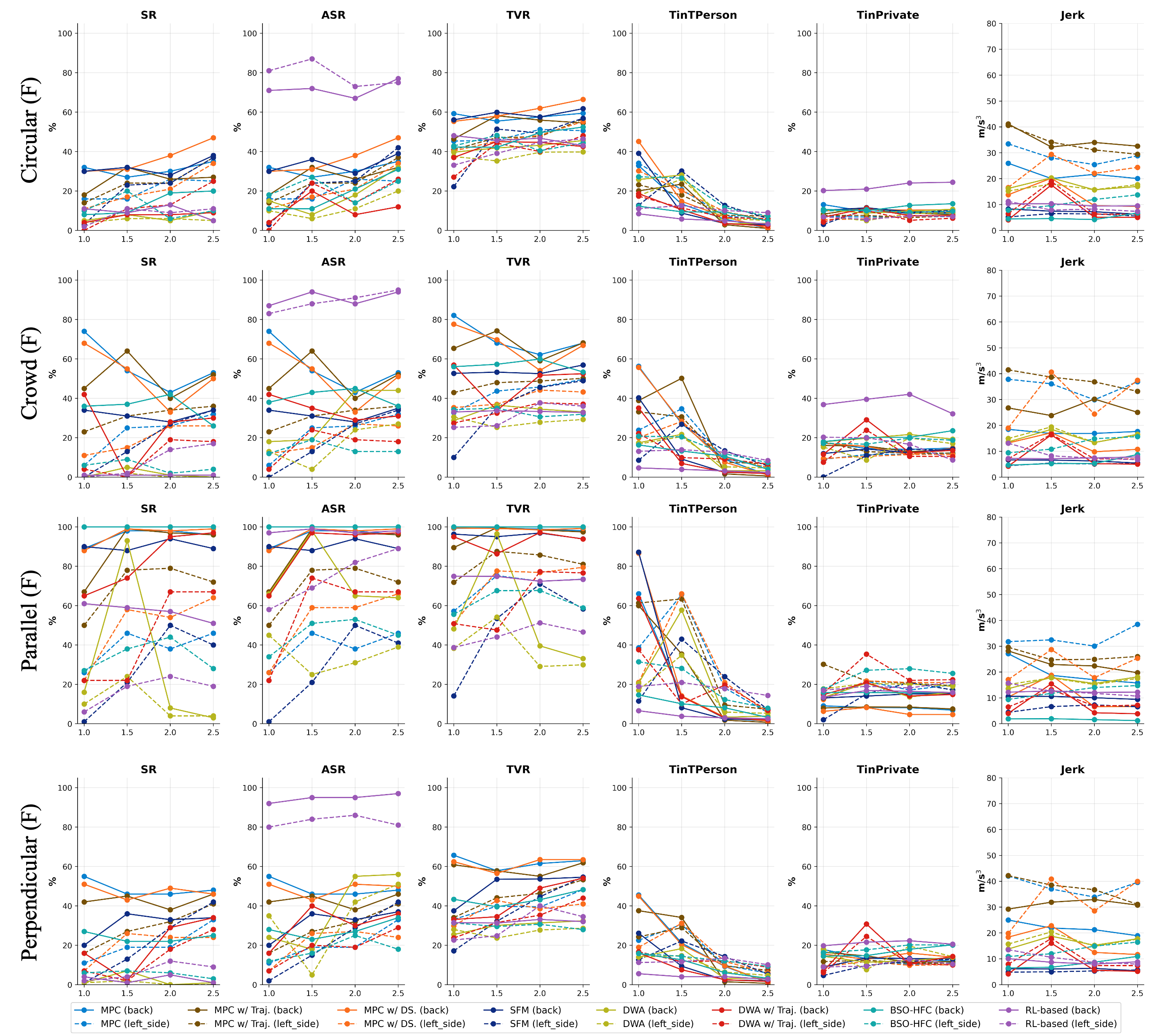}
    \caption{Performance under varying following distances in dynamic crowd flows. The following distance varies from 1.0 m to 2.5 m to assess its impact on planner performance.}
    \label{fig:expDynamicFDist}
    \vspace{-0.3cm}
\end{figure*}

\begin{figure*}[t]
    \centering
    \includegraphics[width=\linewidth]{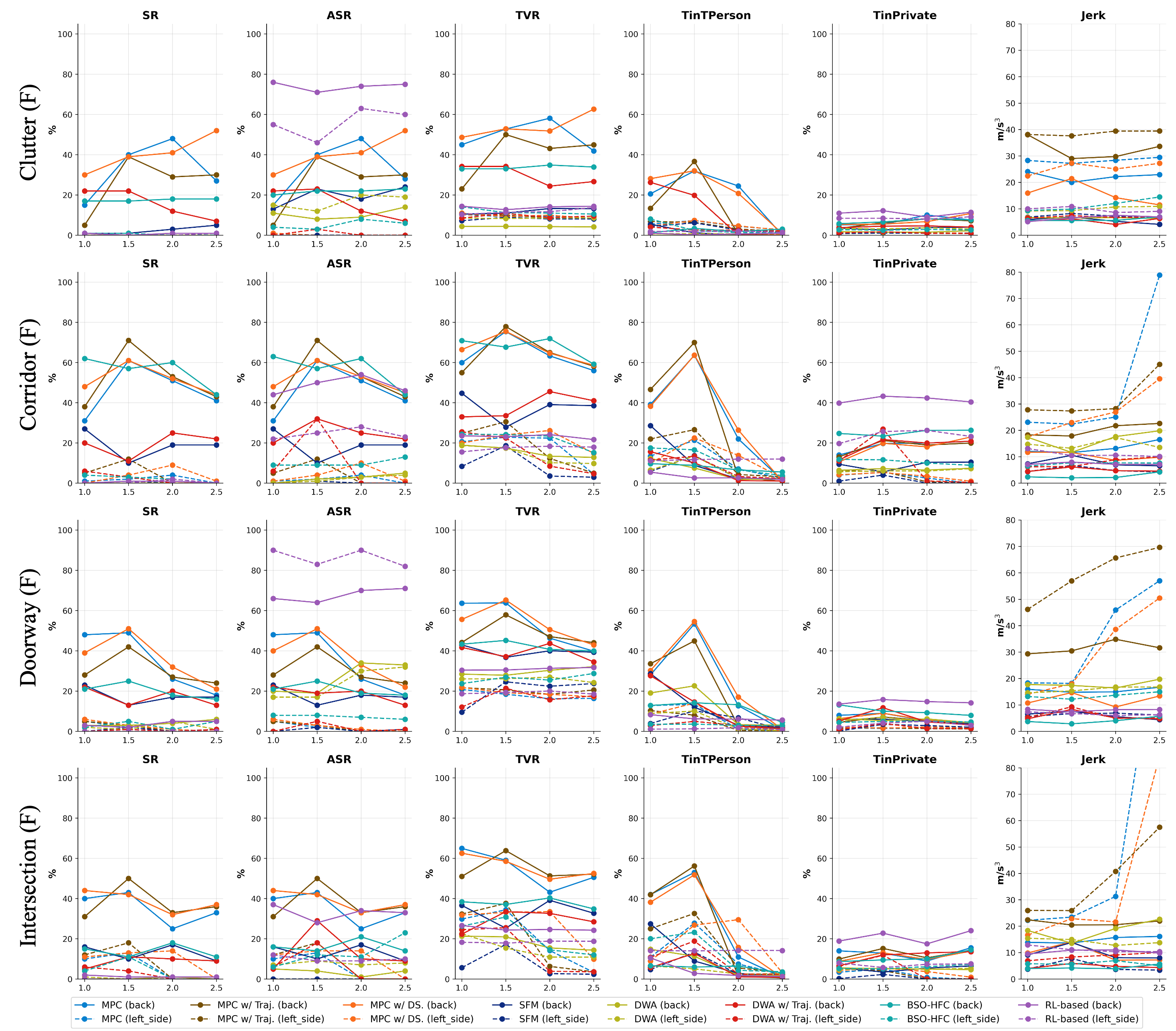}
    \caption{Performance under varying following distances in environmental layouts. The following distance varies from 1.0 m to 2.5 m to assess its impact on planner performance.}
    \label{fig:expLayoutFDist}
    \vspace{-0.3cm}
\end{figure*}

Here, we detail our experimental results of every single scenario under different numbers of humans, occupied conditions and following distances. Fig.~\ref{fig:expDynamicHuman} and Fig.~\ref{fig:expLayoutHuman} show the performance under varying numbers of humans in dynamic crowd flows and environmental layouts, respectively. Fig.~\ref{fig:expLayoutOccupancy} shows the performance under different environmental occupied conditions in environmental layouts. Fig.~\ref{fig:expDynamicFDist} and Fig.~\ref{fig:expLayoutFDist} show the performance under varying following distances in dynamic crowd flows and environmental layouts, respectively. More evaluation results of other evaluation metrics can be found in the released code.

The obstacle-representation sensitivity study shows higher observed MPC success rates with finer boxes and circular obstacles than with the default coarse boxes in \textit{Layout: Cluttered Space} (Table~\ref{tab:obsSensitivity}). To examine this potential source of implementation bias, we evaluate 100 trials with 20 static obstacles and 30 pedestrians under back-following at 1.5\,m. We compare three representations constructed from 2D LiDAR observations: (i)~native coarse minimum-area bounding boxes using DBSCAN with $\texttt{eps}=2.0$, $\texttt{min\_samples}=6$; (ii)~finer boxes using $\texttt{eps}=0.5$, $\texttt{min\_samples}=3$, matching SFM's clustering settings; and (iii)~minimum-enclosing circles with the same finer clustering settings and circular human representations. MPC's observed SR is 44.0\% in the original evaluation and 46.0\% in the repeat evaluation with coarse boxes, compared with 54.0\% for finer boxes and 56.0\% for circles. This pattern is consistent with coarse clustering hindering passage through gaps by merging adjacent obstacles. Under all three representations, MPC's observed SR remains above those of the non-MPC baselines in this setting, where BSO-HFC and DWA reach 29.0\% and 28.0\%, respectively. These results support the observed MPC advantage across the tested representations in this setting, but do not rule out other implementation effects.

\begin{table}[t]
    
    \centering
    \caption{Obstacle-representation sensitivity in \textit{Layout: Cluttered Space} (20 obstacles, 30 humans, back-following at 1.5\,m, 100 trials).}
    \label{tab:obsSensitivity}
    \scriptsize
    \renewcommand{\arraystretch}{1.15}
    \resizebox{\columnwidth}{!}{%
    \begin{tabular}{l l c}
    \toprule
    \textbf{Planner / Variant} & \textbf{Obstacle Representation (2D LiDAR)} & \textbf{SR$\uparrow$ (\%)} \\
    \midrule
    MPC (default reported) & Coarse boxes ($\texttt{eps}=2.0, \texttt{min\_samples}=6$) & 44.0 \\
    MPC (default repeat evaluation) & Coarse boxes ($\texttt{eps}=2.0, \texttt{min\_samples}=6$) & 46.0 \\
    MPC (fine-box variant) & Fine boxes ($\texttt{eps}=0.5, \texttt{min\_samples}=3$) & 54.0 \\
    MPC (circle variant) & Enclosing circles ($\texttt{eps}=0.5, \texttt{min\_samples}=3$) & \textbf{56.0} \\
    \midrule
    MPC w/ Traj. & Coarse boxes ($\texttt{eps}=2.0, \texttt{min\_samples}=6$) & 52.0 \\
    BSO-HFC & Local occupancy grid + EDT & 29.0 \\
    DWA & Raw LiDAR obstacle point cloud & 28.0 \\
    DWA w/ Traj. & Raw LiDAR obstacle point cloud & 22.0 \\
    SFM & Fine cluster centers ($\texttt{eps}=0.5, \texttt{min\_samples}=3$) & 11.0 \\
    RL-based & Robot-centric inflated occupancy grid & 4.0 \\
    \bottomrule
    \end{tabular}%
    }
\end{table}

\end{document}